\pdfoutput=1
\documentclass[11pt]{article}
\usepackage[preprint]{acl}

\usepackage{times}
\usepackage{latexsym}

\usepackage[T1]{fontenc}
\usepackage[utf8]{inputenc}

\usepackage{microtype}

\usepackage{inconsolata}

\usepackage{graphicx}

\usepackage{hyperref}
\usepackage{url}
\usepackage{algorithm}
\usepackage{algpseudocode}
\usepackage{subcaption} % Add this to the preamble
\usepackage{titletoc}

\usepackage{booktabs}
\usepackage[dvipsnames,table]{xcolor}
\usepackage{tikz}
\usetikzlibrary{matrix, calc}
\usepackage{amsthm}

\newtheorem{definition}{Definition}

\usepackage{mathtools}
\usepackage{multirow}
\usepackage{titlesec}
\usepackage{svg}
\usepackage{booktabs}
\usepackage[utf8]{inputenc}
\usepackage{wrapfig}
\usepackage{pifont}

\definecolor{oursblue}{RGB}{232,240,252}
\usepackage{amsmath,amsfonts,bm}

\def\eqref#1{equation~\ref{#1}}
\def\1{\bm{1}}

\DeclareMathAlphabet{\mathsfit}{\encodingdefault}{\sfdefault}{m}{sl}
\SetMathAlphabet{\mathsfit}{bold}{\encodingdefault}{\sfdefault}{bx}{n}

\usepackage{booktabs,multirow,makecell}
\usepackage{siunitx}
\usepackage[most]{tcolorbox}

\newtcolorbox{settingbox}{
    colback=gray!4,
    colframe=black,
    boxrule=0.5pt,
    arc=1.5pt,
    left=4pt,
    right=4pt,
    top=4pt,
    bottom=4pt,
    before skip=4pt,
    after skip=4pt
}

\usepackage{colortbl}
\usepackage[dvipsnames]{xcolor}
\usepackage{array}
\usepackage[normalem]{ulem}
\useunder{\uline}{\ul}{}

\newcommand{\best}[1]{\textbf{#1}}
\newcommand{\second}[1]{\uline{#1}}

\usepackage{array}
\newcolumntype{C}[1]{>{\centering\arraybackslash}m{#1}}
\newcolumntype{L}[1]{>{\raggedright\arraybackslash}m{#1}}
\newcolumntype{Y}{>{\raggedright\arraybackslash}X}

\usepackage{booktabs}
\usepackage{pifont}

\title{GRID: Scaling Task-Agnostic Inference in Continual Prompt Tuning}
\author{
\textbf{Anushka Tiwari}$^{\ddagger}$\thanks{These authors contributed equally to this work.} \quad
\textbf{Sayantan Pal}$^{\dagger}$\footnotemark[1] \quad
\textbf{Rohini K. Srihari}$^{\dagger}$ \quad
\textbf{Kaiyi Ji}$^{\dagger}$ \\
State University of New York at Buffalo \\
$^{\dagger}$Department of Computer Science and Engineering \\
$^{\ddagger}$Institute for Artificial Intelligence and Data Science \\
\texttt{\{atiwari4, spal5, rohini, kaiyiji\}@buffalo.edu}
}

\definecolor{spal}{RGB}{230,120,20}

\begin{document}\maketitle

% Prompt-based continual learning (CL) provides a parameter-efficient approach for adapting large language models (LLMs) across task sequences. However, most existing methods rely on task-aware inference and maintain a growing set of task-specific prompts, which introduces two major challenges: (1) severe performance degradation on earlier tasks when task identifiers are unavailable for prompt selection at inference time, 
% and (2) limited scalability due to prompt memory accumulation as task sequences grow.
% In this paper, we present GRID, a unified framework designed to address these challenges. GRID incorporates an output-space–aware decoding mechanism that enhances backward transfer by leveraging representative inputs and automatic label semantic normalization, alongside a 
% gradient-guided prompt selection strategy that compresses less informative prompts into a 
% single aggregated representation for scalable, memory-efficient continual learning. Extensive experiments on long-sequence and negative-transfer benchmarks show that GRID improves average accuracy and backward transfer, achieves competitive forward transfer, and substantially reduces prompt memory usage across both encoder-decoder and decoder-only architectures, such as T5, Qwen, and LLaMA.

% \end{abstract}

\begin{abstract}
Prompt-based continual learning (CL) offers a parameter-efficient way to adapt large language models (LLMs) across task sequences. However, existing methods often rely on task-aware inference and maintain an expanding set of task-specific prompts, leading to (1) severe performance degradation on earlier tasks when task identifiers are unavailable for prompt selection at inference time, and (2) limited scalability as task sequence grows. We propose GRID, a unified framework designed to address these challenges. GRID incorporates an output-space–aware decoding mechanism that enhances backward transfer by leveraging representative inputs and automatic label semantic normalization, alongside a 
gradient-guided prompt selection strategy that compresses less informative prompts into a 
single aggregated representation for scalable, memory-efficient continual learning. Extensive experiments on long-sequence and negative-transfer benchmarks show that GRID improves backward transfer, achieves competitive forward transfer, and substantially reduces prompt memory across encoder-decoder and decoder-only architectures, including T5, Qwen, and LLaMA. Source code is available \href{https://github.com/AnushkaTi/GRID}{here}.
\end{abstract}

\section{Introduction}

\label{sec:intro}

% \kj{carefully reorganize this introduction to make everything more smooth; let us revise it to make it less overclaimed}

% \vspace{-0.}

Continual learning (CL) \cite{van2019three} enables models to learn from a sequence of tasks without retraining from scratch. CL systems build on prior knowledge while adapting to new tasks, which is crucial in dynamic real-world settings. Recent advancements, especially in NLP \cite{wang2024comprehensive, satapara-srijith-2024-tl}, have focused on three paradigms: \textit{regularization-based methods} \cite{li2017learning, kirkpatrick2017overcoming}, \textit{rehearsal-based methods} \cite{rebuffi2017icarl, sun2019lamol}, and \textit{architecture-based methods} \cite{veniat2020efficient}. While rehearsal-based methods are effective, they are impractical in privacy-sensitive scenarios \cite{kirkpatrick2017overcoming}. With the increasing complexity of pre-trained models \cite{wang2023pre}, full model finetuning has become infeasible. This has led to the rise of parameter-efficient finetuning (PEFT) techniques \cite{ding2023parameter}, such as \textit{prompt tuning} (PT) \cite{lester-etal-2021-power}, which adapts large models by training only a small set of soft prompts, requiring less than 0.01\% of the model's parameters.

\begin{figure}[htbp]
    \centering
    \includegraphics[width=0.5\textwidth]{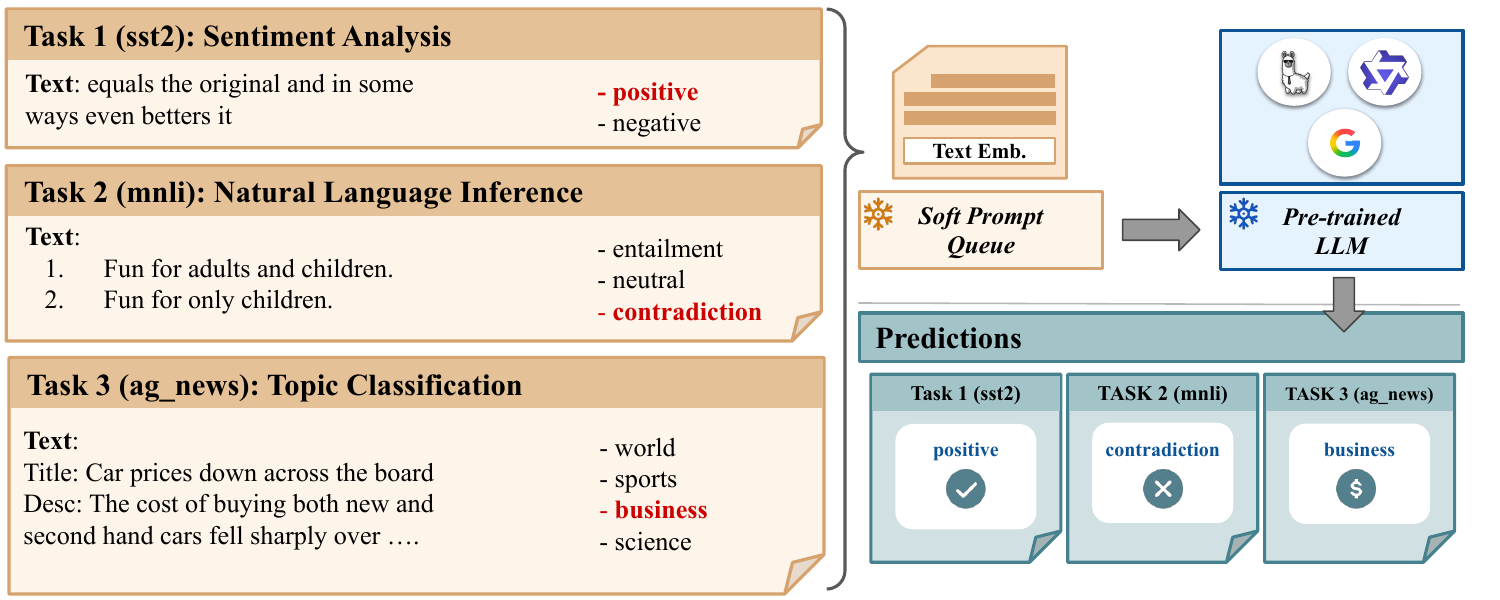}
    \caption{\small Illustration of GRID, task-agnostic inference in prompt-based continual learning. Multiple tasks with distinct input formats and label spaces are processed using a shared, fixed-capacity prompt pool and a frozen pre-trained model, without access to task identities at inference time.}
    \label{fig:intro}
\end{figure}

% Building on this, continual prompt tuning (CPT) extends PT to the CL setting by enabling models to learn task-specific prompts sequentially without modifying the base model \cite{wang2022learning}. Recent advancements in prompt tuning-based continual learning address key challenges like catastrophic forgetting and forward knowledge transfer. \cite{wang2022dualprompt} introduced a dual prompt framework with a shared and task-specific prompt, though this approach faces limitations in retaining knowledge from past tasks. To mitigate these issues, ProgPrompt \cite{razdaibiedina2023progressive}, SHLPT \cite{wu-etal-2024-mitigate}, Q-Tuning \cite{guo2024q} maintains a list of prompts for each task, progressively adding new ones while retaining old ones.

Building on this, continual prompt tuning (CPT) extends PT to the CL setting, enabling sequential task adaptation without modifying the base model \cite{wang2022learning}. Early work maintained a single shared prompt pool queried by input features \cite{wang2022learning}, but retention suffered as a single pool struggled to encode knowledge across diverse tasks. Subsequent methods improved retention by introducing task-specific prompts, either through complementary shared and expert prompt \cite{wang2022dualprompt}, or by maintaining a dedicated prompt per task, notably ProgPrompt \cite{razdaibiedina2023progressive}, SHLPT \cite{wu-etal-2024-mitigate}, Q-Tuning \cite{guo2024q}, achieving strong knowledge preservation.

Although prompt-based CL  has received significant attention, two main challenges remain:  
% \begin{list}{$\bullet$}{\topsep=0.3ex \leftmargin=0.15in \rightmargin=0.in \itemsep =0.022in}
% \item
% \kj{challengs parts here can be shortened}
% First, while major prompt-based CL methods such as Progressive Prompts \cite{razdaibiedina2023progressive}, Q-Tuning \cite{guo2024q}, and SHLPT \cite{wu-etal-2024-mitigate} report zero forgetting, thereby fully preserving prior knowledge,
% they depend on \emph{task-aware prompt selection at inference time}, where task identities are explicitly provided to retrieve the correct prompt for each previous task. In practice, however, task IDs are often unavailable during inference for prompt selection or routing \cite{wang2024comprehensive, liang2023loss}. In practical NLP systems, models often receive free-form inputs, and deployed continual models must handle mixed inputs from past tasks without explicit task labels, requiring task-agnostic inference.
First, major methods such as ProgPrompt \cite{razdaibiedina2023progressive}, Q-Tuning \cite{guo2024q}, and SHLPT \cite{wu-etal-2024-mitigate} depend on \emph{task-aware inference}: the task identity is provided at test time to select the corresponding prompt. This assumption allows them to report near-zero forgetting, but in practice, task IDs are often unavailable during inference for prompt selection \cite{wang2024comprehensive, liang2023loss}. In practical NLP systems, models often receive free-form inputs, and deployed continual models must handle mixed inputs from past tasks without explicit task labels, requiring task-agnostic inference (Figure~\ref{fig:intro}). Although vision-based class-incremental learning (CIL) methods \cite{tran2025boosting, mcdonnell2023ranpac} also consider task-agnostic prediction, they typically assume a shared classification space. However, our setting involves heterogeneous tasks with diverse behaviors and output spaces, making these approaches inapplicable.

% In practice, however, task IDs are often unavailable during inference \cite{wang2024comprehensive, liang2023loss}.

% \item 
Second, existing continual prompt tuning methods struggle to scale efficiently in both time and memory as the prompt queue expands. ProgPrompt \cite{razdaibiedina2023progressive} and SHLPT \cite{wu-etal-2024-mitigate} assign a dedicated soft prompt to each task, causing memory to grow linearly with the task sequence. More recent approaches attempt to mitigate this issue by continually updating the same set of prompts over time \cite{wang2022learning} or by using PCA-based eviction \cite{guo2024q}. However, these solutions either suffer from redundant prompt accumulation due to the lack of effective pruning or merging mechanisms, or incur significant computational overhead (e.g., repeated SVD during eviction).
% \end{list}

% \kj{this paragraph should put to challenges part? something like why class-incremental learning cannot be applied to prompt based CL?}
% \textcolor{blue}{One might ask whether class-incremental learning (CIL) methods from the vision domain address these challenges. However, vision CIL operates under a fundamentally different assumption: each task introduces new object classes into a single shared classifier head, and the goal is selecting the correct class among all seen so far. In this work, we follow the task-incremental (TIL) setting in NLP, where tasks are heterogeneous with different label formats and semantics (e.g., sentiment, entailment, QA). Here the core challenge is producing the correct task behavior without a task identifier, a mismatch in problem structure that makes direct application of vision CIL methods non-trivial.}

As shown in Table~\ref{tab:method_comparison}, existing prompt-based CL methods in NLP fail to simultaneously support task-agnostic inference and a bounded prompt pool. This motivates us to study an important yet underexplored setting for task-agnostic prompt usage:
% \begin{list}{}{\topsep=0.3ex \leftmargin=0.15in \rightmargin=0.in \itemsep =0.022in}
% \item {\em Task identifiers are unavailable at inference time for prompt selection, 
% % The task identity is unknown at inference time, 
% and the prompt pool has a fixed capacity. In this setting, the model cannot (1) guarantee zero forgetting by relying on task-specific prompts, nor (2) indefinitely store all prompts as the task sequence grows. Consequently, the model must perform prediction using the entire available prompt pool, while prompt pruning or merging becomes necessary to maintain the bounded size of the pool.} 
% \end{list}

\begin{settingbox}
% \noindent\textbf{Task-Agnostic Bounded-Memory Setting.}
\emph{
Task identifiers are unavailable at inference time for prompt selection, and the prompt pool has a fixed capacity. The model therefore cannot rely on task-specific prompt routing or indefinitely store all prompts as the task sequence grows. Instead, it performs prediction using the available prompt pool, while pruning or merging prompts to maintain bounded memory.
}
\end{settingbox}
This paper focuses on this setting and makes the following contributions.
\begin{list}{$\bullet$}{\topsep=0.3ex \leftmargin=0.15in \rightmargin=0.in \itemsep =0.022in}

\item We observe that existing prompt-based continual learning approaches often struggle 
under task-agnostic prompt usage scenarios.
Our experiments reveal that when 
task identities are unavailable at inference time for prompt selection, 
performance on earlier tasks degrades substantially after training on new ones, with models frequently producing incorrect or ambiguous outputs. Without explicit task cues, the model may generate labels from unrelated tasks encountered during pretraining or earlier learning, or hallucinate entirely invalid outputs not belonging to any seen task's label set.
% This issue is especially pronounced for encoder--decoder architectures, which generate labels as free text. 
% Without explicit task cues, the model may generate labels from unrelated tasks encountered during pretraining or earlier learning. 

% We refer to this phenomenon as \emph{latent forgetting}, denoting the degradation in performance on earlier tasks 
% under task-agnostic prompt usage evaluation. 
% \kj{maybe slightly mention about hallucination and I feel we need not to mention this 'latent forgetting' concept but just use plain words to explain.}
% Similar observations have been reported in prior work (e.g., \cite{guo2024q}), showing that forward transfer often persists while backward transfer suffers severely in this setting.

\item We introduce \textbf{GRID} (Figure \ref{fig:main}), a bounded-memory framework for task-agnostic prompt-based continual learning that addresses the above limitations through two key components: (i) a decoding mechanism that leverages representative inputs and output-space-aware decoding to control the output space of pre-trained LLMs, thereby ensuring label consistency and improving backward transfer (BWT) without relying on task IDs for prompt selection; and (ii) a gradient-guided prompt selection strategy that dynamically evaluates prompt usefulness and merges less informative prompts, significantly reducing memory usage while maintaining both forward and backward transfer performance.
\end{list}

\begin{table}[h]
\centering
\scriptsize
\begin{tabular}{l|c|c}
\toprule
\textbf{Method} & \textbf{Task-Agnostic Inference} & \textbf{Bounded Prompt Memory} \\
\midrule
ProgPrompt & \ding{55} & \ding{55} \\
SHLPT      & \ding{55} & \ding{55} \\
Q-Tuning   & \ding{55} & Partial  \\
\textbf{GRID} & \ding{51} & \ding{51} \\
\bottomrule
   \end{tabular}
\caption{Comparison of prompt-based CL methods under two practical deployment constraints. Unlike prior methods that require task identifiers at inference or allow unbounded prompt growth, GRID satisfies both constraints jointly.}
\label{tab:method_comparison}
\end{table}

\vspace{-0.4cm}

\section{Related Work}
\label{sec:rel_work}
% \vspace{-0.2cm}

\subsection{Continual Learning}  Continual learning (CL) involves learning from a sequence of tasks without full access to previous task data, aiming to preserve prior knowledge and enable positive transfer. A key challenge is catastrophic forgetting \citep{mccloskey1989catastrophic}, where updates to model parameters on new data erode earlier knowledge. CL strategies are categorized into three approaches: memory-based methods store and replay past task data to mitigate forgetting \citep{shin2017continual, bang2021rainbow}; regularization-based methods penalize deviations from important parameters to retain knowledge without stored data \citep{kirkpatrick2017overcoming, zenke2017continual}. \cite{du-etal-2024-unlocking} adopts a gradient-masking strategy by updating only high-activation model parameters, achieving task-agnostic and rehearsal-free CL; and architecture-based methods expand the model by adding new components for each task \citep{ yoon2018lifelong, tiwari2026turning}. These methods face scalability issues in large pre-trained models, motivating the development of parameter-efficient CL techniques using lightweight components like prompts and adapters \citep{xu2023parameterefficientfinetuningmethodspretrained, ruckle-etal-2021-adapterdrop}. 

% \vspace{0.2cm}

\subsection{Continual Prompt Tuning} 
Prompt tuning \citep{lester-etal-2021-power, li2021prefix, gu2021ppt, wang2023multitask} adapts large language models (LLMs) by learning a small set of continuous vectors, or soft prompts, prepended to the input tokens. Unlike full finetuning, it updates only the prompts while freezing model parameters, achieving competitive or superior performance with lower computational and memory cost. Continual prompt tuning (CPT) \citep{zhu-etal-2022-continual, yin-etal-2022-contintin, ermis2022memory, wang2022dualprompt} extends this idea to the continual learning (CL) setting, where models adapt to evolving task sequences. A substantial body of work has focused on improving CPT's ability to retain and transfer knowledge, using techniques such as prompt concatenation \citep{razdaibiedina2023progressive}, parameter sharing \citep{wang2022learning}, and weighted prompt summation \citep{jiang2023towards}. However, existing CPT methods often depend on memory buffers 
\citep{zhu-etal-2022-continual, ermis2022memory}, assume task-aware 
prompt selection that can break down under task-agnostic inference 
\citep{guo2024q}, or maintain expanding prompt pools that become less 
scalable as the task sequence grows 
\citep{razdaibiedina2023progressive, wang2022dualprompt, smith2023coda}.

% However, existing approaches often rely on large memory buffers to mitigate forgetting \citep{zhu-etal-2022-continual, ermis2022memory}, impractical in privacy- or resource-constrained scenarios. Moreover, task-specific prompts suffer from latent forgetting under task-agnostic inference \citep{guo2024q}, while methods that expand prompt pools over tasks \citep{razdaibiedina2023progressive, wang2022dualprompt, smith2023coda} face scalability and efficiency bottlenecks as tasks accumulate.

% \vspace{-0.7cm}

\subsection{Gradient-Based Data Selection} Gradient-based strategies have been extensively studied in data selection and coreset construction \citep{aljundi2019gradient}, where the goal is to identify a representative or influential subset of training data \citep{borovicka2012selecting, rolf2021representation}. Methods such as Coreset Selection \citep{killamsetty2021gradmatchgradientmatchingbased, mirzasoleiman2020coresetsdataefficienttrainingmachine,hao2023bilevel} and Gradient Matching \citep{aljundi2019gradient} leverage the similarity or norm of training gradients to preserve performance while reducing dataset sizes. 
% Inspired by these ideas, recent works have begun to explore gradient-based criteria for parameter-efficient tuning. 
In the context of prompt learning, some recent works have begun to reduce the prompt memory. For example, 
Q-Tuning \citep{guo2024q} uses a PCA-based eviction and L2P \citep{wang2022learning} keeps updating the same set of prompts over time. 

% In contrast, 
% our work draws inspiration from the coreset literature and extends it to the prompt space by dynamically evaluating gradient norms of prompt embeddings, allowing us to merge less informative prompts and improve scalability under long task sequences.

\vspace{-0.2cm}

\section{Background and Challenges}
\vspace{-0.1cm}

\subsection{Problem Setup}
\label{subsec:formulation}
% \vspace{-0.2cm}

We consider a continual learning (CL) setting in which a model encounters a sequence of \(N\) tasks \(\mathcal{T} = \{T_1, T_2, \dots, T_N\}\), where each task \(T_i\) is associated with a labeled dataset \(D_i = \{(x_j, y_j)\}_{j=1}^{|D_i|}\). Here, \(x_j\) denotes an input instance and \(y_j \in \mathcal{Y}_i\) is the corresponding label. The model is built upon a pretrained encoder-decoder or decoder-only language model \(f(\cdot; \theta)\), whose parameters \(\theta\) are kept fixed throughout learning. Rather than finetuning \(\theta\), we adapt the model to each task \(T_i\) by learning a soft prompt \(\mathbf{p}_i \in \mathbb{R}^{l \times d}\), where \(l\) denotes the prompt length and \(d\) the embedding dimension. 
After observing tasks \(\{T_1, \dots, T_{t-1}\}\), we maintain a pool of learned prompts
$\mathcal{P} \subseteq \{\mathbf{p}_1, \dots, \mathbf{p}_{t-1}\}$, 
which serves as a memory of past task adaptations.

When a new task \(T_t\) arrives, we initialize a new prompt \(\mathbf{p}_t\) and train it using data from \(D_t\), concatenated with the existing prompt queue \(\mathcal{P}\). The backbone model \(f(\cdot; \theta)\) remains frozen during training; only the new prompt \(\mathbf{p}_t\) is updated. Let \(\mathcal{P}^{(t)} = \mathcal{P} \cup \{\mathbf{p}_t\}\) denote the prompt configuration used during training for task \(T_t\). The model prediction is then given by
$
\hat{y} = f(x; \mathcal{P}^{(t)}).$
% \begin{align}
%     \hat{y} = f(x; \mathcal{P}^{(t)}).
%     \label{eq:model_pred}
% \end{align}
The training objective is to minimize: $\mathcal{L}_t = \mathbb{E}_{(x,y)\sim D_t}  [\ell(f(x; \mathcal{P}^{(t)}), y) ]$, 
% \begin{equation}
%     \mathcal{L}_t = \mathbb{E}_{(x,y)\sim D_t} \left[ \ell(f(x; \mathcal{P}^{(t)}), y) \right]
%     \label{eq:training_loss}
% \end{equation}
where \(\ell(\cdot)\) is the token-level cross-entropy loss.

\subsection{Task-Agnostic Inference}

% {\bf Inference Settings.}
% We distinguish between two inference modes commonly studied in continual learning: 1) \textbf{Task-Aware Inference:} The task identity is known at inference time. The model retrieves the corresponding prompt \(\mathbf{p}_i\) from \(\mathcal{P}\) for prediction on task \(T_i\). 2) \textbf{Task-Agnostic Inference:} The task identity is not known. The model cannot rely on selecting a task-specific prompt. Instead, the model performs prediction using the entire available prompt queue \(\mathcal{P} = \{\mathbf{p}_1, \dots, \mathbf{p}_{t-1}\}\), which may optionally include an aggregated prompt from earlier filtering stages. The concatenated prompt queue is prepended to the input and passed to the model. Formally, the prediction is given by $\hat{y} = f(x; \mathcal{P})$. 
As motivated before, we focus on the following more realistic task-agnostic inference setting:

% \begin{enumerate}
% %     \item 
%     \textit{\bf 1) Task-Aware Inference:} The task identity is known at inference time. The model selects the corresponding prompt \(\mathbf{p}_i\) from the prompt pool \(\mathcal{P}\) to perform prediction on task \(T_i\).
    
    % \item 

\begin{definition}[Task-Agnostic Inference]
Task identifiers are unavailable at inference time for prompt selection.
As a result, the model cannot rely on task-specific prompt selection.
Instead, it performs prediction using the entire available prompt pool 
$\mathcal{P} \subseteq \{\mathbf{p}_1, \dots, \mathbf{p}_{t-1}\}$, which may optionally include an aggregated prompt derived from previous filtering stages.
The concatenated prompt sequence is prepended to the input and passed to the model.
The prediction is defined as:
$\hat{y} = f(x; \mathcal{P})$.
\end{definition}

 % \textcolor{blue}{Consistent with standard continual learning evaluation protocols \cite{lopez2017gradient}, we assume that the dataset-specific output space is known at evaluation time, while prompt--task correspondence remains unavailable.}

% \end{enumerate}
% \begin{equation}
%     \hat{y} = f(x; \mathcal{P})
%     \label{eq:inference_pred}
% \end{equation}
The task-agnostic evaluation objective  becomes: $
    \mathcal{L}_{\text{TA}} = \sum_{T_k \in \mathcal{T}_{\text{past}}} \mathbb{E}_{(x,y)\sim D_k} \left[ \ell(f(x; \mathcal{P}), y) \right].$
This formulation evaluates how well the concatenated prompt pool enables generalization to earlier tasks without needing to retrieve or know their corresponding individual prompts.
This setting appears in two important CL scenarios:
1) \textit{Online learning}, where a new task arrives and the model must evaluate or detect alignment with previously seen tasks \cite{aljundi2019onlinecontinuallearningmaximally}.
2) \textit{Retroactive evaluation}, where a model trained on a task sequence is later evaluated on older tasks without access to their individual prompts \cite{Chaudhry_2018, de2019episodic}.

\vspace{0.12cm}

\noindent {\bf Challenges.} First, in task-agnostic settings, unconstrained generative decoding often causes \emph{label drift} and \emph{hallucination}, where models generate semantically related but incorrect or syntactically invalid labels or produce unseen labels as likelihood mass spreads over an expanding vocabulary (Table~\ref{tab:qual_grid}).
Second, prior methods such as ProgPrompt~\cite{razdaibiedina2023progressive} and SHLPT~\cite{wu-etal-2024-mitigate} rely on human-specified label mappings (e.g., $0 \rightarrow$ ``negative''), which vary across datasets and introduce semantic inconsistencies that hinder cross-task generalization (Appendix, Table~\ref{appendix:incons_label}). Third, existing approaches lack explicit task awareness, treating even semantically related tasks independently, leading to overlapping prompt representations and reduced transfer. Finally, under task-agnostic evaluation, prompt memory grows unbounded as predictions rely on the entire prompt pool, with no principled mechanism for pruning or merging redundant prompts~\cite{razdaibiedina2023progressive}. Together, these issues motivate the need for a structured and scalable framework for prompt-based continual learning under task-agnostic conditions.

% \vspace{-0.2cm}
% \subsection{Why Not Vision Class-Incremental Learning (CIL) ?}
% In this section, we clarify why vision-based CIL methods \cite{smith2023coda, mcdonnell2023ranpac, wu-etal-2002-boosting} do not directly address our setting. Vision CIL operates under a fundamentally different assumption: each task introduces new object classes into a single shared classifier head, and the goal is selecting the correct class among all seen so far \cite{bako2026addressing}. In this work, we follow structurally different setting in NLP, where tasks are heterogeneous with different input formats, label semantics, and output spaces (e.g., sentiment, entailment, QA) \cite{bako2026addressing, wu-etal-2024-mitigate}. Here the core challenge is producing the correct task behavior without a task identifier, a mismatch in problem structure that makes direct application of vision CIL methods non-trivial. Therefore, task-agnostic inference in our setting is not merely a class-selection problem; the model must infer the appropriate task behavior and output space without a task identifier.

\subsection{Why Not Vision Class-Incremental Learning (CIL)?}

In this section, we clarify why existing vision class-incremental learning (CIL) methods \cite{smith2023coda,mcdonnell2023ranpac,wu-etal-2002-boosting} do not directly address our setting. Vision CIL is built on a fundamentally different problem formulation: each task introduces new object classes into a shared classification space, and the model incrementally expands its ability to distinguish among all previously seen classes \cite{bako2026addressing}. As a result, the primary challenge is preserving discriminative decision boundaries over an enlarging label set under a unified classifier head.

In contrast, the NLP continual learning setting often involves heterogeneous tasks with substantially different objectives, input structures, label semantics, and output spaces, such as sentiment analysis, natural language inference, and question answering \cite{bako2026addressing,wu-etal-2024-mitigate}. Unlike vision CIL, these tasks cannot be naturally unified as class expansion within a single shared label space. Consequently, task-agnostic inference in our setting is not merely a class-selection problem. Instead, the model must implicitly infer the underlying task type and activate the appropriate task behavior and output space without access to a task identifier. This structural mismatch makes the direct application of vision CIL methods to NLP task-agnostic continual learning non-trivial.

% \subsection{Why Not Vision Class-Incremental Learning (CIL)?}

% We clarify why vision-based CIL methods~\cite{smith2023coda, mcdonnell2023ranpac, wu-etal-2002-boosting} do not directly address our setting. Vision CIL typically assumes a unified classification space: each new task introduces additional object classes, and the model predicts the correct class among all seen classes using a shared classifier head~\cite{bako2026addressing}. In contrast, we consider a structurally different NLP setting, where tasks have heterogeneous input formats, label semantics, and output spaces, such as sentiment analysis, entailment, and question answering~\cite{wang2018glue, wang2019superglue, razdaibiedina2023progressive, wu-etal-2024-mitigate}. Therefore, task-agnostic inference in our setting is not merely a class-selection problem; the model must infer the appropriate task behavior and output space without a task identifier. This mismatch makes direct application of vision CIL methods non-trivial.

\begin{table}[t]
\small
\centering
\resizebox{\columnwidth}{!}{%
\begin{tabular}{@{}
  C{0.8cm}
  L{5.5cm}
  C{1cm}
  C{1.8cm}
@{}}
\toprule
\rowcolor{gray!25}
\textbf{Task} & \textbf{Input Text} & \textbf{Label} & \textbf{ProgPrompt}\\
\midrule
BoolQ & Did you ever lecture at Harvard? & true & \textcolor{RoyalBlue}{true</s>}\\
BoolQ & The imperialist nation wanted to strangle the economy of its colony. & false & \textcolor{RoyalBlue}{<pad>Fal}\\
BoolQ & The neoclassical canon was rooted in traditional European aesthetics. & true & \textcolor{RoyalBlue}{<pad>Fal}\\
MNLI  & Some of the buildings around the city square … colonial period. & entailment & \textcolor{BrickRed}{<pad>True </s>}\\
MNLI  & The U.S. Army acceded … to keep U.S. forces in place. & entailment & \textcolor{BrickRed}{<pad>4.0}\\
\bottomrule
\end{tabular}%
}
\caption{\small Qualitative comparison under task-agnostic inference. 
\textcolor{RoyalBlue}{Blue} indicates \textit{label drift}, where Progressive Prompts predict syntactically incorrect labels.
\textcolor{BrickRed}{Red} highlights \textit{hallucination}, where the model generates invalid outputs not part of the task’s label set.} 
\label{tab:qual_grid}
% \vspace{-0.8cm}
\end{table}

\begin{figure*}[htbp]
    \centering
\includegraphics[width=1\textwidth]{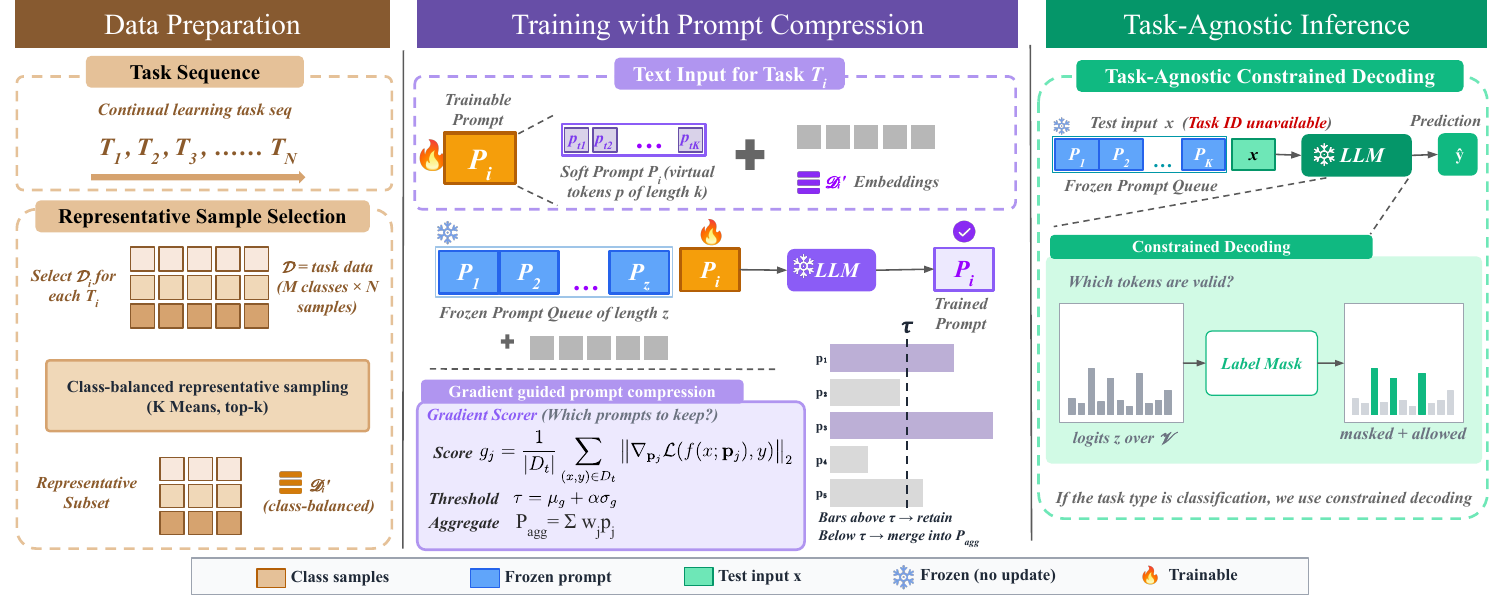}    

% \includesvg[width=0.95\textwidth]{images/main/main_2.svg}

    \caption{\small Overview of GRID. Given a continual task sequence, GRID first selects representative samples for each task, then trains a task-specific soft prompt while keeping previously learned prompts frozen. After each task, gradient-guided prompt compression scores the frozen prompt queue and retains or merges prompts based on their contribution to the current task. At inference time, GRID uses the compressed prompt queue without requiring a task identifier, and optionally applies constrained decoding for finite-output tasks. 
    % \kj{re-design and simplify; can use AI for help; too large}
    % Together, GRID supports task-agnostic inference while reducing forgetting and memory overhead.
    }
 % \kj{discribe this image a little bit in this caption.}
    \label{fig:main}
\end{figure*}

\begin{minipage}[ht]{0.45\textwidth}
\tiny
\begin{algorithm}[H]
\scriptsize
\caption{Prompt Selection and Aggregation}
\label{alg:prompt_selection}
\begin{algorithmic}[1]
\Require Prompt pool $\mathcal{P}=\{\mathbf{p}_j\}$, new task data $D_t$, thresholds $\alpha, \beta$
\Ensure Updated prompt pool $\mathcal{P}'$
\For{each $\mathbf{p}_j \in \mathcal{P}$}
    \State Compute gradient norm $g_j$ via Eq.~\ref{eq:g_j}
\EndFor
\State Compute $\mu_g$, $\sigma_g$; threshold $\tau$ via Eq.~\ref{eq:tau}
\State Classify prompts into $\mathcal{P}_{high}$ and $\mathcal{P}_{low}$ via Eq.~\ref{eq:phigh}
\If{$\mathcal{P}_{low} \neq \emptyset$}
    \State Compute aggregation $\mathbf{p}_{\text{agg}}$ via Eq.~\ref{eq:pagg}
    \State $\mathcal{P}' \gets \mathcal{P}_{high} \cup \{\mathbf{p}_{\text{agg}}\}$
\Else
    \State $\mathcal{P}' \gets \mathcal{P}_{high}$
\EndIf
\State Append new prompt $\mathbf{p}_t$ to $\mathcal{P}'$
\State \Return $\mathcal{P}'$
\end{algorithmic}
\end{algorithm}
\end{minipage}
% \vspace{-0.3cm}
\section{Method}
\label{sec:method}
% \vspace{-0.1cm}

To address the aforementioned challenges,  
we propose a unified framework called \textbf{GRID}: \textbf{G}radient-based prompt selection with \textbf{R}epresentative sample selection, label \textbf{I}dentification, and constrained \textbf{D}ecoding. GRID integrates two complementary components: 1) An input pipeline that enhances backward retention and output consistency by selecting representative samples, performing task identification, and applying constrained decoding; 2) A gradient-based prompt scoring mechanism that reduces prompt pool size by identifying and merging less informative prompts while preserving relevant task knowledge. 
% Figure~\ref{fig:main} for the pipeline and Algorithm \ref{alg:prompt_selection} for the Prompt Selection and Aggregation algorithm. Other algorithm details in Appendix ~\ref{append:alg}.
Figure~\ref{fig:main} illustrates the overall GRID pipeline, while Algorithm~\ref{alg:prompt_selection} details the Prompt Selection and Aggregation procedure. Additional details are provided in Appendix~\ref{append:alg}.

% ===============================================================
% A. Representative Samples, Task Type, Constrained Decoding 
% ===============================================================
% \vspace{-0.2cm}
\subsection{Input Construction for Stable Decoding}
% \vspace{-0.2cm}

In task-agnostic inference, unconstrained decoding often leads to \emph{label drift}, where predictions correspond to unrelated tasks or pretraining artifacts. To address this, we reformulate task-agnostic inference over a refined input space built from three components:

\textbf{1) Representative Input Sampling.}  
To construct a compact and informative training subset, we select \(k\) representative samples per class via clustering, improving upon the random sampling strategy used in prior work \citep{razdaibiedina2023progressive}. The data set \(\mathcal{D} = \{(x_j, y_j)\}\) is partitioned by label \(y \in \mathcal{Y}\), and the sentences \(\mathbf{e}_j = f_{\text{embed}}(x_j)\) are calculated using a pre-trained embedding model (\texttt{all-MiniLM-L6-v2}; \citep{reimers-2019-sentence-bert}). K-Means clustering is performed within each class to ensure intra-class diversity, and the top \(k/C\) samples closest to each cluster center are selected based on cosine similarity:
% \begin{equation*}
$\text{sim}(\mathbf{e}, \mathbf{c}) = \frac{\mathbf{e} \cdot \mathbf{c}}{\|\mathbf{e}\| \|\mathbf{c}\|}. $
% \end{equation*}
This yields a balanced subset \(\mathcal{D}_{\text{rep}}\) that spans the semantic space of each class. 

% We empirically find that increasing beyond 1k samples per class, the accuracy either saturates or improves negligibly, indicating that larger sample sizes may not result in substantial gains in performance (See Appendix~\ref{Appendix} Table~\ref{samples}), while clustering improves both data efficiency and generalization.

% \textbf{2) Label Identification.}  
% In many datasets, task categories are not explicitly provided, and labels often appear in non-descriptive formats (e.g., \{0,1\} or \{choice1, choice2\}) that are ambiguous without task context. To resolve this, we implement a hierarchical task identification module that infers the task type \(t^*\) given a candidate label set \(\mathcal{Y}_i\) and a sample input \(x\). The process begins with \textit{rule-based heuristics}, which match label tokens and input structure to predefined task templates (e.g., ``positive/negative'' for sentiment analysis, or premise-hypothesis pairs with ``entailment/contradiction'' for NLI). If this fails, we fallback to \textit{zero-shot task classification} using a lightweight generative language model (e.g., \href{https://huggingface.co/microsoft/Phi-3.5-mini-instruct}{Phi-3.5}). The model is prompted to infer the task type and remap non-descriptive labels into meaningful textual tokens:
% % \begin{equation*}
% $(t^*, \tilde{\mathcal{Y}}_i) = \textbf{LLM}(x, \mathcal{Y}_i).$
% % \end{equation*}
% This remapping enables consistent semantic interpretation across tasks, making it possible to apply decoding constraints in the task-agnostic setting.

\textbf{2) Label Identification.}
Many datasets lack explicit task metadata and use non-descriptive labels (e.g., \{0,1\}, \{choice1, choice2\}), which are ambiguous in task-agnostic inference. 
We address this with a hierarchical label identification module that infers the task type $t^*$ from an input $x$ and candidate label set $\mathcal{Y}_i$. 
First, rule-based heuristics match label tokens and input structure to known task templates (e.g., sentiment or NLI). 
If unsuccessful, we apply zero-shot task classification using a lightweight generative model (e.g., Phi-3.5) to infer the task and semantically remap labels:
$(t^*, \tilde{\mathcal{Y}}_i) = \textbf{LLM}(x, \mathcal{Y}_i)$.
This normalization enables consistent label semantics and supports constrained decoding in the task-agnostic setting.

\textbf{3) Constrained Decoding.}  
With remapped label set \(\mathcal{L}_i = \{\ell_1, \dots, \ell_K\}\), we apply constrained decoding at inference. At each decoding step \(t\), we restrict the softmax to only allow tokens from \(\mathcal{L}_i\):
\begin{align*}
\tilde{P}(y_t \mid y_{<t}, x) =& \text{softmax}(M \odot \mathbf{z}_t)\\  M_j =& \mathbf{1}[v_j \in \mathcal{L}_i],
\end{align*}
where \(\mathbf{z}_t\) are the raw logits and \(M\) is a binary mask over the vocabulary \(\mathcal{V}\).

% ===============================================================
% B. Prompt Pool Compression via Gradient-Based Selection 
% ===============================================================
% \vspace{-0.2cm}
\subsection{Prompt Pool Compression via Gradient-Guided Selection}
\begin{figure}[t]
\centering
\includegraphics[width=0.45\textwidth]{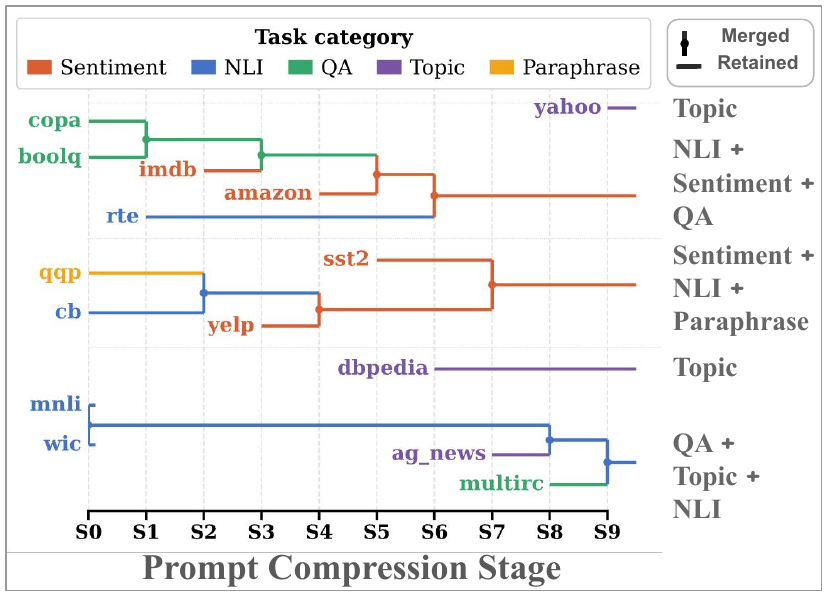}
% \vspace{-0.4cm}
\caption{\small
Prompt pool compression trajectory under GRID for Order L1.
Each leaf represents a task-specific prompt, with colors denoting task categories.
Branches indicate prompt consolidation events during compression, while solo branches denote prompts retained independently.
The trajectory shows that GRID forms interpretable compressed groups across related task types while preserving isolated prompts when merging is less suitable.
}
 % \kj{discribe this image a little bit in this caption.}
\label{fig:compression}
\end{figure}

Prompt-based CL methods typically allocate one prompt per task, causing the prompt pool to grow as $\mathcal{O}(N)$. To achieve scalable long-horizon continual learning, we propose a dynamic selection--compression strategy based on gradient relevance. Let $\mathcal{P} = \{\mathbf{p}_1, \dots, \mathbf{p}_{t-1}\}$ be the prompt pool before task $T_t$. For each $\mathbf{p}_j \in \mathcal{P}$, we compute the average gradient norm  over new task data $D_t$:
\begin{align}\label{eq:g_j}
    g_j = \tfrac{1}{|D_t|} \sum_{(x,y)\in D_t} \|\nabla_{\mathbf{p}_j} L_{t}(f(x; \mathbf{p}_j), y)\|_2. 
\end{align}
A large $g_j$ indicates that the task substantially updates $\mathbf{p}_j$, suggesting distinct knowledge worth preserving. Conversely, small $g_j$ values imply redundancy with the current task. Prompts are partitioned using the threshold
\begin{align}\label{eq:tau}
\tau = \mu_g + \alpha \sigma_g,
\end{align}
where $\mu_g, \sigma_g$ are the mean and standard deviation of $\{g_j\}$, and $\alpha$ is a tunable hyperparameter; for sensitivity analysis see Appendix \S\ref{apdx:alpha}.
\begin{align}\label{eq:phigh}
\mathcal{P}_{high} =& \{\mathbf{p}_j : g_j > \tau\}, \nonumber\\
\mathcal{P}_{low} =& \{\mathbf{p}_j : g_j < \tau\}. 
\end{align}
Although ~\eqref{eq:g_j} only measures prompt-task interaction, we observed that low-gradient prompts are often highly redundant (average cosine similarity $\geq 0.87$, Euclidean radius $R < 0.45$). When similarity is high, discarding them has little effect; but when $\mathcal{P}_{low}$ is diverse, removal risks losing transferable information. To mitigate this, we aggregate $\mathcal{P}_{low}$ into a single embedding $\mathbf{p}_{agg}$ using gradient-weighted averaging:
\begin{align}\label{eq:pagg}
\mathbf{p}_{agg} = \sum_{\mathbf{p}_j \in \mathcal{P}_{low}} w_j \mathbf{p}_j, 
\,
w_j = \frac{g_j}{\sum_{\mathbf{p}_k \in \mathcal{P}_{low}} g_k}.
\end{align}
This ensures that relatively more informative low-gradient prompts contribute proportionally. The updated pool for task $T_t$ becomes 
$\mathcal{P}' = \mathcal{P}_{high} \cup \{\mathbf{p}_{agg}\}.$  
This mechanism re-evaluates the existing prompts, retains high-gradient prompts, and merges low-gradient redundant ones into a single aggregated prompt which preserves critical knowledge while substantially reducing memory and inference costs. \emph{Notably, experiments show minimal degradation under compression, confirming that low-gradient prompts contribute little to future tasks.} Figure~\ref{fig:compression} illustrates this compression process for a task sequence.

\section{Experiments}
\label{sec:exp}

% \textbf{Datasets.} Following \citet{razdaibiedina2023progressive}, we evaluate in few-shot continual-learning ``long-sequence'' settings with six 15-task sequences (order L1-L6). L1-L3 are taken from prior work, while L4-L6 are newly constructed to study task difficulty progressions (easy$\rightarrow$hard, hard$\rightarrow$easy, and mixed). In addition, we introduce a \emph{Negative Transfer} benchmark (order NT1--NT3), each containing 9 tasks with deliberately dissimilar transitions to induce transfer degradation. Detailed construction details are provided in the appendix~\ref{sec:order_create}.

% \textcolor{blue}{We evaluate on both \textbf{encoder--decoder} and \textbf{decoder-only} language models to demonstrate its generality across architectures. Following prior continual learning work \citep{wu-etal-2024-mitigate, qin2021lfpt5, zhu-etal-2022-continual, lester-etal-2021-power, wang2023multitask, liang2023prompts}, we use T5 and Flan-T5 models, covering sizes from T5-small (60M) to T5-large (770M), and additionally T5-3B to study scalability. To extend our evaluation beyond encoder--decoder models, we further include decoder-only models, namely \texttt{Qwen/Qwen3-4B-Instruct} and \texttt{meta-llama/Llama-2-7b-hf}, covering both instruction-tuned and base pretrained autoregressive LLMs. This expanded model suite allows us to assess the effectiveness of GRID across different decoding paradigms.}

We evaluate on six long-sequence benchmarks (L1–L6, 15 tasks each), where L1–L3 follow prior work \citet{razdaibiedina2023progressive}, L4–L6 are newly constructed to study task difficulty progressions (easy$\rightarrow$hard, hard$\rightarrow$easy, and mixed), and NT1–NT3 form a Negative Transfer benchmark of nine deliberately dissimilar tasks. To evaluate GRID beyond standard classification-style benchmarks, we introduce \textbf{MTCL-Bench}, a 15-task Math-Tool continual learning benchmark combining Competition MATH, GSM8K~\cite{hendrycksmath2021, cobbe2021gsm8k}, and seven category-level ToolAlpaca tasks~\cite{tang2023toolalpaca}. Construction details are in Appendix \S\ref{sec:order_create}.

\begin{table}[ht]
% \vspace{-0.4cm}
\centering
\scriptsize

\setlength\tabcolsep{4pt}
\begin{tabular}{@{}l|ccccccc|c@{}}
\toprule
\rowcolor{gray!25}
\multirow{1}{*}{\textbf{Method}} 
  & \textbf{L1} & \textbf{L2} & \textbf{L3} & \textbf{L4} & \textbf{L5} & \textbf{L6} 
  & \textbf{DR} & \textbf{Avg} \\
\midrule
Finetune              & 8.3 & 8.7 & 7.8 & 7.9 & 8.1 & 8.9 & \ding{55} & 8.3 \\
Prompt Tuning         & 8.8 & 9.5 & 8.1 & 9.3 & 9.4 & 9.4 & \ding{55} & 9.1 \\
Data Replay           & 56.2 & 54.3 & 53.5 & 54.8 & 54.2 & 55.3 & \ding{51} & 54.7 \\ 
LFPT5                 & 70.8 & 69.2 & 69.4 & 68.2 & 69.4 & 68.5 & \ding{51} & 69.3 \\
Per-task Prompt       & 75.0 & 75.6 & 76.2 & 74.8 & 75.9 & 73.6 & \ding{55} & 75.2 \\    
ProgPrompt            & 75.7 & 78.6 & 74.3 & 75.05 & 77.10 & 75.46 & \ding{55} & 76.0 \\
SHLPT                 & 77.4 & 77.9 & 78.8 & 78.4 & 78.2 & \textbf{76.2} & \ding{55} & 77.8 \\ 
\midrule
\rowcolor{oursblue}
GRID\textsuperscript{*} & \textbf{79.1} & \textbf{80.7} & \textbf{81.1} & \textbf{79.0} & \textbf{79.8} & 75.5 & \ding{55} & \textbf{79.2} \\
\bottomrule
\end{tabular}
\caption{\small Average test-set accuracy on long-sequence order L1–L6 with T5-large. 
The \textbf{DR} column indicates whether the method uses data replay (\ding{51}) or not (\ding{55}).}
\label{Table2}
% \vspace{-0.2cm}
\end{table}

\vspace{-0.3cm}
\section{Results and Discussion}
\vspace{-0.1cm}
\begin{table*}[t]
\centering
% \scriptsize
\setlength\tabcolsep{4.2pt}
\renewcommand{\arraystretch}{1.18}

\resizebox{0.92\textwidth}{!}{
\begin{tabular}{c l  ccc  ccc  ccc  ccc}
\toprule
\multirow{2}{*}{\textbf{Order}} & \multirow{2}{*}{\textbf{Method}}
& \multicolumn{3}{c}{\textbf{T5-Large}}
& \multicolumn{3}{c}{\textbf{T5-3B}}
& \multicolumn{3}{c}{\textbf{Qwen-3-4B}}
& \multicolumn{3}{c}{\textbf{LLaMA-2-7B}} \\
\cmidrule(lr){3-5} \cmidrule(lr){6-8} \cmidrule(lr){9-11} \cmidrule(lr){12-14}
& & \textbf{Acc$\uparrow$} & \textbf{BWT$\uparrow$} & \textbf{FTC$\downarrow$}
  & \textbf{Acc$\uparrow$} & \textbf{BWT$\uparrow$} & \textbf{FTC$\downarrow$}
  & \textbf{Acc$\uparrow$} & \textbf{BWT$\uparrow$} & \textbf{FTC$\downarrow$}
  & \textbf{Acc$\uparrow$} & \textbf{BWT$\uparrow$} & \textbf{FTC$\downarrow$} \\
\midrule

% ---------- L1 ----------
\multirow{3}{*}{\textbf{L1}}
& ProgPrompt
& 75.70 & -0.7275 & 77
& 76.10 & -0.8475 & 80
& 79.70 & -0.8987 & 82
& 78.98 & -0.8575 & 80 \\
& SHLPT
& \second{77.40} & \second{-0.6123} & \second{64}
& \second{78.10} & \second{-0.6954} & \second{70}
& \second{80.95} & \second{-0.7265} & \second{75}
& \second{79.40} & \second{-0.6923} & \second{73} \\
\rowcolor{oursblue}

& \textbf{GRID}
& \best{79.10} & \best{-0.3243} & \best{11}
& \best{80.13} & \best{-0.3832} & \best{14}
& \best{83.54} & \best{-0.4243} & \best{18}
& \best{81.98} & \best{-0.3844} & \best{17} \\
\addlinespace[1.5pt]
\midrule

% ---------- L2 ----------
\multirow{3}{*}{\textbf{L2}}
& ProgPrompt
& \second{78.60} & -0.7625 & 72
& \second{79.56} & -0.8054 & 78
& \best{83.56} & -0.8421 & 81
& 81.87 & -0.8121 & 78 \\
& SHLPT
& 77.90 & \second{-0.6870} & \second{58}
& 79.33 & \second{-0.7458} & \second{65}
& \second{82.68} & \second{-0.7970} & \second{69}
& \second{82.12} & \second{-0.7645} & \second{70} \\
\rowcolor{oursblue}

& \textbf{GRID}
& \best{80.70} & \best{-0.3336} & \best{5}
& \best{81.40} & \best{-0.3726} & \best{9}
& 82.32 & \best{-0.4198} & \best{15}
& \best{83.12} & \best{-0.4387} & \best{11} \\
\addlinespace[1.5pt]
\midrule

% ---------- L3 ----------
\multirow{3}{*}{\textbf{L3}}
& ProgPrompt
& 74.30 & -0.6137 & 80
& 76.76 & -0.6932 & 85
& 78.32 & -0.6934 & 89
& 77.33 & -0.6932 & 87 \\
& SHLPT
& \second{78.80} & \second{-0.5174} & \second{69}
& \second{79.32} & \second{-0.5964} & \second{74}
& \second{80.54} & \second{-0.6344} & \second{78}
& \second{80.65} & \second{-0.6345} & \second{75} \\
\rowcolor{oursblue}

& \textbf{GRID}
& \best{81.10} & \best{-0.3979} & \best{18}
& \best{81.65} & \best{-0.4279} & \best{19}
& \best{83.87} & \best{-0.4812} & \best{23}
& \best{84.76} & \best{-0.4954} & \best{22} \\
\addlinespace[1.5pt]

\midrule

% ---------- L4 ----------
\multirow{3}{*}{\textbf{L4}}
& ProgPrompt
& 75.05 & -0.6257 & 87
& 77.43 & -0.6657 & 89
& 81.54 & -0.6957 & 92
& 80.21 & -0.6832 & 89 \\
& SHLPT
& \second{78.40} & \second{-0.5042} & \second{73}
& \second{79.87} & \second{-0.5542} & \second{78}
& 81.42 & \second{-0.6021} & \second{85}
& \second{81.43} & \second{-0.5943} & \second{81} \\
\rowcolor{oursblue}

& \textbf{GRID}
& \best{79.00} & \best{-0.3912} & \best{26}
& \best{80.43} & \best{-0.4321} & \best{29}
& \best{82.43} & \best{-0.4842} & \best{34}
& \best{81.53} & \best{-0.4565} & \best{33} \\
\addlinespace[1.5pt]

\midrule

% ---------- L5 ----------
\multirow{3}{*}{\textbf{L5}}
& ProgPrompt
& 77.10 & -0.6351 & 71
& 78.43 & -0.6790 & 74
& 80.21 & -0.7254 & 79
& 80.43 & -0.7187 & 79 \\
& SHLPT
& \second{78.20} & \second{-0.4187} & \second{58}
& \second{79.43} & \second{-0.4954} & \second{63}
& \second{82.76} & \second{-0.5644} & \second{68}
& \second{81.64} & \second{-0.5698} & \second{75} \\
\rowcolor{oursblue}

& \textbf{GRID}
& \best{79.80} & \best{-0.2956} & \best{10}
& \best{79.43} & \best{-0.3576} & \best{15}
& \best{83.43} & \best{-0.3588} & \best{19}
& \best{82.54} & \best{-0.3654} & \best{20} \\
\addlinespace[1.5pt]

\midrule

% ---------- L6 ----------
\multirow{3}{*}{\textbf{L6}}
& ProgPrompt
& \second{75.46} & -0.6416 & 85
& 77.54 & -0.6754 & 89
& \second{81.64} & \second{-0.6926} & 92
& 79.65 & \second{-0.6837} & 88 \\
& SHLPT
& 76.20 & \second{-0.5840} & \second{62}
& \second{78.43} & \second{-0.6425} & \second{67}
& \best{82.54} & -0.7736 & \second{67}
& \second{81.65} & -0.7653 & \second{66} \\
\rowcolor{oursblue}

& \textbf{GRID}
& \best{75.50} & \best{-0.3512} & \best{13}
& \best{77.35} & \best{-0.3753} & \best{16}
& 81.26 & \best{-0.4132} & \best{18}
& \best{80.65} & \best{-0.4036} & \best{18} \\
\addlinespace[1.5pt]

\midrule

% ---------- Avg ----------
\multirow{3}{*}{\textbf{Avg}}
& ProgPrompt
& 76.00 & -0.6677 & 78.7
& 77.64 & -0.7277 & 82.5
& 80.83 & -0.7580 & 85.8
& 79.75 & -0.7414 & 83.5 \\
& SHLPT
& \second{77.80} & \second{-0.5539} & \second{64.0}
& \second{79.08} & \second{-0.6216} & \second{69.5}
& \second{81.82} & \second{-0.6830} & \second{73.7}
& \second{81.15} & \second{-0.6701} & \second{73.3} \\
\rowcolor{oursblue}

& \textbf{GRID}
& \best{79.20} & \best{-0.3490} & \best{13.8}
& \best{80.06} & \best{-0.3915} & \best{17.0}
& \best{82.81} & \best{-0.4303} & \best{21.2}
& \best{82.43} & \best{-0.4240} & \best{20.2} \\
\bottomrule
\end{tabular}}
\caption{\small Performance under task-agnostic prompt usage across task orders (L1--L6). We report only the strongest ProgPrompt and SHLPT
baselines and our GRID method  for brevity. 
We report Avg. Accuracy (Acc$\uparrow$; higher is better), Backward Transfer (BWT$\uparrow$; less negative is better),
and Forgotten Task Count (FTC$\downarrow$; lower is better) for each backbone. Best and second-best are \best{bold} and \second{underlined}.}
\label{tab:dec}
\end{table*}

\subsection{Average Accuracy across tasks}
% Our method excels in the long-sequence experiments (L1-L6); Table~\ref{Table2}), achieving an average accuracy of 79.2\% on T5-large and surpassing competitive baselines such as SHLPT, the most recent state-of-the-art method.
%  These results demonstrate that gradient-guided prompt compression effectively reduces redundancy while preserving task-relevant knowledge. Importantly, the gains extend beyond encoder--decoder models: \textcolor{blue}{GRID also yields consistent improvements on decoder-only backbones (Table~\ref{tab:dec}), including \texttt{Qwen-3-4B} and \texttt{LLaMA-2-7B}, indicating that the proposed framework generalizes across architectures and decoding paradigms.}
%  Under the Negative Transfer benchmarks (Order NT1–NT3; Table~\ref{Table6}), we also observe clear improvements (e.g., +3.8\% on FT5-base, +3.6\% on FT5-large), highlighting its robustness even in low-task-similarity scenarios.

GRID achieves strong performance in long-sequence settings (L1--L6; Table~\ref{Table2}, ~\ref{tab:dec}), reaching 79.2\% average accuracy on T5-large and outperforming competitive baselines, including SHLPT, the current state of the art. These gains indicate that gradient-guided prompt compression reduces redundancy while preserving task-relevant knowledge. Notably, the improvements extend to decoder-only models (Table~\ref{tab:dec}), with consistent gains on \texttt{Qwen-3-4B} and \texttt{LLaMA-2-7B}, demonstrating robustness across architectures and decoding paradigms. On the Negative Transfer benchmarks (NT1--NT3; Table~\ref{Table6}), GRID continues to outperform baselines (e.g., +3.8\% on FT5-base, +3.6\% on FT5-large), highlighting its effectiveness under low task similarity. On MTCL-Bench, GRID remains competitive (Table~\ref{tab:mtcl_bench_results}), improving average accuracy on LLaMA while maintaining stronger retention than ProgPrompt on both backbones.

\begin{figure}[t]
    \centering
\includegraphics[width=0.45\textwidth]{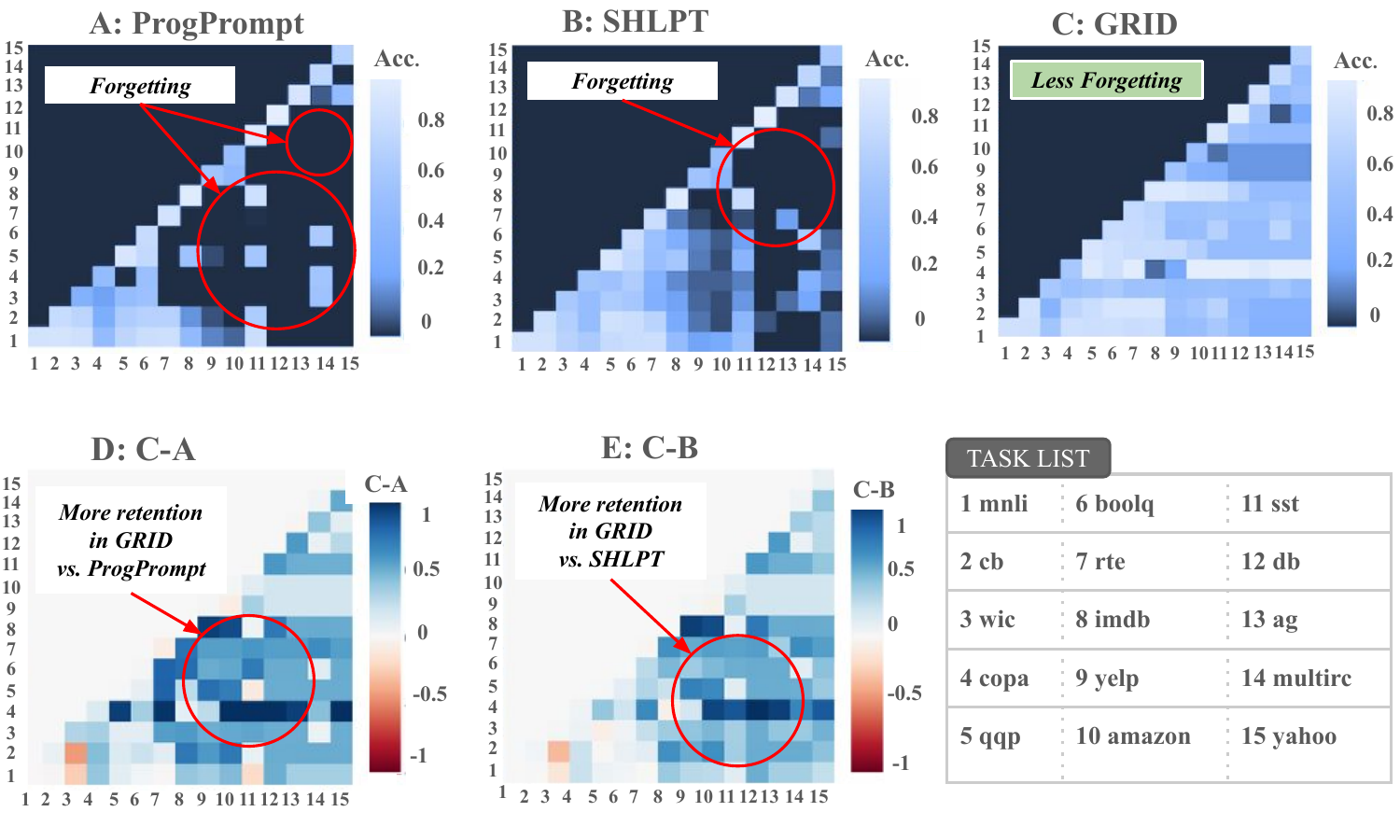}
    % \vspace{-0.4cm}

    \caption{\small Heatmaps of backward transfer scores on previous tasks for Order L1. 
(A) Progressive Prompts, (B) SHLPT, (C) GRID, and differences (D) C–A, (E) C–B.}
 % \kj{discribe this image a little bit in this caption.}
    \label{fig:O8_heat}
\end{figure}

% \begin{table}[htbp]
% \centering
% \scriptsize
% \setlength{\tabcolsep}{4pt}
% \renewcommand{\arraystretch}{1.08}
% \begin{tabular}{llcc}
% \toprule
% \textbf{Model} & \textbf{Method} 
% & \textbf{Avg. Acc.} & \textbf{BWT} \\
% \midrule

% \multirow{5}{*}{\shortstack[l]{LLaMA-3.1\\8B-Instruct}}
% & Pretrained & 45.90 & -- \\
% & Continual Prompt Tuning & 16.89 & -42.67 \\
% & ProgPrompt & 49.15 & -35.09 \\
% & \cellcolor{oursblue}GRID 
%   & \cellcolor{oursblue}\textbf{50.55} 
%   & \cellcolor{oursblue}\textbf{-25.34} \\
% & Per-task LoRA SFT & 51.53 & -- \\

% \midrule

% \multirow{5}{*}{\shortstack[l]{Qwen3-4B\\Instruct}}
% & Pretrained & 45.16 & -- \\
% & Continual Prompt Tuning & 14.12 & -7.34 \\
% & ProgPrompt & 54.85 & -6.12 \\
% & \cellcolor{oursblue}GRID 
%   & \cellcolor{oursblue}53.63 
%   & \cellcolor{oursblue}\textbf{-5.14} \\
% & Per-task LoRA SFT & \textbf{55.97} & -- \\

% \bottomrule
% \end{tabular}
% \caption{\small Results on MTCL-Bench. Avg. Acc. and BWT denote average accuracy and backward transfer, respectively. All values are reported as percentages. \kj{can be better}}
% \label{tab:mtcl_bench_results}
% \end{table}

\subsection{Backward Transfer Analysis}

Table~\ref{tab:dec} reports BWT across long-sequence settings and model variants. GRID consistently exhibits substantially less negative BWT than ProgPrompt and SHLPT, reducing forgetting by nearly half on average, with relative improvements exceeding 50\% across all seq-to-seq model variants and similar trends on decoder-only backbones. While larger models achieve higher accuracy (Table~\ref{tab:dec},~\ref{tab:enc-dec}), they suffer greater forgetting, whereas smaller models retain prior knowledge better, likely due to less aggressive adaptation. MTCL-Bench further confirms GRID's retention benefit: BWT improves from $-35.09$ to $-25.34$ on LLaMA and from $-6.12$ to $-5.14$ on Qwen (Table~\ref{tab:mtcl_bench_results}).

Figure~\ref{fig:O8_heat} visualizes BWT dynamics for Order L1, where GRID (C) shows brighter lower-triangle regions than ProgPrompt (A) and SHLPT (B), with difference heatmaps (D,E) highlighting widespread gains particularly for early tasks. Figure~\ref{fig:bwt_and_analysis} further shows consistently less negative BWT as tasks accumulate (a), and substantial per-task gains on challenging tasks such as \texttt{boolq}, \texttt{sst2}, and \texttt{dbpedia\_14} (b); Figure~\ref{fig:cluster} explains higher forgetting on outlier tasks like \texttt{wic} through their distinct semantic cluster. Additional comparisons across L1--L6 and task-order analyses are provided in Appendix \S\ref{sec:order_analysis} and Table~\ref{tab:bwt_comparison}.

\begin{table}[htbp]
\centering
\scriptsize
\setlength{\tabcolsep}{3.8pt}
\renewcommand{\arraystretch}{1.10}
\begin{tabular}{llccc}
\toprule
\rowcolor{gray!25}
\textbf{Model} & \textbf{Method} & \textbf{Is CL?} 
& \textbf{Acc$\uparrow$} & \textbf{BWT$\uparrow$} \\
\midrule

\multirow{6}{*}{\shortstack[l]{LLaMA-3.1\\8B-Instruct}}
& Pretrained & \ding{55} & 45.90 & -- \\
& Per-task LoRA SFT & \ding{55} & 51.53 & -- \\
\cmidrule(lr){2-5}
& Prompt Tuning & \ding{51} & 16.89 & -42.67 \\
& ProgPrompt & \ding{51} & \underline{49.15} & \underline{-35.09} \\
& \cellcolor{oursblue}GRID 
  & \cellcolor{oursblue}\ding{51} 
  & \cellcolor{oursblue}\textbf{50.55} 
  & \cellcolor{oursblue}\textbf{-25.34} \\

\midrule

\multirow{6}{*}{\shortstack[l]{Qwen3-4B\\Instruct}}
& Pretrained & \ding{55} & 45.16 & -- \\
& Per-task LoRA SFT & \ding{55} & 55.97 & -- \\
\cmidrule(lr){2-5}
& Prompt Tuning & \ding{51} & 14.12 & -7.34 \\
& ProgPrompt & \ding{51} & \textbf{54.85} & \underline{-6.12} \\
& \cellcolor{oursblue}GRID 
  & \cellcolor{oursblue}\ding{51} 
  & \cellcolor{oursblue}\underline{53.63} 
  & \cellcolor{oursblue}\textbf{-5.14} \\

\bottomrule
\end{tabular}
\caption{\footnotesize Results: MTCL-Bench. Best in \textbf{bold} and the second-best \underline{underlined}. All scores in percentages.}
\label{tab:mtcl_bench_results}
\end{table}

\vspace{-0.1cm}

\subsection{Forgotten Task Count}

% We define the Forgotten Task Count (FTC) as the number of tasks whose accuracy falls below a threshold relative to their standalone performance.  
% Formally, task $T_i$ is forgotten at step $t$ if $a_i^{(t)} < \tau \cdot \min_j a_j^{(0)}$, where $a_i^{(0)}$ is the accuracy of $T_i$ when trained alone and $\tau \in (0,1)$.  
% Using $0$ as a cutoff is misleading since accuracies may degrade to small but nonzero values; our formulation instead provides a principled absolute threshold.\footnotemark. GRID reduces forgetting dramatically compared to both ProgPrompt and SHLPT. On average, GRID forgets only 13.8 tasks, compared to 78.7 for ProgPrompt and 64.0 for SHLPT. These results highlight that GRID mitigates catastrophic forgetting in long-horizon continual learning, while preserving task-relevant knowledge in a compact form. For results across different model variants, please refer to Table~\ref{tab:forgotten_tasks_comparison} in the Appendix.

GRID substantially reduces catastrophic forgetting compared to ProgPrompt and SHLPT, forgetting only 13.8 tasks on average versus 78.7 and 64.0, respectively. This highlights GRID’s ability to preserve task-relevant knowledge over long task horizons while maintaining a compact prompt memory. (Details in the Appendix §\ref{apdx:ftc}.)

\subsection{Scalability to Larger Models}
\vspace{-0.2cm}

\begin{figure}[htbp]
\centering

\includegraphics[width=0.45\textwidth]{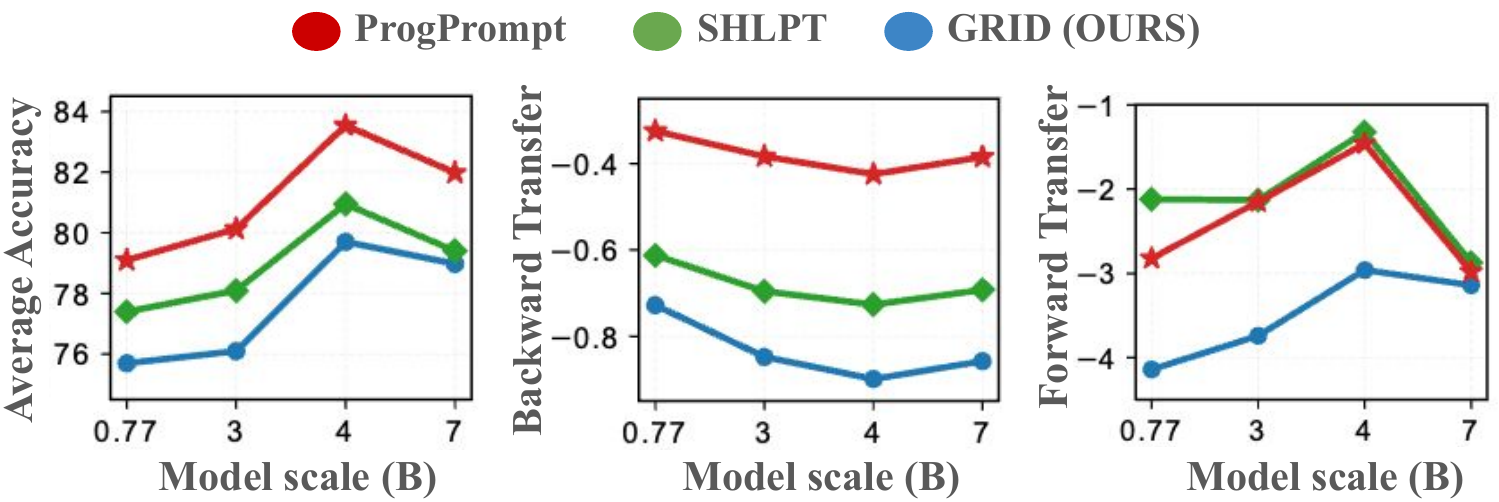}
% \vspace{-0.4cm}

\caption{\small Effect of model scaling across T5-Large (0.77B), T5-XL (3B), Qwen-3 (4B), and LLaMA-2 (7B).}

    \label{fig:model_scale}
\end{figure}

\vspace{-0.2cm}

Results on larger models (Table~\ref{tab:dec}) show that GRID preserves strong accuracy and reduced forgetting at scale. Figure~\ref{fig:model_scale} further demonstrates that this behavior remains stable as model size increases, indicating that GRID scales effectively to large language models.

% For $L1$, accuracy improves from 74.86 to 78.54, BWT improves from $-0.7161$ to $-0.4445$, indicating reduced forgetting, and forgotten tasks drop from 69 to 21, demonstrating much stronger retention of past knowledge. 
\footnotetext{Amazon had the lowest standalone accuracy ($\sim$0.50); setting $\tau=0.4$ yields an absolute threshold of 0.20.}

\vspace{-0.1cm}

\subsection{Training and inference overhead.}
GRID introduces a one-time gradient-scoring step per retained prompt when a new task arrives, incurring $O\!\left(\sum_{t=1}^{N} |P_t|\right)$ total overhead across $N$ tasks. Empirically, this adds $\sim$12.9--15.6 minutes per task (2.5--3.1 min/prompt $\times$ ${\sim}5$ prompts), totaling $\sim$144 minutes across Tasks 6--15. Overall wall-clock time remains comparable to ProgPrompt (27 vs.\ 24 h on A100; 12 vs.\ 11 h on H100). At inference, retaining ${\sim}5$ prompts instead of 15 yields a 5--20\% speedup for long inputs (${\sim}2048$ tokens) and 30--50\% for shorter inputs (${\sim}256$ tokens). Full details appear in Appendix \S\ref{apdx:overhead}.

% \vspace{-0.6cm}

\subsection{Ablation Study }
% \vspace{-0.4cm}

\begin{table}[t]
\centering
\tiny
\setlength\tabcolsep{6pt}
\begin{tabular}{
l
l
S[table-format = -1.4]
S[table-format = 3.0]
l
}
\toprule
\rowcolor{gray!25}
\textbf{Model} & \textbf{Variant} & {\textbf{Avg}} & {\textbf{Memory}} & {\textbf{GPU (h:m)}} \\
\midrule
\multirow{5}{*}{T5-large}
& \textbf{(0) G.R.I.D.}     & \bfseries -0.3511 & \bfseries 200 & \texttt{27:08} \\
& (1) w/o G                 & -0.3490 & 600 & \texttt{25:35} \\
& (2) w/o G,D               & -0.6863 & 600 & \texttt{26:42} \\
& (3) w/o G,D,R             & -0.6954 & 600 & \texttt{24:58} \\
& (4) w/o all               & -0.7012 & 600 & \texttt{23:58} \\
\midrule
\multirow{5}{*}{Qwen-3-4B}
& \textbf{(0) G.R.I.D.}     & \bfseries -0.4418 & \bfseries 200 & \texttt{84:12} \\
& (1) w/o G                 & -0.4393 & 600 & \texttt{81:38} \\
& (2) w/o G,D               & -0.7958 & 600 & \texttt{81:35} \\
& (3) w/o G,D,R             & -0.8009 & 600 & \texttt{80:46} \\
& (4) w/o all               & -0.8114 & 600 & \texttt{78:47} \\
\bottomrule
\end{tabular}
% \vspace{-0.4cm}
\caption{\small  Ablation study. We report average BWT (avg over L1-L3), prompt memory size in KB, and GPU time per run.}
\label{ablation:compact}
\end{table}

% \vspace{-0.2cm}
We conduct an ablation on task orders L1–L3, evaluating BWT across four GRID variants that remove gradient-based selection (G), constrained decoding (D), representative input selection (R), or all components (Table~\ref{ablation:compact}). The full model achieves the best BWT on both T5-large and Qwen-3-4B, with results indicating constrained decoding as the primary contributor to retention, while gradient-based selection mainly improves scalability (200KB vs.\ 600KB for ProgPrompt). Detailed ablation in Appendix \S\ref{sec:detailed_abalation}

\vspace{-0.2cm}

\section{Conclusion}

% We introduce \textbf{GRID}, a unified framework for scalable prompt-based continual learning under task-agnostic inference. By integrating constrained decoding with gradient-guided prompt selection and compression, GRID enables consistent label generation, compact prompt memory, and improved knowledge retention. Across settings, GRID substantially improves backward transfer by reducing forgotten tasks, while maintaining competitive forward transfer.

% In this work, we introduce \textbf{GRID}, a unified framework that makes prompt-based continual learning both \textit{scalable} and \textit{resilient to forgetting}. Unlike prior approaches that rely on task-aware inference or accumulate ever-growing prompt pools, GRID addresses two central challenges: \textit{latent forgetting under task-agnostic inference} and the \textit{inefficiency of unbounded prompt memory}. GRID combines \textit{constrained decoding} with \textit{gradient-guided prompt selection and compression}, enabling consistent label generation, compact memory usage, and improved long-horizon retention. Empirically, GRID substantially improves backward transfer by reducing forgotten tasks by a large margin, while maintaining competitive forward transfer.
\vspace{-0.2cm}
In this work, we introduce GRID, a unified framework that makes prompt-based continual learning both scalable and robust to forgetting. Unlike prior approaches that depend on task-aware inference or accumulate ever-growing prompt pools, GRID tackles two key challenges: \textit{forgetting under task-agnostic inference} and the \textit{inefficiency of unbounded prompt memory}. It integrates \textit{constrained decoding} with \textit{gradient-guided prompt selection and compression}, enabling consistent label generation, compact memory, and improved retention, while substantially enhancing backward transfer and preserving competitive forward transfer.

% Future work could extend GRID to also enable positive backward knowledge transfer, allowing new tasks to refine earlier prompts and further boost their performance. Additional directions include scaling to much longer task streams, applying the framework to other types of foundation models.  

\section*{Limitations}

While GRID demonstrates strong improvements in backward transfer and scalability, it has several primary limitations:

% \vspace{0.5em}
First, although inference does not require task identifiers, GRID assumes access to task boundaries during training, which may not hold in fully task-free or streaming settings. Extending GRID to operate without explicit task demarcations during training is an interesting direction for future work.

% \vspace{0.5em}
Second, our decoding mechanism uses task output spaces for label normalization and constrained prediction, which may limit applicability to fully open-ended generation; however, MTCL-Bench partly addresses this through math tasks that require generating the correct answer rather than selecting from fixed labels.

% \vspace{0.5em}
Third, to fully assess the
scalability of our method, future work should test it on significantly longer task streams.

\section*{Ethical Considerations}
GRID is a methodological contribution to prompt-based continual learning and introduces no user-facing applications. All experiments use publicly available benchmarks under their original licenses, and the fully structured MTCL-Bench constructed from publicly available datasets will be released upon acceptance to support reproducibility. GRID inherits biases and risks from the underlying pretrained language models, but we identify no additional ethical risks beyond those common to LLMs and continual learning systems.

\bibliography{anthology_0, anthology_1, custom}

\begin{thebibliography}{59}
\providecommand{\natexlab}[1]{#1}

\bibitem[{Aljundi et~al.(2019{\natexlab{a}})Aljundi, Caccia, Belilovsky, Caccia, Lin, Charlin, and Tuytelaars}]{aljundi2019onlinecontinuallearningmaximally}
Rahaf Aljundi, Lucas Caccia, Eugene Belilovsky, Massimo Caccia, Min Lin, Laurent Charlin, and Tinne Tuytelaars. 2019{\natexlab{a}}.
\newblock \href {https://arxiv.org/abs/1908.04742} {Online continual learning with maximally interfered retrieval}.
\newblock \emph{Preprint}, arXiv:1908.04742.

\bibitem[{Aljundi et~al.(2019{\natexlab{b}})Aljundi, Lin, Goujaud, and Bengio}]{aljundi2019gradient}
Rahaf Aljundi, Min Lin, Baptiste Goujaud, and Yoshua Bengio. 2019{\natexlab{b}}.
\newblock Gradient based sample selection for online continual learning.
\newblock \emph{Advances in neural information processing systems}, 32.

\bibitem[{Bako and Kalita(2026)}]{bako2026addressing}
John Bako and Jugal Kalita. 2026.
\newblock Addressing catastrophic forgetting in class-incremental learning—a survey.

\bibitem[{Bang et~al.(2021)Bang, Kim, Yoo, Ha, and Choi}]{bang2021rainbow}
Jihwan Bang, Heesu Kim, YoungJoon Yoo, Jung-Woo Ha, and Jonghyun Choi. 2021.
\newblock Rainbow memory: Continual learning with a memory of diverse samples.
\newblock In \emph{Proceedings of the IEEE/CVF conference on computer vision and pattern recognition}, pages 8218--8227.

\bibitem[{Borovicka et~al.(2012)Borovicka, Jirina~Jr, Kordik, and Jirina}]{borovicka2012selecting}
Tomas Borovicka, Marcel Jirina~Jr, Pavel Kordik, and Marcel Jirina. 2012.
\newblock Selecting representative data sets.
\newblock \emph{Advances in data mining knowledge discovery and applications}, 12:43--70.

\bibitem[{Chaudhry et~al.(2018)Chaudhry, Dokania, Ajanthan, and Torr}]{Chaudhry_2018}
Arslan Chaudhry, Puneet~K. Dokania, Thalaiyasingam Ajanthan, and Philip H.~S. Torr. 2018.
\newblock \href {https://doi.org/10.1007/978-3-030-01252-6_33} {\emph{Riemannian Walk for Incremental Learning: Understanding Forgetting and Intransigence}}, page 556–572.
\newblock Springer International Publishing.

\bibitem[{Cobbe et~al.(2021)Cobbe, Kosaraju, Bavarian, Chen, Jun, Kaiser, Plappert, Tworek, Hilton, Nakano, Hesse, and Schulman}]{cobbe2021gsm8k}
Karl Cobbe, Vineet Kosaraju, Mohammad Bavarian, Mark Chen, Heewoo Jun, Lukasz Kaiser, Matthias Plappert, Jerry Tworek, Jacob Hilton, Reiichiro Nakano, Christopher Hesse, and John Schulman. 2021.
\newblock Training verifiers to solve math word problems.
\newblock \emph{arXiv preprint arXiv:2110.14168}.

\bibitem[{de~Masson~D'Autume et~al.(2019)de~Masson~D'Autume, Ruder, Kong, and Yogatama}]{de2019episodic}
Cyprien de~Masson~D'Autume, Sebastian Ruder, Lingpeng Kong, and Dani Yogatama. 2019.
\newblock Episodic memory in lifelong language learning.
\newblock \emph{Advances in Neural Information Processing Systems}, 32.

\bibitem[{Ding et~al.(2023)Ding, Qin, Yang, Wei, Yang, Su, Hu, Chen, Chan, Chen et~al.}]{ding2023parameter}
Ning Ding, Yujia Qin, Guang Yang, Fuchao Wei, Zonghan Yang, Yusheng Su, Shengding Hu, Yulin Chen, Chi-Min Chan, Weize Chen, and 1 others. 2023.
\newblock Parameter-efficient fine-tuning of large-scale pre-trained language models.
\newblock \emph{Nature Machine Intelligence}, 5(3):220--235.

\bibitem[{Du et~al.(2024)Du, Cheng, Luo, Qiu, Huang, Cheung, Cheng, and Fu}]{du-etal-2024-unlocking}
Wenyu Du, Shuang Cheng, Tongxu Luo, Zihan Qiu, Zeyu Huang, Ka~Chun Cheung, Reynold Cheng, and Jie Fu. 2024.
\newblock \href {https://doi.org/10.18653/v1/2024.findings-emnlp.379} {Unlocking continual learning abilities in language models}.
\newblock In \emph{Findings of the Association for Computational Linguistics: EMNLP 2024}, pages 6503--6522, Miami, Florida, USA. Association for Computational Linguistics.

\bibitem[{Ermis et~al.(2022)Ermis, Zappella, Wistuba, Rawal, and Archambeau}]{ermis2022memory}
Beyza Ermis, Giovanni Zappella, Martin Wistuba, Aditya Rawal, and Cedric Archambeau. 2022.
\newblock Memory efficient continual learning with transformers.
\newblock \emph{Advances in Neural Information Processing Systems}, 35:10629--10642.

\bibitem[{Gu et~al.(2021)Gu, Han, Liu, and Huang}]{gu2021ppt}
Yuxian Gu, Xu~Han, Zhiyuan Liu, and Minlie Huang. 2021.
\newblock Ppt: Pre-trained prompt tuning for few-shot learning.
\newblock \emph{arXiv preprint arXiv:2109.04332}.

\bibitem[{Guo et~al.(2024)Guo, Xu, Fu, Liu, Dong, and Wang}]{guo2024q}
Yanhui Guo, Shaoyuan Xu, Jinmiao Fu, Jia Liu, Chaosheng Dong, and Bryan Wang. 2024.
\newblock Q-tuning: Queue-based prompt tuning for lifelong few-shot language learning.
\newblock \emph{arXiv preprint arXiv:2404.14607}.

\bibitem[{Hao et~al.(2023)Hao, Ji, and Liu}]{hao2023bilevel}
Jie Hao, Kaiyi Ji, and Mingrui Liu. 2023.
\newblock Bilevel coreset selection in continual learning: A new formulation and algorithm.
\newblock \emph{Advances in Neural Information Processing Systems}, 36:51026--51049.

\bibitem[{Hendrycks et~al.(2021)Hendrycks, Burns, Kadavath, Arora, Basart, Tang, Song, and Steinhardt}]{hendrycksmath2021}
Dan Hendrycks, Collin Burns, Saurav Kadavath, Akul Arora, Steven Basart, Eric Tang, Dawn Song, and Jacob Steinhardt. 2021.
\newblock Measuring mathematical problem solving with the math dataset.
\newblock \emph{arXiv preprint arXiv:2103.03874}.

\bibitem[{Jiang et~al.(2023)Jiang, Jiang, Xue, Zhang, Zhou, Lian, and Wei}]{jiang2023towards}
Gangwei Jiang, Caigao Jiang, Siqiao Xue, James~Y Zhang, Jun Zhou, Defu Lian, and Ying Wei. 2023.
\newblock Towards anytime fine-tuning: Continually pre-trained language models with hypernetwork prompt.
\newblock \emph{arXiv preprint arXiv:2310.13024}.

\bibitem[{Killamsetty et~al.(2021)Killamsetty, Sivasubramanian, Ramakrishnan, De, and Iyer}]{killamsetty2021gradmatchgradientmatchingbased}
Krishnateja Killamsetty, Durga Sivasubramanian, Ganesh Ramakrishnan, Abir De, and Rishabh Iyer. 2021.
\newblock \href {https://arxiv.org/abs/2103.00123} {Grad-match: Gradient matching based data subset selection for efficient deep model training}.
\newblock \emph{Preprint}, arXiv:2103.00123.

\bibitem[{Kirkpatrick et~al.(2017)Kirkpatrick, Pascanu, Rabinowitz, Veness, Desjardins, Rusu, Milan, Quan, Ramalho, Grabska-Barwinska et~al.}]{kirkpatrick2017overcoming}
James Kirkpatrick, Razvan Pascanu, Neil Rabinowitz, Joel Veness, Guillaume Desjardins, Andrei~A Rusu, Kieran Milan, John Quan, Tiago Ramalho, Agnieszka Grabska-Barwinska, and 1 others. 2017.
\newblock Overcoming catastrophic forgetting in neural networks.
\newblock \emph{Proceedings of the national academy of sciences}, 114(13):3521--3526.

\bibitem[{Lester et~al.(2021)Lester, Al-Rfou, and Constant}]{lester-etal-2021-power}
Brian Lester, Rami Al-Rfou, and Noah Constant. 2021.
\newblock \href {https://doi.org/10.18653/v1/2021.emnlp-main.243} {The power of scale for parameter-efficient prompt tuning}.
\newblock In \emph{Proceedings of the 2021 Conference on Empirical Methods in Natural Language Processing}, pages 3045--3059, Online and Punta Cana, Dominican Republic. Association for Computational Linguistics.

\bibitem[{Li and Liang(2021)}]{li2021prefix}
Xiang~Lisa Li and Percy Liang. 2021.
\newblock Prefix-tuning: Optimizing continuous prompts for generation.
\newblock \emph{arXiv preprint arXiv:2101.00190}.

\bibitem[{Li and Hoiem(2017)}]{li2017learning}
Zhizhong Li and Derek Hoiem. 2017.
\newblock Learning without forgetting.
\newblock \emph{IEEE transactions on pattern analysis and machine intelligence}, 40(12):2935--2947.

\bibitem[{Liang and Li(2023)}]{liang2023loss}
Yan-Shuo Liang and Wu-Jun Li. 2023.
\newblock Loss decoupling for task-agnostic continual learning.
\newblock \emph{Advances in Neural Information Processing Systems}, 36:11151--11167.

\bibitem[{Lopez-Paz and Ranzato(2017)}]{10.5555/3295222.3295393}
David Lopez-Paz and Marc'Aurelio Ranzato. 2017.
\newblock Gradient episodic memory for continual learning.
\newblock In \emph{Proceedings of the 31st International Conference on Neural Information Processing Systems}, NIPS'17, page 6470–6479, Red Hook, NY, USA. Curran Associates Inc.

\bibitem[{Maas et~al.(2011)Maas, Daly, Pham, Huang, Ng, and Potts}]{maas2011learning}
Andrew Maas, Raymond~E Daly, Peter~T Pham, Dan Huang, Andrew~Y Ng, and Christopher Potts. 2011.
\newblock Learning word vectors for sentiment analysis.
\newblock In \emph{Proceedings of the 49th annual meeting of the association for computational linguistics: Human language technologies}, pages 142--150.

\bibitem[{McCloskey and Cohen(1989)}]{mccloskey1989catastrophic}
Michael McCloskey and Neal~J Cohen. 1989.
\newblock Catastrophic interference in connectionist networks: The sequential learning problem.
\newblock In \emph{Psychology of learning and motivation}, volume~24, pages 109--165. Elsevier.

\bibitem[{McDonnell et~al.(2023)McDonnell, Gong, Parvaneh, Abbasnejad, and Van~den Hengel}]{mcdonnell2023ranpac}
Mark~D McDonnell, Dong Gong, Amin Parvaneh, Ehsan Abbasnejad, and Anton Van~den Hengel. 2023.
\newblock Ranpac: Random projections and pre-trained models for continual learning.
\newblock \emph{Advances in Neural Information Processing Systems}, 36:12022--12053.

\bibitem[{Mirzasoleiman et~al.(2020)Mirzasoleiman, Bilmes, and Leskovec}]{mirzasoleiman2020coresetsdataefficienttrainingmachine}
Baharan Mirzasoleiman, Jeff Bilmes, and Jure Leskovec. 2020.
\newblock \href {https://arxiv.org/abs/1906.01827} {Coresets for data-efficient training of machine learning models}.
\newblock \emph{Preprint}, arXiv:1906.01827.

\bibitem[{Pan(2010)}]{pan2010q}
SJ—Yang Pan. 2010.
\newblock Q.: A survey on transfer learning.
\newblock \emph{IEEE Transactions on Knowledge and Data Engineering}, 22(10):1345--1359.

\bibitem[{Qin and Joty(2021)}]{qin2021lfpt5}
Chengwei Qin and Shafiq Joty. 2021.
\newblock Lfpt5: A unified framework for lifelong few-shot language learning based on prompt tuning of t5.
\newblock \emph{arXiv preprint arXiv:2110.07298}.

\bibitem[{Razdaibiedina et~al.(2023)Razdaibiedina, Mao, Hou, Khabsa, Lewis, and Almahairi}]{razdaibiedina2023progressive}
Anastasia Razdaibiedina, Yuning Mao, Rui Hou, Madian Khabsa, Mike Lewis, and Amjad Almahairi. 2023.
\newblock \href {https://openreview.net/forum?id=UJTgQBc91_} {Progressive prompts: Continual learning for language models}.
\newblock In \emph{The Eleventh International Conference on Learning Representations}.

\bibitem[{Rebuffi et~al.(2017)Rebuffi, Kolesnikov, Sperl, and Lampert}]{rebuffi2017icarl}
Sylvestre-Alvise Rebuffi, Alexander Kolesnikov, Georg Sperl, and Christoph~H Lampert. 2017.
\newblock icarl: Incremental classifier and representation learning.
\newblock In \emph{Proceedings of the IEEE conference on Computer Vision and Pattern Recognition}, pages 2001--2010.

\bibitem[{Reimers and Gurevych(2019)}]{reimers-2019-sentence-bert}
Nils Reimers and Iryna Gurevych. 2019.
\newblock \href {https://arxiv.org/abs/1908.10084} {Sentence-bert: Sentence embeddings using siamese bert-networks}.
\newblock In \emph{Proceedings of the 2019 Conference on Empirical Methods in Natural Language Processing}. Association for Computational Linguistics.

\bibitem[{Rolf et~al.(2021)Rolf, Worledge, Recht, and Jordan}]{rolf2021representation}
Esther Rolf, Theodora~T Worledge, Benjamin Recht, and Michael Jordan. 2021.
\newblock Representation matters: Assessing the importance of subgroup allocations in training data.
\newblock In \emph{International Conference on Machine Learning}, pages 9040--9051. PMLR.

\bibitem[{R{\"u}ckl{\'e} et~al.(2021)R{\"u}ckl{\'e}, Geigle, Glockner, Beck, Pfeiffer, Reimers, and Gurevych}]{ruckle-etal-2021-adapterdrop}
Andreas R{\"u}ckl{\'e}, Gregor Geigle, Max Glockner, Tilman Beck, Jonas Pfeiffer, Nils Reimers, and Iryna Gurevych. 2021.
\newblock \href {https://doi.org/10.18653/v1/2021.emnlp-main.626} {{AdapterDrop}: {O}n the efficiency of adapters in transformers}.
\newblock In \emph{Proceedings of the 2021 Conference on Empirical Methods in Natural Language Processing}, pages 7930--7946, Online and Punta Cana, Dominican Republic. Association for Computational Linguistics.

\bibitem[{Satapara and Srijith(2024)}]{satapara-srijith-2024-tl}
Shrey Satapara and P.~K. Srijith. 2024.
\newblock \href {https://doi.org/10.18653/v1/2024.emnlp-main.676} {{TL}-{CL}: Task and language incremental continual learning}.
\newblock In \emph{Proceedings of the 2024 Conference on Empirical Methods in Natural Language Processing}, pages 12123--12142, Miami, Florida, USA. Association for Computational Linguistics.

\bibitem[{Shin et~al.(2017)Shin, Lee, Kim, and Kim}]{shin2017continual}
Hanul Shin, Jung~Kwon Lee, Jaehong Kim, and Jiwon Kim. 2017.
\newblock Continual learning with deep generative replay.
\newblock \emph{Advances in neural information processing systems}, 30.

\bibitem[{Smith et~al.(2023)Smith, Karlinsky, Gutta, Cascante-Bonilla, Kim, Arbelle, Panda, Feris, and Kira}]{smith2023coda}
James~Seale Smith, Leonid Karlinsky, Vyshnavi Gutta, Paola Cascante-Bonilla, Donghyun Kim, Assaf Arbelle, Rameswar Panda, Rogerio Feris, and Zsolt Kira. 2023.
\newblock Coda-prompt: Continual decomposed attention-based prompting for rehearsal-free continual learning.
\newblock In \emph{Proceedings of the IEEE/CVF conference on computer vision and pattern recognition}, pages 11909--11919.

\bibitem[{Sun et~al.(2019)Sun, Ho, and Lee}]{sun2019lamol}
Fan-Keng Sun, Cheng-Hao Ho, and Hung-Yi Lee. 2019.
\newblock Lamol: Language modeling for lifelong language learning.
\newblock \emph{arXiv preprint arXiv:1909.03329}.

\bibitem[{Tang et~al.(2023)Tang, Deng, Lin, Han, Liang, and Sun}]{tang2023toolalpaca}
Qiaoyu Tang, Ziliang Deng, Hongyu Lin, Xianpei Han, Qiao Liang, and Le~Sun. 2023.
\newblock \href {https://arxiv.org/abs/2306.05301} {Toolalpaca: Generalized tool learning for language models with 3000 simulated cases}.
\newblock \emph{Preprint}, arXiv:2306.05301.

\bibitem[{Tiwari and Ji(2026)}]{tiwari2026turning}
Anushka Tiwari and Kaiyi Ji. 2026.
\newblock Turning back without forgetting: Selective backward refinement for parameter-efficient continual learning.
\newblock \emph{arXiv preprint arXiv:2606.01379}.

\bibitem[{Tran et~al.(2025)Tran, Tran, Doan, Tran, Phung, Than, and Le}]{tran2025boosting}
Quyen Tran, Tung~Lam Tran, Khanh Doan, Toan Tran, Dinh Phung, Khoat Than, and Trung Le. 2025.
\newblock Boosting multiple views for pretrained-based continual learning.
\newblock In \emph{The Thirteenth International Conference on Learning Representations}.

\bibitem[{Van~de Ven and Tolias(2019)}]{van2019three}
Gido~M Van~de Ven and Andreas~S Tolias. 2019.
\newblock Three scenarios for continual learning.
\newblock \emph{arXiv preprint arXiv:1904.07734}.

\bibitem[{Veniat et~al.(2020)Veniat, Denoyer, and Ranzato}]{veniat2020efficient}
Tom Veniat, Ludovic Denoyer, and Marc'Aurelio Ranzato. 2020.
\newblock Efficient continual learning with modular networks and task-driven priors.
\newblock \emph{arXiv preprint arXiv:2012.12631}.

\bibitem[{Wang et~al.(2019)Wang, Pruksachatkun, Nangia, Singh, Michael, Hill, Levy, and Bowman}]{wang2019superglue}
Alex Wang, Yada Pruksachatkun, Nikita Nangia, Amanpreet Singh, Julian Michael, Felix Hill, Omer Levy, and Samuel Bowman. 2019.
\newblock Superglue: A stickier benchmark for general-purpose language understanding systems.
\newblock \emph{Advances in neural information processing systems}, 32.

\bibitem[{Wang et~al.(2018)Wang, Singh, Michael, Hill, Levy, and Bowman}]{wang2018glue}
Alex Wang, Amanpreet Singh, Julian Michael, Felix Hill, Omer Levy, and Samuel~R Bowman. 2018.
\newblock Glue: A multi-task benchmark and analysis platform for natural language understanding.
\newblock \emph{arXiv preprint arXiv:1804.07461}.

\bibitem[{Wang et~al.(2023{\natexlab{a}})Wang, Li, Wu, Hovy, and Sun}]{wang2023pre}
Haifeng Wang, Jiwei Li, Hua Wu, Eduard Hovy, and Yu~Sun. 2023{\natexlab{a}}.
\newblock Pre-trained language models and their applications.
\newblock \emph{Engineering}, 25:51--65.

\bibitem[{Wang et~al.(2024)Wang, Zhang, Su, and Zhu}]{wang2024comprehensive}
Liyuan Wang, Xingxing Zhang, Hang Su, and Jun Zhu. 2024.
\newblock A comprehensive survey of continual learning: Theory, method and application.
\newblock \emph{IEEE Transactions on Pattern Analysis and Machine Intelligence}.

\bibitem[{Wang et~al.(2023{\natexlab{b}})Wang, Panda, Karlinsky, Feris, Sun, and Kim}]{wang2023multitask}
Zhen Wang, Rameswar Panda, Leonid Karlinsky, Rogerio Feris, Huan Sun, and Yoon Kim. 2023{\natexlab{b}}.
\newblock \href {https://openreview.net/forum?id=Nk2pDtuhTq} {Multitask prompt tuning enables parameter-efficient transfer learning}.
\newblock In \emph{The Eleventh International Conference on Learning Representations}.

\bibitem[{Wang et~al.(2022{\natexlab{a}})Wang, Zhang, Ebrahimi, Sun, Zhang, Lee, Ren, Su, Perot, Dy et~al.}]{wang2022dualprompt}
Zifeng Wang, Zizhao Zhang, Sayna Ebrahimi, Ruoxi Sun, Han Zhang, Chen-Yu Lee, Xiaoqi Ren, Guolong Su, Vincent Perot, Jennifer Dy, and 1 others. 2022{\natexlab{a}}.
\newblock Dualprompt: Complementary prompting for rehearsal-free continual learning.
\newblock In \emph{European conference on computer vision}, pages 631--648. Springer.

\bibitem[{Wang et~al.(2022{\natexlab{b}})Wang, Zhang, Lee, Zhang, Sun, Ren, Su, Perot, Dy, and Pfister}]{wang2022learning}
Zifeng Wang, Zizhao Zhang, Chen-Yu Lee, Han Zhang, Ruoxi Sun, Xiaoqi Ren, Guolong Su, Vincent Perot, Jennifer Dy, and Tomas Pfister. 2022{\natexlab{b}}.
\newblock Learning to prompt for continual learning.
\newblock In \emph{Proceedings of the IEEE/CVF conference on computer vision and pattern recognition}, pages 139--149.

\bibitem[{Wang et~al.(2020)Wang, Mehta, Poczos, and Carbonell}]{wang-etal-2020-efficient}
Zirui Wang, Sanket~Vaibhav Mehta, Barnabas Poczos, and Jaime Carbonell. 2020.
\newblock \href {https://doi.org/10.18653/v1/2020.emnlp-main.39} {Efficient meta lifelong-learning with limited memory}.
\newblock In \emph{Proceedings of the 2020 Conference on Empirical Methods in Natural Language Processing (EMNLP)}, pages 535--548, Online. Association for Computational Linguistics.

\bibitem[{Wu et~al.(2024)Wu, Jiang, and Lian}]{wu-etal-2024-mitigate}
Chenyuan Wu, Gangwei Jiang, and Defu Lian. 2024.
\newblock \href {https://aclanthology.org/2024.findings-acl.650} {Mitigate negative transfer with similarity heuristic lifelong prompt tuning}.
\newblock In \emph{Findings of the Association for Computational Linguistics ACL 2024}, pages 10944--10959, Bangkok, Thailand and virtual meeting. Association for Computational Linguistics.

\bibitem[{Wu et~al.(2002)Wu, Ngai, Carpuat, Larsen, and Yang}]{wu-etal-2002-boosting}
Dekai Wu, Grace Ngai, Marine Carpuat, Jeppe Larsen, and Yongsheng Yang. 2002.
\newblock \href {https://aclanthology.org/W02-2035} {Boosting for named entity recognition}.
\newblock In \emph{{COLING}-02: The 6th Conference on Natural Language Learning 2002 ({C}o{NLL}-2002)}.

\bibitem[{Xu et~al.(2023)Xu, Xie, Qin, Tao, and Wang}]{xu2023parameterefficientfinetuningmethodspretrained}
Lingling Xu, Haoran Xie, Si-Zhao~Joe Qin, Xiaohui Tao, and Fu~Lee Wang. 2023.
\newblock \href {https://arxiv.org/abs/2312.12148} {Parameter-efficient fine-tuning methods for pretrained language models: A critical review and assessment}.
\newblock \emph{Preprint}, arXiv:2312.12148.

\bibitem[{Yin et~al.(2022)Yin, Li, and Xiong}]{yin-etal-2022-contintin}
Wenpeng Yin, Jia Li, and Caiming Xiong. 2022.
\newblock \href {https://doi.org/10.18653/v1/2022.acl-long.218} {{C}on{T}in{T}in: Continual learning from task instructions}.
\newblock In \emph{Proceedings of the 60th Annual Meeting of the Association for Computational Linguistics (Volume 1: Long Papers)}, pages 3062--3072, Dublin, Ireland. Association for Computational Linguistics.

\bibitem[{Yoon et~al.(2018)Yoon, Yang, Lee, and Hwang}]{yoon2018lifelong}
Jaehong Yoon, Eunho Yang, Jeongtae Lee, and Sung~Ju Hwang. 2018.
\newblock \href {https://openreview.net/forum?id=Sk7KsfW0-} {Lifelong learning with dynamically expandable networks}.
\newblock In \emph{International Conference on Learning Representations}.

\bibitem[{Zenke et~al.(2017)Zenke, Poole, and Ganguli}]{zenke2017continual}
Friedemann Zenke, Ben Poole, and Surya Ganguli. 2017.
\newblock Continual learning through synaptic intelligence.
\newblock In \emph{International conference on machine learning}, pages 3987--3995. PMLR.

\bibitem[{Zhang et~al.(2015)Zhang, Zhao, and LeCun}]{zhang2015character}
Xiang Zhang, Junbo Zhao, and Yann LeCun. 2015.
\newblock Character-level convolutional networks for text classification.
\newblock \emph{Advances in neural information processing systems}, 28.

\bibitem[{Zhu et~al.(2022)Zhu, Li, Mi, Zhu, and Huang}]{zhu-etal-2022-continual}
Qi~Zhu, Bing Li, Fei Mi, Xiaoyan Zhu, and Minlie Huang. 2022.
\newblock \href {https://doi.org/10.18653/v1/2022.acl-long.80} {Continual prompt tuning for dialog state tracking}.
\newblock In \emph{Proceedings of the 60th Annual Meeting of the Association for Computational Linguistics (Volume 1: Long Papers)}, pages 1124--1137, Dublin, Ireland. Association for Computational Linguistics.

\end{thebibliography}

% \clearpage
% \appendix

\clearpage
\onecolumn

\appendix
\startcontents[appendix]

\section*{Contents of the Appendix}
\printcontents[appendix]{}{1}{}

\clearpage
\twocolumn

% \section{Appendix}

\section{Algorithmic 
Details}\label{append:alg}

Algorithm~\ref{alg:rep_selection} presents the representative sample selection procedure used to construct compact yet diverse task-specific training subsets.

\begin{minipage}[htbp]{0.45\textwidth}
% \tiny
\begin{algorithm}[H]
\footnotesize
\caption{\small Representative Sample Selection}
\label{alg:rep_selection}
\begin{algorithmic}[1]
\Require Dataset $\mathcal{D} = \{(x_j, y_j)\}$, samples per class $k$, embedding model $f_{\text{embed}}$, number of clusters $C$
\Ensure Subset $\mathcal{D}_{\text{rep}}$
\State Initialize $\mathcal{D}_{\text{rep}} \gets \emptyset$
\For{each label $y$}
    \State $\mathcal{D}_y \gets \{x_j \mid y_j = y\}$
    \State Embed each $x_j$: $\mathbf{e}_j \gets f_{\text{embed}}(x_j)$
    \State Run K-Means on $\{\mathbf{e}_j\}$ into $C$ clusters
    \For{each cluster $c$}
        \State Select top $k/C$ samples by similarity to center
        \State Add to $\mathcal{D}_{\text{rep}}$
    \EndFor
    \If{selected < $k$}
        \State Add random samples to reach $k$ total
    \EndIf
\EndFor
\State Shuffle and return $\mathcal{D}_{\text{rep}}$
\end{algorithmic}
\end{algorithm}
\end{minipage}
% \hfill

\section{Full Experimental Setup and Configurations}
\label{sup_details}

\begin{table*}[htbp]
  \centering
  \scriptsize
  \setlength{\tabcolsep}{5pt}
  \begin{tabular}{l l l l l}
    \toprule
    \rowcolor{gray!25}
    \textbf{Dataset} & \textbf{Category} & \textbf{Task} & \textbf{Domain} & \textbf{Metric} \\
    \midrule
    Yelp & CL benchmark & sentiment analysis & Yelp reviews & acc \\
    Amazon & CL benchmark & sentiment analysis & Amazon reviews & acc \\
    DBpedia & CL benchmark & topic classification & Wikipedia & acc \\
    Yahoo & CL benchmark & QA & Yahoo Q\&A & acc \\
    AG News & CL benchmark & topic classification & news & acc \\
    MNLI & GLUE & NLI & various & acc \\
    QQP & GLUE & paraphrase detection & Quora & acc/F1 \\
    RTE & GLUE & NLI & news, Wikipedia & acc \\
    SST-2 & GLUE & sentiment analysis & movie reviews & acc \\
    WiC & SuperGLUE & word sense disambiguation & lexical db & acc \\
    CB & SuperGLUE & NLI & various & acc \\
    COPA & SuperGLUE & QA & blogs, encyclopedia & acc \\
    BoolQ & SuperGLUE & boolean QA & Wikipedia & acc \\
    MultiRC & SuperGLUE & QA & various & F1/EM \\
    IMDB & Other & sentiment analysis & movie reviews & acc \\
    \bottomrule
  \end{tabular}
    \caption{\footnotesize Overview of the 15 datasets used in our CL experiments, including their evaluation metrics. Datasets from CL benchmark \citep{zhang2015character}, GLUE \citep{wang2018glue}, and SuperGLUE \citep{wang2019superglue} were utilized, along with the IMDB movie reviews dataset. For tasks with two evaluation metrics, we report the average as the final performance measure.}
  \label{tab:datasets_overview}
\end{table*}

\begin{table}[t]
\centering
\scriptsize
\setlength{\tabcolsep}{6pt}
\renewcommand{\arraystretch}{1.08}
\begin{tabular}{c|l|l|l}
\toprule
\rowcolor{gray!25}
\textbf{\#} & \textbf{Dataset name} & \textbf{Task} & \textbf{Metric} \\
\midrule
1  & task639  & dialogue generation     & Rouge-L \\
2  & task1590 & dialogue generation     & Rouge-L \\
3  & task1729 & dialogue generation     & Rouge-L \\
4  & task181  & information extraction  & Rouge-L \\
5  & task748  & information extraction  & Rouge-L \\
6  & task1510 & information extraction  & Rouge-L \\
7  & task002  & question answering      & Rouge-L \\
8  & task073  & question answering      & Rouge-L \\
9  & task591  & question answering      & Rouge-L \\
10 & task511  & summarization           & Rouge-L \\
11 & task1290 & summarization           & Rouge-L \\
12 & task1572 & summarization           & Rouge-L \\
13 & task363  & sentiment analysis      & accuracy \\
14 & task875  & sentiment analysis      & accuracy \\
15 & task1687 & sentiment analysis      & accuracy \\
\bottomrule
\end{tabular}
\caption{\footnotesize Details of the SuperNI benchmark used for open-ended instruction-based continual learning. The benchmark includes dialogue generation, information extraction, question answering, summarization, and sentiment analysis tasks.}
\label{tab:superni_details}
\end{table}

\begin{table*}[htbp]
  \centering
  \scriptsize
  \setlength{\tabcolsep}{6pt}
  \renewcommand{\arraystretch}{1.15}
  \begin{tabular}{l p{14cm}}
    \toprule
    \rowcolor{gray!25}
    \textbf{Order} & \textbf{Task Sequence} \\
    \midrule
    L1 & mnli $\rightarrow$ cb $\rightarrow$ wic $\rightarrow$ copa $\rightarrow$ qqp $\rightarrow$ boolq $\rightarrow$ rte $\rightarrow$ imdb $\rightarrow$ yelp $\rightarrow$ amazon $\rightarrow$ sst2 $\rightarrow$ dbpedia $\rightarrow$ ag $\rightarrow$ multirc $\rightarrow$ yahoo \\
    L2 & multirc $\rightarrow$ boolq $\rightarrow$ wic $\rightarrow$ mnli $\rightarrow$ cb $\rightarrow$ copa $\rightarrow$ qqp $\rightarrow$ rte $\rightarrow$ imdb $\rightarrow$ sst2 $\rightarrow$ dbpedia $\rightarrow$ ag $\rightarrow$ yelp $\rightarrow$ amazon $\rightarrow$ yahoo \\
    L3 & yelp $\rightarrow$ amazon $\rightarrow$ mnli $\rightarrow$ cb $\rightarrow$ copa $\rightarrow$ qqp $\rightarrow$ rte $\rightarrow$ imdb $\rightarrow$ sst2 $\rightarrow$ dbpedia $\rightarrow$ ag $\rightarrow$ yahoo $\rightarrow$ multirc $\rightarrow$ boolq $\rightarrow$ wic \\
    \midrule
    L4 & sst2 $\rightarrow$ imdb $\rightarrow$ yelp $\rightarrow$ amazon $\rightarrow$ ag $\rightarrow$ yahoo $\rightarrow$ dbpedia $\rightarrow$ mnli $\rightarrow$ rte $\rightarrow$ cb $\rightarrow$ qqp $\rightarrow$ copa $\rightarrow$ boolq $\rightarrow$ wic $\rightarrow$ multirc \\
    L5 & multirc $\rightarrow$ wic $\rightarrow$ boolq $\rightarrow$ copa $\rightarrow$ qqp $\rightarrow$ cb $\rightarrow$ rte $\rightarrow$ mnli $\rightarrow$ dbpedia $\rightarrow$ yahoo $\rightarrow$ ag $\rightarrow$ amazon $\rightarrow$ yelp $\rightarrow$ imdb $\rightarrow$ sst2 \\
    L6 & sst2 $\rightarrow$ copa $\rightarrow$ ag $\rightarrow$ imdb $\rightarrow$ mnli $\rightarrow$ yahoo $\rightarrow$ rte $\rightarrow$ yelp $\rightarrow$ qqp $\rightarrow$ cb $\rightarrow$ amazon $\rightarrow$ dbpedia $\rightarrow$ boolq $\rightarrow$ wic $\rightarrow$ multirc \\
    \midrule
    NT1 & multirc $\rightarrow$ wic $\rightarrow$ mnli $\rightarrow$ cb $\rightarrow$ rte $\rightarrow$ qqp $\rightarrow$ yahoo $\rightarrow$ yelp $\rightarrow$ amazon \\
    NT2 & amazon $\rightarrow$ yelp $\rightarrow$ yahoo $\rightarrow$ qqp $\rightarrow$ rte $\rightarrow$ cb $\rightarrow$ mnli $\rightarrow$ wic $\rightarrow$ multirc \\
    NT3 & yahoo $\rightarrow$ mnli $\rightarrow$ amazon $\rightarrow$ cb $\rightarrow$ yelp $\rightarrow$ rte $\rightarrow$ qqp $\rightarrow$ multirc $\rightarrow$ wic \\
    \midrule
    MTCL-Bench & math-prealgebra $\rightarrow$ math-algebra $\rightarrow$ math-geometry $\rightarrow$ math-counting-probability $\rightarrow$ math-number-theory $\rightarrow$ math-intermediate-algebra $\rightarrow$ math-precalculus $\rightarrow$ gsm8k $\rightarrow$ tool-development $\rightarrow$ tool-geocoding $\rightarrow$ tool-games-comics $\rightarrow$ tool-transportation $\rightarrow$ tool-finance $\rightarrow$ tool-video $\rightarrow$ tool-health \\
    \bottomrule
  \end{tabular}
    \caption{\footnotesize Task sequence orders used in our CL experiments. Orders L1--L3 correspond to prior long-sequence CL benchmarks, while L4--L6 are our proposed custom sequences designed to study easy-to-hard, hard-to-easy, and mixed task progressions. NT1--NT3 denote negative-transfer orders constructed from deliberately dissimilar tasks. MTCL-Bench denotes our Math-Tool continual learning benchmark, which extends evaluation to mathematical reasoning and tool-selection tasks.}
  \label{tab:task_sequence_orders_full}
\end{table*}

\begin{table}[htbp]
\centering
\scriptsize
\setlength{\tabcolsep}{6pt}
\renewcommand{\arraystretch}{1.15}
\begin{tabular}{c|p{0.78\linewidth}}
\toprule
\rowcolor{gray!25}
\textbf{Order} & \textbf{SuperNI Benchmark} \\
\midrule
1 &
task1572 $\rightarrow$ task363 $\rightarrow$ task1290 $\rightarrow$ task181 $\rightarrow$
task002 $\rightarrow$ task1510 $\rightarrow$ task639 $\rightarrow$ task1729 $\rightarrow$
task073 $\rightarrow$ task1590 $\rightarrow$ task748 $\rightarrow$ task511 $\rightarrow$
task591 $\rightarrow$ task1687 $\rightarrow$ task875 \\
\midrule
2 &
task748 $\rightarrow$ task073 $\rightarrow$ task1590 $\rightarrow$ task639 $\rightarrow$
task1572 $\rightarrow$ task1687 $\rightarrow$ task591 $\rightarrow$ task363 $\rightarrow$
task1510 $\rightarrow$ task1729 $\rightarrow$ task181 $\rightarrow$ task511 $\rightarrow$
task002 $\rightarrow$ task1290 $\rightarrow$ task875 \\
\bottomrule
\end{tabular}
\caption{ \footnotesize Task orders used for the SuperNI benchmark.}
\label{tab:superni_orders}
\end{table}

\subsection{Task Sequence Orders and Benchmark Construction}
\label{sec:order_create}

\begin{figure}[htbp]
    \centering
        \includegraphics[width=\linewidth]{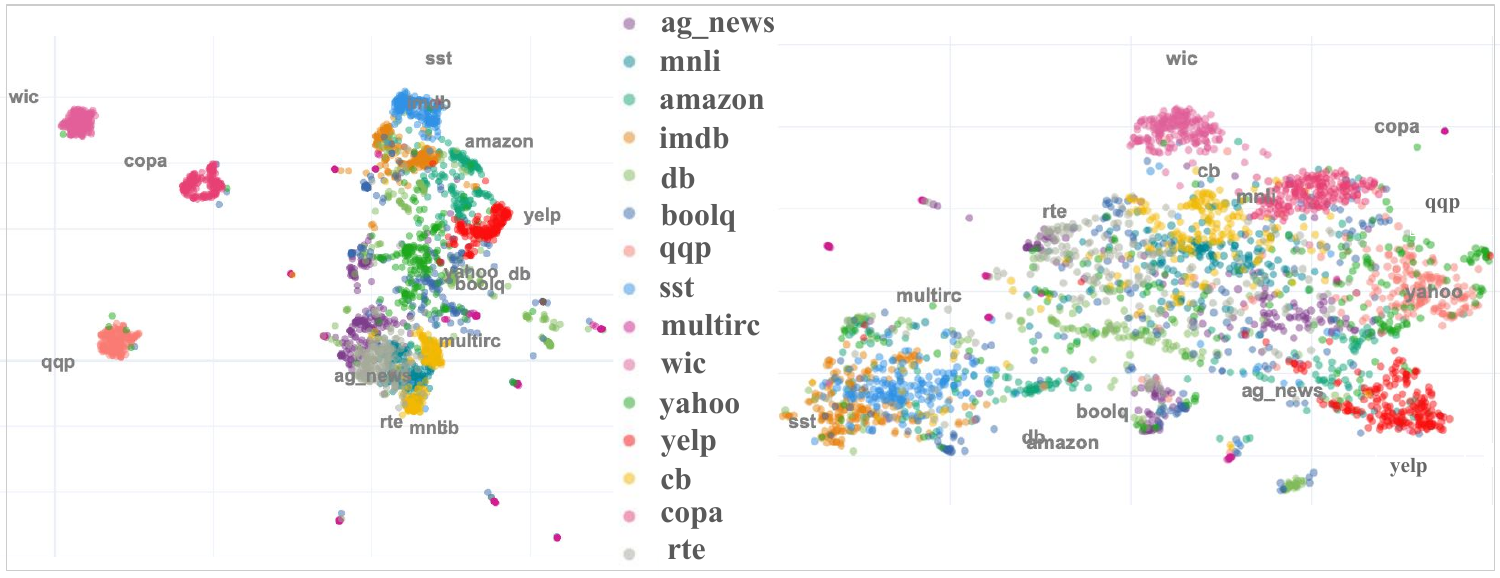}    % \vspace{-0.25cm}
    \caption{\small
    (a) Left - Dataset embeddings using EmbeddingGemma-300M.
    (b) Right - Dataset embeddings using sentence-transformers/all-MiniLM-L6-v2
    }
    \label{fig:cluster}
\end{figure}

\paragraph{Long-Sequence CL Setting:}
Following the approach of \citep{razdaibiedina2023progressive}, we consider a total of 15 distinct tasks. These consist of five datasets from the CL benchmark \citep{zhang2015character}: AG News (topic classification), Amazon Reviews (sentiment analysis), Yelp Reviews (sentiment analysis), DBpedia (Wikipedia text classification), Yahoo Answers (Q\&A classification), four tasks from the GLUE benchmark (MNLI, QQP, RTE, SST2) \citep{wang2018glue}, five tasks from the SuperGLUE benchmark (WiC, CB, COPA, MultiRC, BoolQ) \citep{wang2019superglue}, and the IMDB movie reviews dataset for sentiment analysis \citep{maas2011learning}. Figure~\ref{fig:cluster} visualizes task embedding distributions under two sentence encoders, highlighting semantic clustering and separation among tasks in the sequence. We evaluate our methods across six different task sequence orders. Three of these sequences follow standard long-sequence continual learning orders that have been used in prior works (e.g., L1–L3) \citep{razdaibiedina2023progressive, guo2024q}. We additionally propose three novel task orders: (1) \textit{Order L4: easy-to-hard}, where tasks are grouped by category and simpler tasks such as binary sentiment classification are introduced first, followed by more complex reasoning tasks like MultiRC and COPA; (2) \textit{Order L5: hard-to-easy}, which reverses this progression; and (3) \textit{Order L6: mixed}, where tasks from different categories and difficulty levels are randomly interleaved. These variations allow us to assess how task presentation order and semantic similarity affect forgetting, transfer, and model generalization in long-horizon continual learning scenarios.

{\bf Negative Transfer Benchmark.} To evaluate the robustness of our approach against negative transfer \citep{pan2010q}, we introduce a \textit{Negative Transfer Benchmark} (NT1, NT2, NT3) inspired by the SHLPT framework \citep{wu-etal-2024-mitigate}. Specifically, we identify task pairs that exhibit negative transfer based on the analysis provided in the SHLPT. Using these insights, we create three task sequences that are likely to cause negative transfer, meaning earlier tasks make it harder to learn later ones. This helps us test how well our method handles such challenges.

\begin{table}[htbp]
\centering
\scriptsize
\setlength{\tabcolsep}{6pt}
\begin{tabular}{lcc}
\toprule
\rowcolor{gray!25}
\textbf{Tool Category} & \textbf{\# Examples} & \textbf{\# Tools} \\
\midrule
Development & 298 & 36 \\
Geocoding & 287 & 32 \\
Games \& Comics & 276 & 29 \\
Transportation & 215 & 21 \\
Finance & 138 & 14 \\
Video & 124 & 14 \\
Health & 122 & 12 \\
\bottomrule
\end{tabular}
\caption{\footnotesize Statistics of the selected ToolAlpaca categories used in the Math--Tool continual learning benchmark.}
\label{tab:tool_stats}
\end{table}

{\bf Math-Tool Continual Learning Benchmark (MTCL-Bench)} To extend evaluation beyond classification-style continual learning, we construct a 15-task Math-Tool benchmark. For mathematical reasoning, we use competition\_math \citep{hendrycksmath2021} and GSM8K, treating the seven Competition MATH subjects as separate tasks and GSM8K as the eighth task. For each Competition MATH subject, we select 1K examples for training and a separate set of 100 examples for testing. For GSM8K, we sample 1K training examples from the official training split and 100 examples from the official test split. For tool use, we process ToolAlpaca \citep{tang2023toolalpaca} into seven category-level tool-selection tasks: Development, Geocoding, Games \& Comics, Transportation, Finance, Video, and Health. Each ToolAlpaca instance is converted into a category-level tool-selection example and split using a 90/10 stratified split by category and gold tool; category statistics are shown in Table~\ref{tab:tool_stats}.

\begin{table}[t]
\centering
\scriptsize
\setlength{\tabcolsep}{3.5pt}
\renewcommand{\arraystretch}{1.1}
\begin{tabular}{l p{2.55cm} p{2.55cm}}
\toprule
\rowcolor{gray!25}
\textbf{Argument} & \textbf{Autoregressive} & \textbf{Seq-to-seq} \\
\midrule
Architecture & Autoregressive & Seq-to-seq \\
Base model & Qwen3-4B-ins; LLaMA-2-7B, LLaMA-3.1-8B-ins & T5 (and Flan-T5) \\
Number of tasks & 15 & 15 \\
Prefix length & 10 & 10 \\
Max sequence length & 256 & 256 \\
Learning rate & 0.03 & 0.3 \\
Batch size & 16 & 8 \\
Samples per class ($k$) & 1000 & 1000 \\
Training epochs & 5, 3 (MTCL-Bench) & 10 \\
Frozen backbone & Yes & Yes \\
Evaluation setting & Task-agnostic & Task-agnostic \\
Fixed test data & Yes & Yes \\
Early stopping & Enabled & Enabled \\
\bottomrule
\end{tabular}
\caption{\small Training configurations used in our experiments for autoregressive and seq-to-seq backbones.}
\label{tab:training_config}
\end{table}

\subsection{Detailed Experimental Configurations}
\label{sec:experiment_detail}

We report task-agnostic continual prompt-learning results on long-sequence benchmarks (L1--L6) and negative-transfer sequences (NT1--NT3). Unless otherwise noted, we use a prompt length of 10 tokens and $k{=}1000$ representative samples per class. We train seq-to-seq models for 10 epochs and autoregressive models for 5 epochs, following the configurations summarized in Table~\ref{tab:training_config}.

\subsubsection{Baselines Details.} 
We compare against seven continual-learning baselines:\footnote{We do not include Q-Tuning \citep{guo2024q} due to its incompatible setup and unavailable code under company restrictions.} 
\textbf{Finetune} \citep{wang-etal-2020-efficient}, 
\textbf{Prompt Tuning} \citep{lester-etal-2021-power, qin2021lfpt5}, 
\textbf{Data Replay} \citep{de2019episodic}, 
\textbf{LFPT5} \citep{qin2021lfpt5}, 
\textbf{Per-task Prompt} \citep{lester-etal-2021-power, qin2021lfpt5}, 
\textbf{ProgPrompt} \citep{razdaibiedina2023progressive}, and 
\textbf{SHLPT} \citep{wu-etal-2024-mitigate}. 
We report average test accuracy, backward transfer (BWT), and forward transfer (FWT) \citep{10.5555/3295222.3295393}, averaged over three runs.

For the MTCL-Bench, we additionally compare GRID against four baselines to the new setting: (i) pretrained models evaluated directly on each task without training, (ii) per-task SFT, where a separate LoRA adapter is trained independently for each task to estimate task-specific upper-bound performance, (iii) ProgPrompt, where task prompts are accumulated without pruning or merging, and (iv) continual Prompt Tuning, where a single soft prompt is sequentially updated across tasks.

\subsubsection{Models.}
To demonstrate architectural generality, we evaluate both \textbf{seq-to-seq} and \textbf{autoregressive} backbones. For seq-to-seq, we use T5 and Flan-T5 across \texttt{small}, \texttt{base}, \texttt{large}, and \texttt{3B} variants. For autoregressive evaluation, we use \texttt{Qwen/Qwen3-4B-Instruct} and \texttt{meta-llama/Llama-2-7b-hf} under the same task-agnostic setup. For MTCL-Bench, which requires stronger mathematical reasoning and tool-selection capabilities, we use \texttt{Qwen/Qwen3-4B-Instruct-2507} and \texttt{meta-llama/Llama-3.1-8B-Instruct}. Training configuration in  Table~\ref{tab:training_config}.

\subsubsection{Prompt Initialization Strategies.}
For both architectures, we freeze the backbone parameters and optimize only task-specific soft prompts. 
For instruction-tuned autoregressive models, prompts are initialized using a short natural-language instruction that enumerates the task label names (e.g., ``Classify into: \ldots''), and the corresponding token embeddings are used to seed the 10-token soft prompt. 
For seq-to-seq models, prompts are initialized directly in the embedding space. For MTCL-Bench, we use task-type-specific initialization: math prompts are initialized with ``Solve the math problem step by step and write the final answer in boxed form,'' while tool-use prompts are initialized with ``Select the best tool from the candidate tool list for the user request.
We also experimented with random prompt initialization and observed a slight but not statistically significant degradation in performance compared to instruction-based initialization.

\subsubsection{Representative sample selection.}
We construct a refined training subset per class using the clustering-based representative selection procedure in Algorithm~\ref{alg:rep_selection}. We embed inputs using \texttt{all-MiniLM-L6-v2} and run K-means to select examples closest to cluster centers, ensuring coverage and diversity while keeping a fixed per-class budget.

% \section{Additional Results}

% \subsection{Impact of Task Order on Backward Transfer}
% \label{sec:order_analysis}

% An important trend observed in Table~\ref{tab:forgotten_tasks_comparison} is that \textbf{Order L5 (hard-to-easy)} consistently outperforms \textbf{Order L4 (easy-to-hard)} in terms of both backward transfer (BWT) and the number of forgotten tasks. For an identical set of tasks, Order L5 yields less negative BWT scores, indicating reduced forgetting, and consistently results in fewer forgotten tasks. This suggests that introducing more challenging tasks earlier in training promotes stronger and more stable representations, leading to improved long-term retention. 

% We further observe that \textbf{Order L6 (mixed)} performs better than Order L4 but slightly worse than Order L5, indicating that interleaving tasks of varying difficulty offers a reasonable trade-off between stability and flexibility. Overall, these results highlight task order as a critical factor in long-horizon continual learning, with hard-to-easy curricula providing the most robust retention under task-agnostic prompt-based learning.

\section{Additional Results}

\subsection{Impact of Task Order on Backward Transfer}
\label{sec:order_analysis}

Table~\ref{tab:bwt_comparison} provides a detailed analysis of backward transfer (BWT) across 
multiple long-sequence task orderings (L1--L6) and model scales. 
A consistent pattern emerges across all seq-to-seq model variants: 
\textbf{Order L5 (hard-to-easy)} yields the least negative BWT, indicating significantly reduced forgetting 
compared to \textbf{Order L4 (easy-to-hard)}. 
For identical task sets, L5 improves average BWT by over 50\% relative to ProgPrompt, 
demonstrating that exposing the model to more challenging tasks earlier leads to stronger and more stable prompt representations.

\begin{figure}[t]
    \centering
    \small

    \begin{minipage}{0.48\columnwidth}
        \centering
        \includegraphics[width=\linewidth]{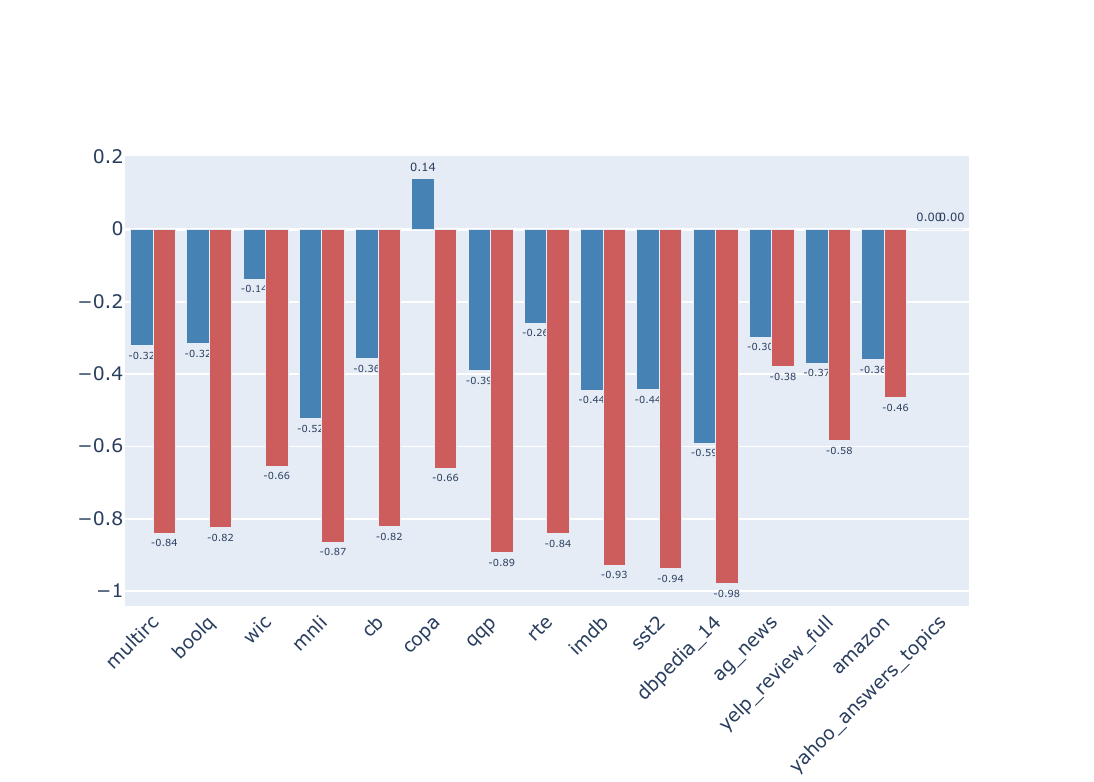}
        \caption*{\small (a) Order L2.}
    \end{minipage}
    \hfill
    \begin{minipage}{0.48\columnwidth}
        \centering
        \includegraphics[width=\linewidth]{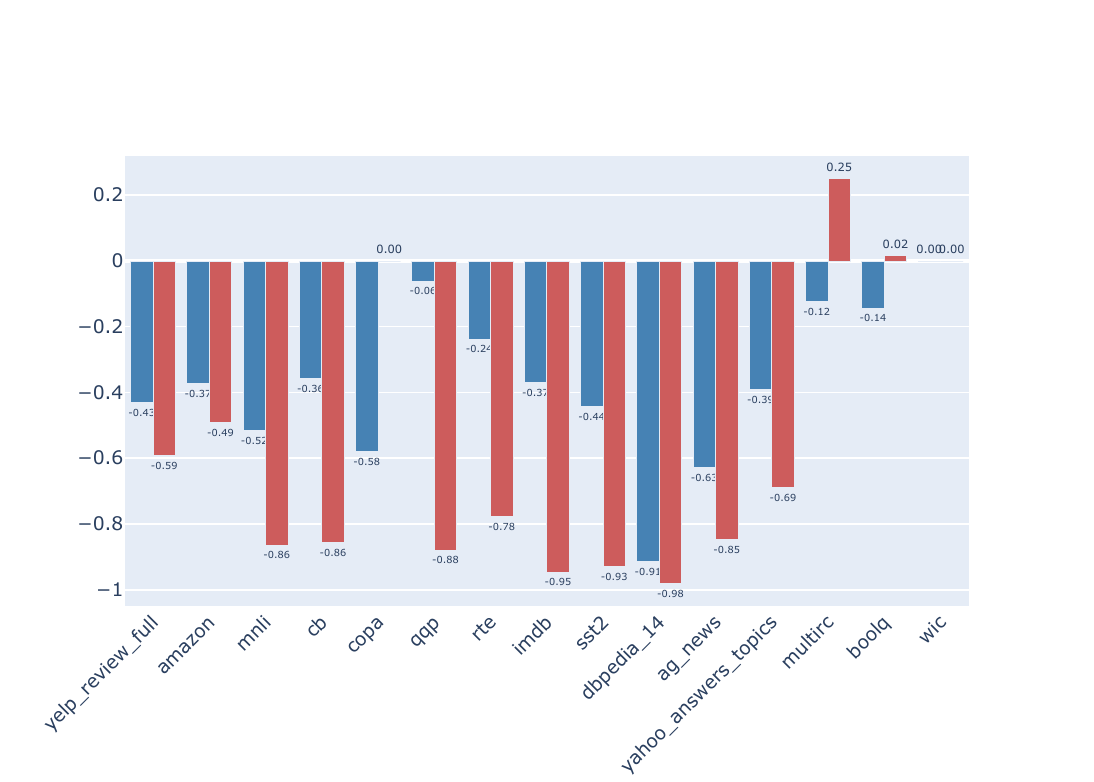}
        \caption*{\small (b) Order L3.}
    \end{minipage}

    \caption{\small
    Per-task BWT comparison between our method (blue) and the ProgPrompt (red).
    Positive bars indicate improved retention of prior tasks.
    Our method shows consistent BWT gains across task orders, demonstrating its effectiveness
    in mitigating forgetting across diverse task types.
    }
    \label{fig:bwt_l1_l2}
\end{figure}

\begin{figure}[t]
    \centering

    \begin{minipage}{0.32\columnwidth}
        \centering
        \includegraphics[width=\linewidth]{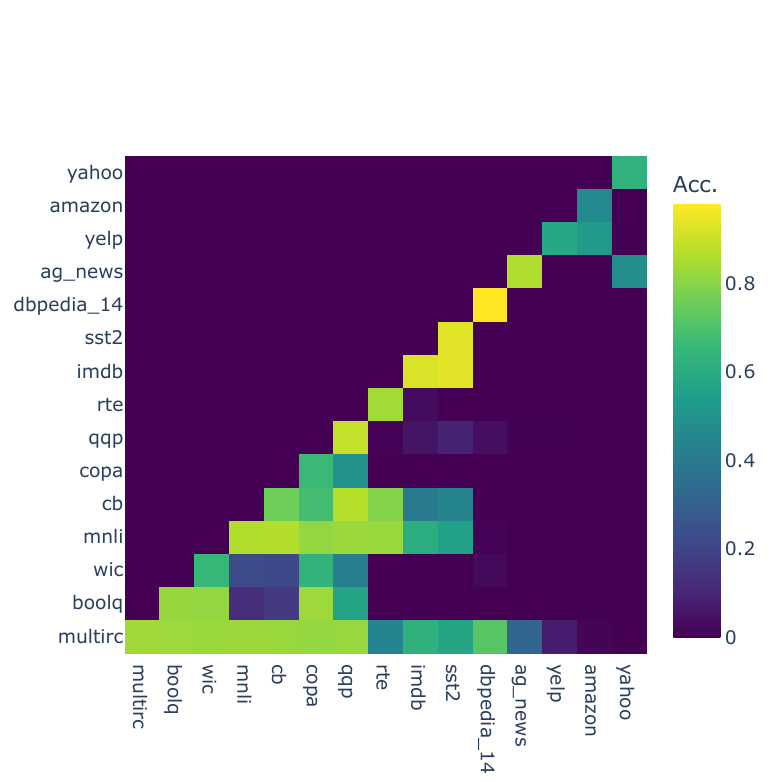}
        \caption*{\small (A) ProgPrompt}
    \end{minipage}
    \hfill
    \begin{minipage}{0.32\columnwidth}
        \centering
        \includegraphics[width=\linewidth]{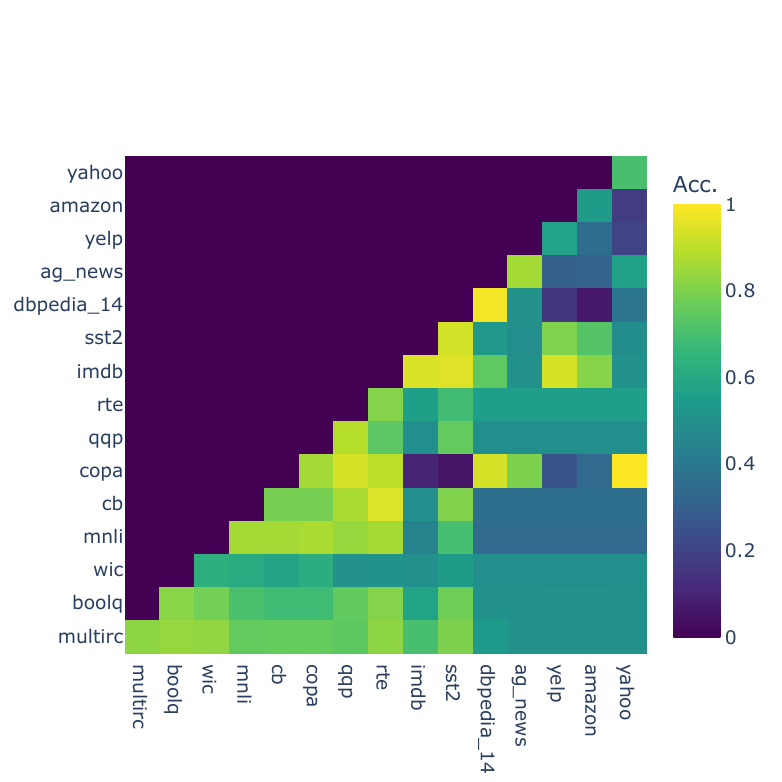}
        \caption*{\small (B) GRID}
    \end{minipage}
    \hfill
    \begin{minipage}{0.32\columnwidth}
        \centering
        \includegraphics[width=\linewidth]{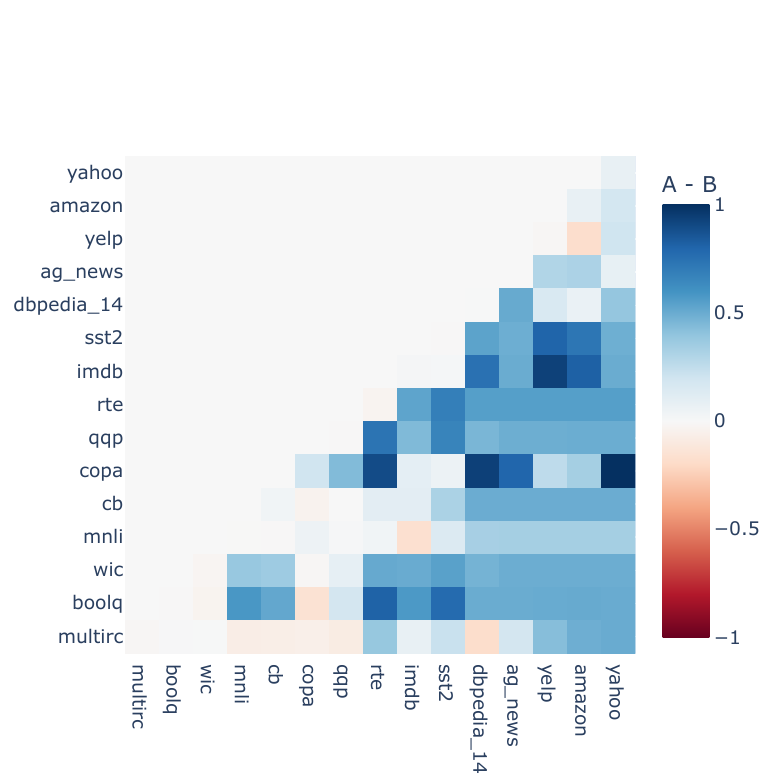}
        \caption*{\small (C) B-A}
    \end{minipage}

    \caption{\small Heatmaps of backward transfer scores on previous tasks for Order L2.
    Brighter values indicate better retention of earlier tasks.
    }
    \label{fig:o5_heat}
\end{figure}

\begin{figure}[t]
    \centering
    \small

    \begin{minipage}{0.48\columnwidth}
        \centering
        \includegraphics[width=\linewidth]{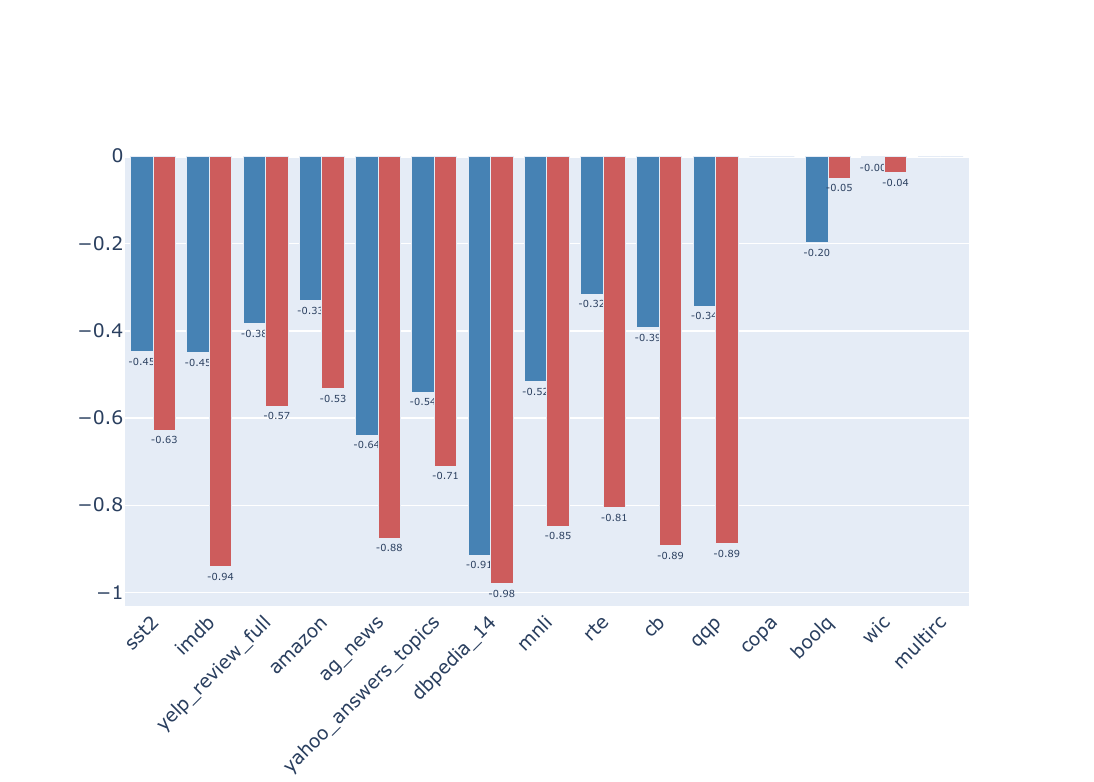}
        \caption*{\small (a) Order L4.}
    \end{minipage}
    \hfill
    \begin{minipage}{0.48\columnwidth}
        \centering
        \includegraphics[width=\linewidth]{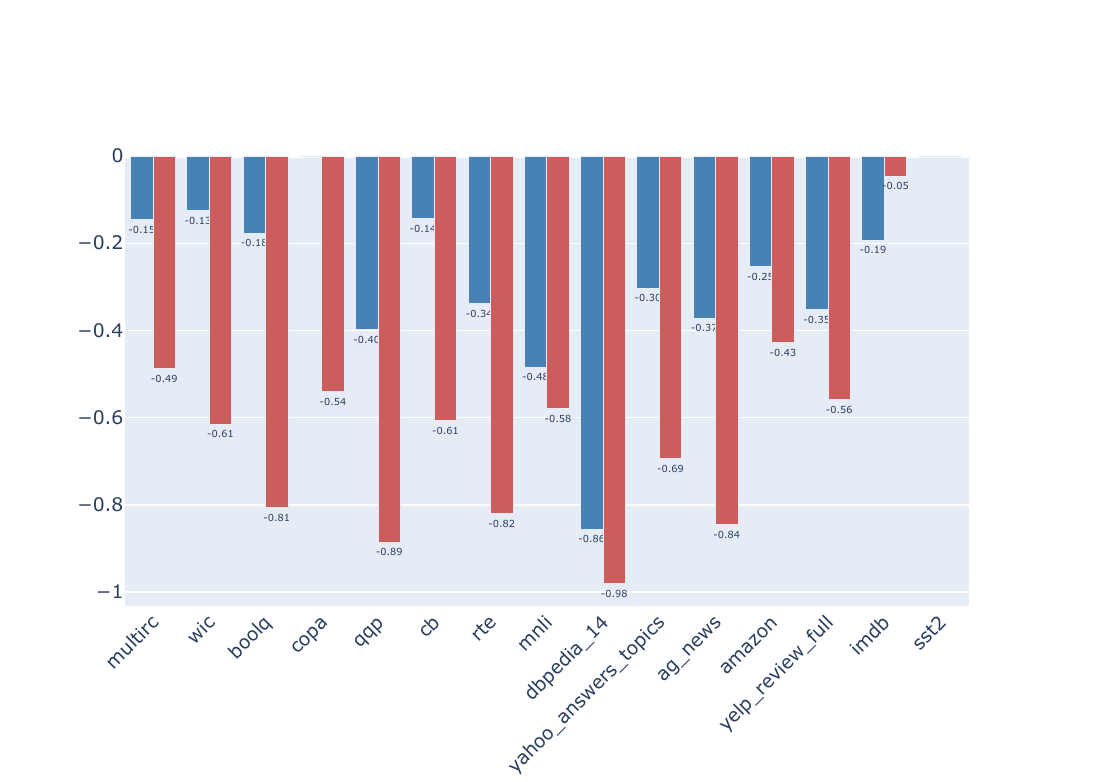}
        \caption*{\small (b) Order L5.}
    \end{minipage}

    \caption{\small
    Per-task BWT comparison between our method (blue) and the ProgPrompt (red).
    Positive bars indicate improved retention of prior tasks.
    Our method shows consistent BWT gains across task orders, demonstrating its effectiveness
    in mitigating forgetting across diverse task types.
    }
    \label{fig:bwt_l4_l5}
\end{figure}

\begin{figure}[t]
    \centering

    \begin{minipage}{0.32\columnwidth}
        \centering
        \includegraphics[width=\linewidth]{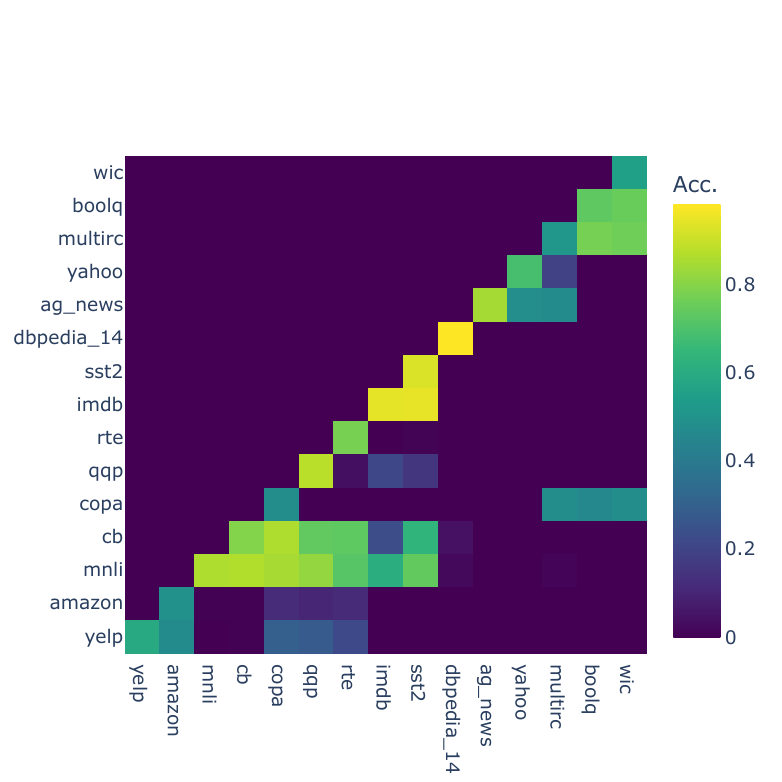}
        \caption*{\small (A) ProgPrompt}
    \end{minipage}
    \hfill
    \begin{minipage}{0.32\columnwidth}
        \centering
        \includegraphics[width=\linewidth]{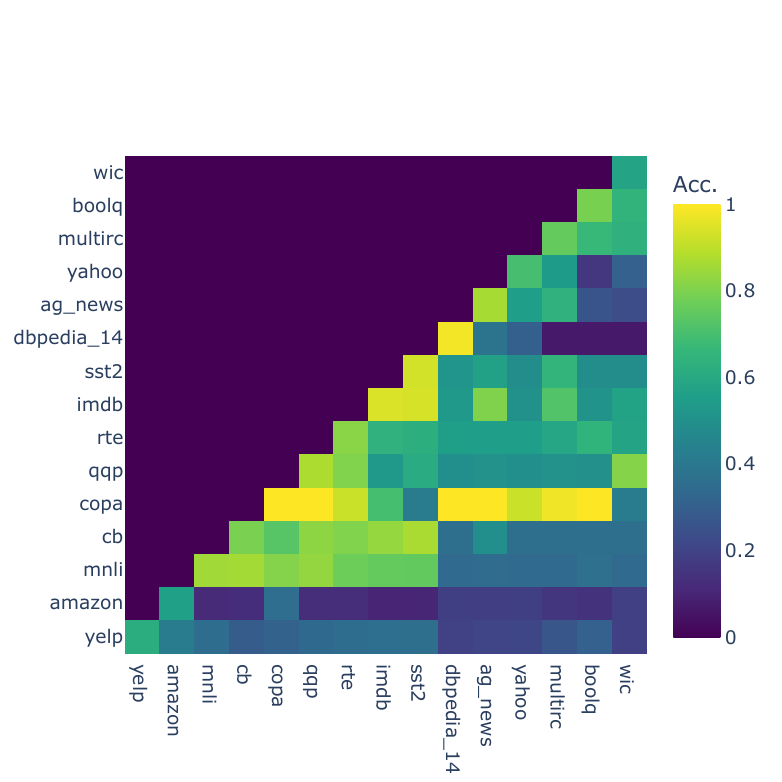}
        \caption*{\small (B) GRID}
    \end{minipage}
    \hfill
    \begin{minipage}{0.32\columnwidth}
        \centering
        \includegraphics[width=\linewidth]{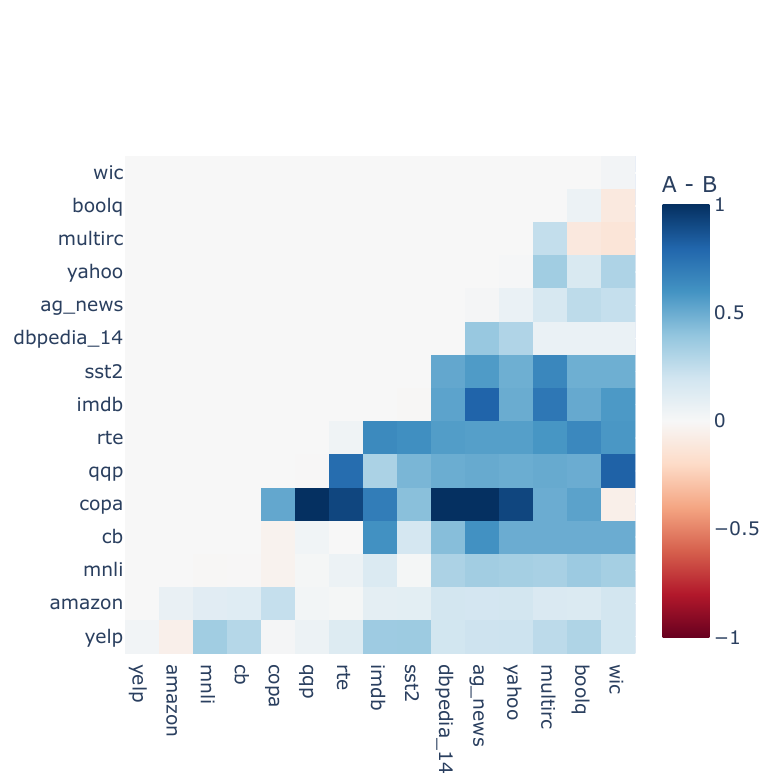}
        \caption*{\small (C) B--A}
    \end{minipage}

    \caption{\small
    Heatmaps of backward transfer scores on previous tasks for Order L3.
    Brighter values indicate better retention of earlier tasks.
    }
    \label{fig:o6_heat}
\end{figure}

\begin{figure}[t]
    \centering

    \begin{minipage}{0.32\columnwidth}
        \centering
        \includegraphics[width=\linewidth]{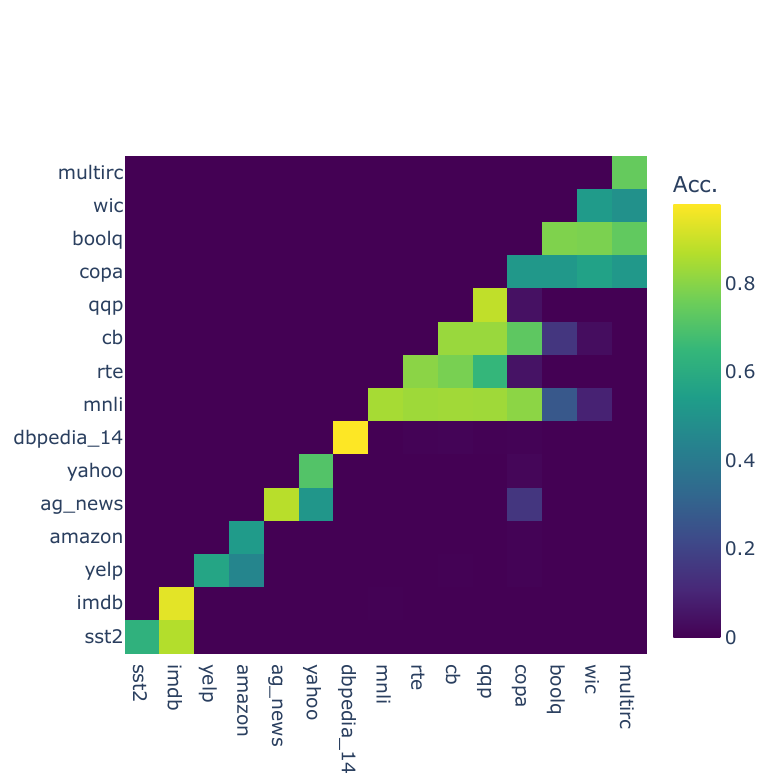}
        \caption*{\small (A) ProgPrompt}
    \end{minipage}
    \hfill
    \begin{minipage}{0.32\columnwidth}
        \centering
        \includegraphics[width=\linewidth]{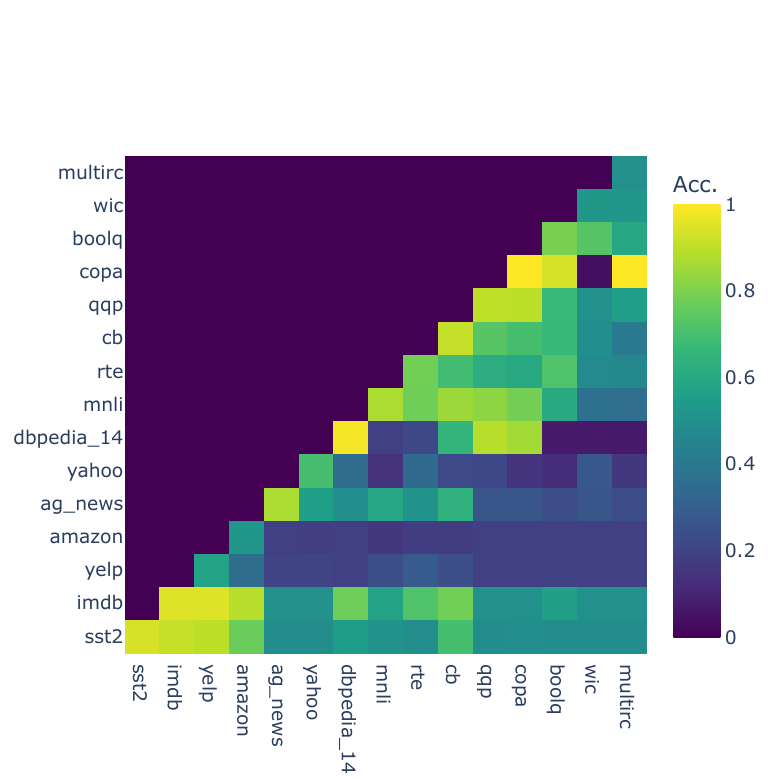}
        \caption*{\small (B) GRID}
    \end{minipage}
    \hfill
    \begin{minipage}{0.32\columnwidth}
        \centering
        \includegraphics[width=\linewidth]{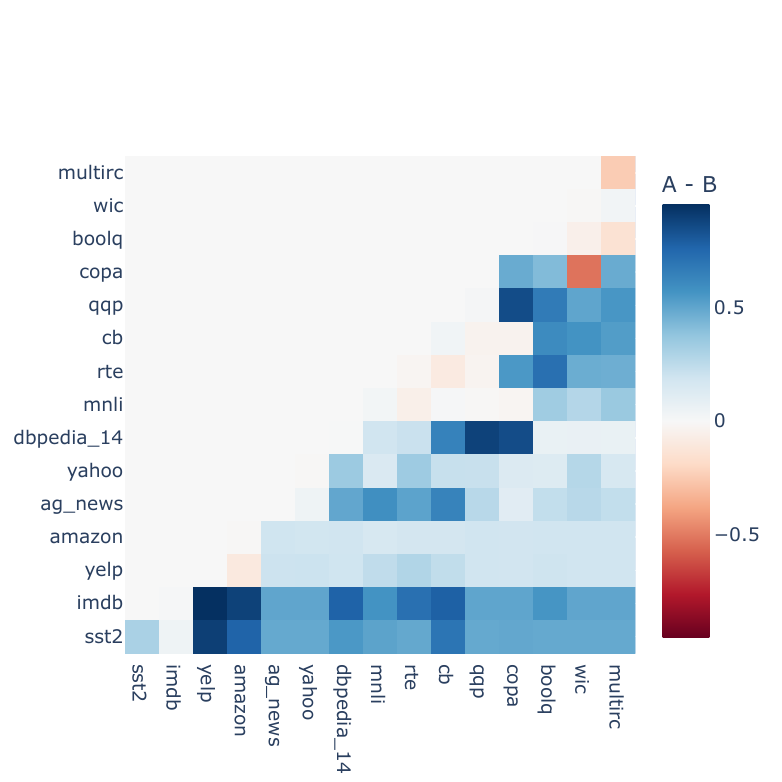}
        \caption*{\small (C) B--A}
    \end{minipage}

    \caption{\small
    Heatmaps of backward transfer scores on previous tasks for Order L4.
    Brighter values indicate better retention of earlier tasks.
    }
    \label{fig:oa_heat}
\end{figure}

\begin{figure}[t]
    \centering

    \begin{minipage}{0.32\columnwidth}
        \centering
        \includegraphics[width=\linewidth]{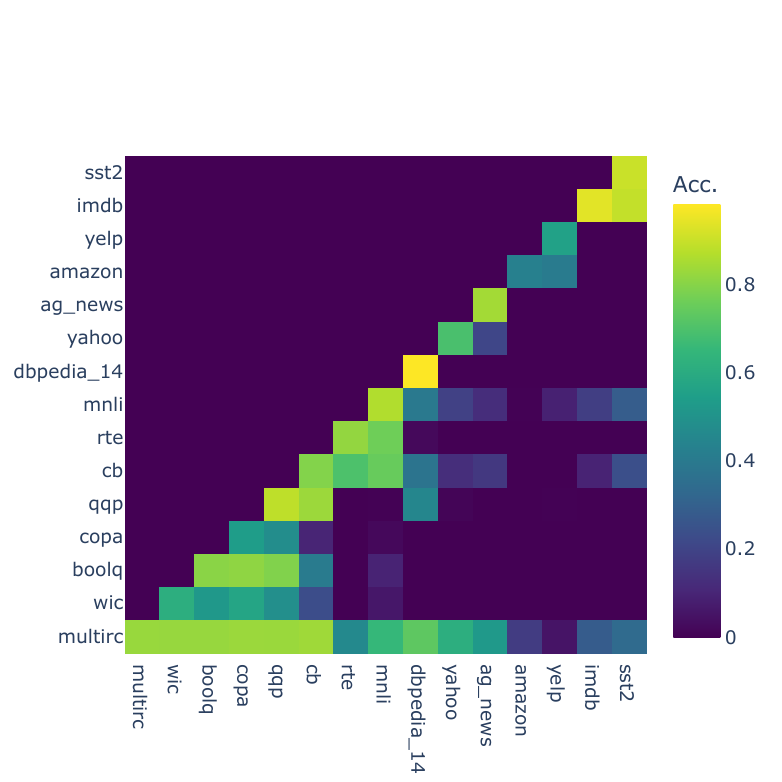}
        \caption*{\small (A) ProgPrompt}
    \end{minipage}
    \hfill
    \begin{minipage}{0.32\columnwidth}
        \centering
        \includegraphics[width=\linewidth]{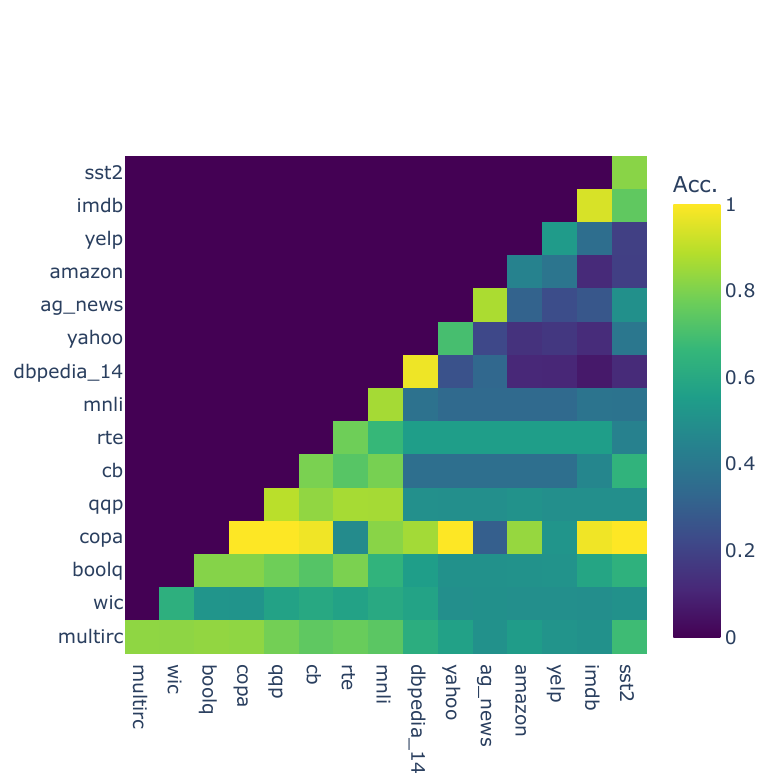}
        \caption*{\small (B) GRID}
    \end{minipage}
    \hfill
    \begin{minipage}{0.32\columnwidth}
        \centering
        \includegraphics[width=\linewidth]{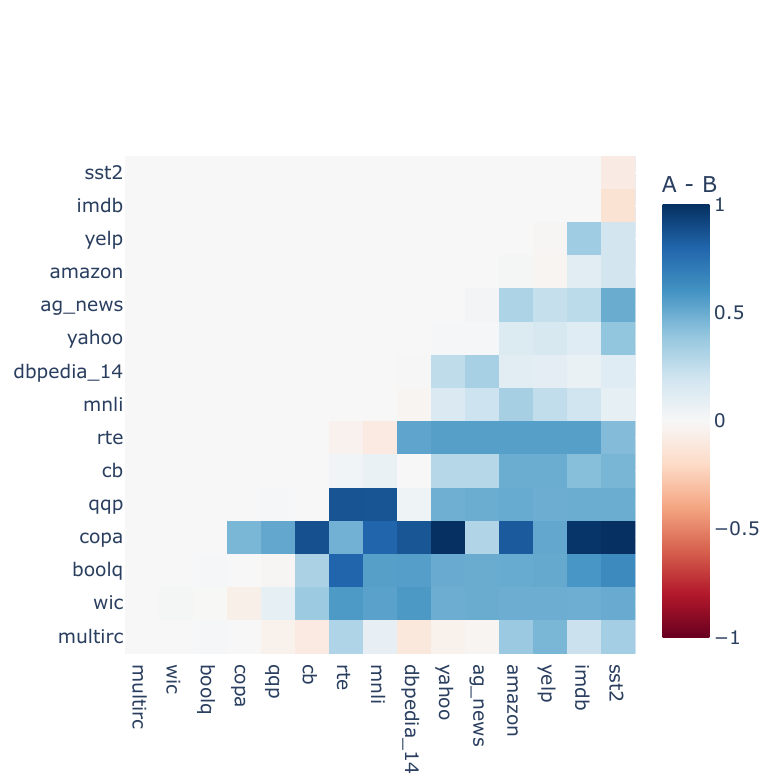}
        \caption*{\small (C) B--A}
    \end{minipage}

    \caption{\small
    Heatmaps of backward transfer scores on previous tasks for Order L5.
    Brighter values indicate better retention of earlier tasks.
    }
    \label{fig:ob_heat}
\end{figure}

\begin{figure}[t]
    \centering

    \begin{minipage}{0.32\columnwidth}
        \centering
        \includegraphics[width=\linewidth]{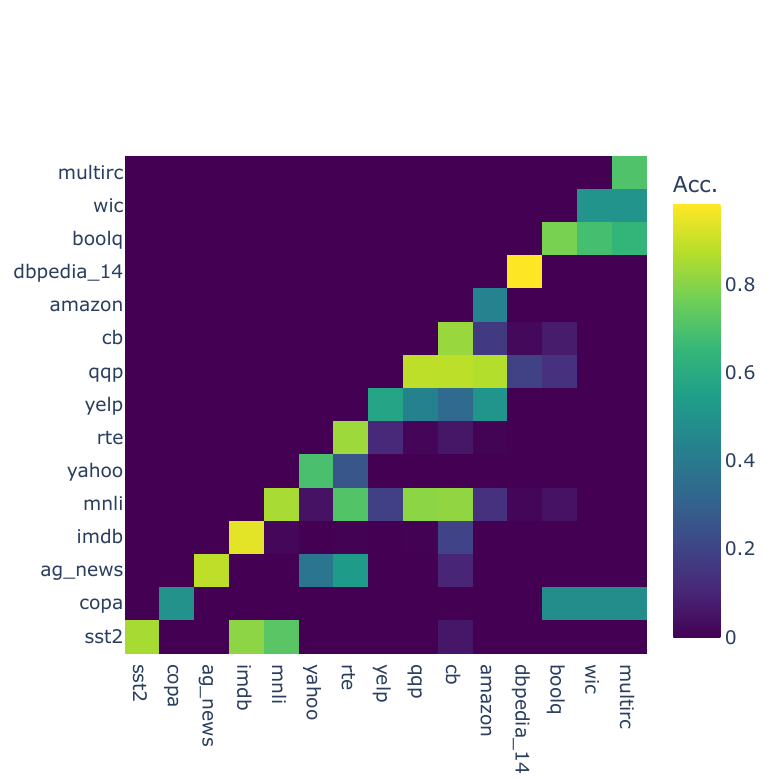}
        \caption*{\small (A) ProgPrompt}
    \end{minipage}
    \hfill
    \begin{minipage}{0.32\columnwidth}
        \centering
        \includegraphics[width=\linewidth]{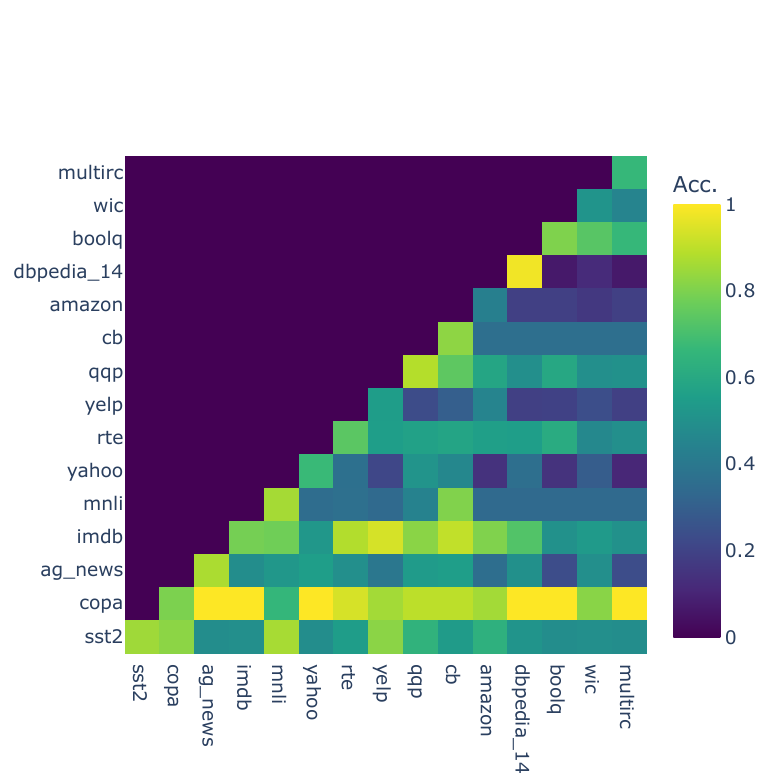}
        \caption*{\small (B) GRID}
    \end{minipage}
    \hfill
    \begin{minipage}{0.32\columnwidth}
        \centering
        \includegraphics[width=\linewidth]{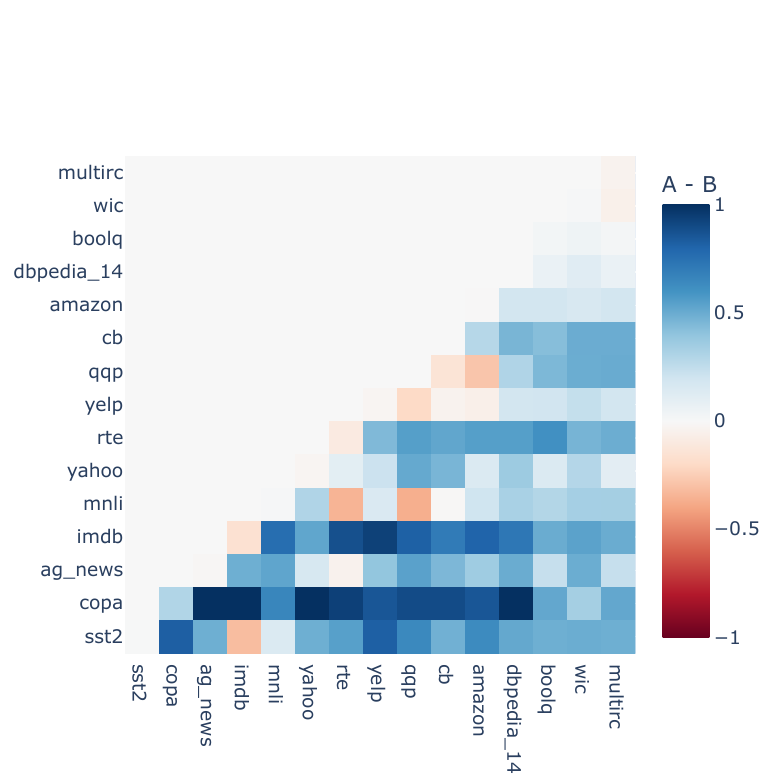}
        \caption*{\small (C) B--A}
    \end{minipage}

    \caption{\small
    Heatmaps of backward transfer scores on previous tasks for Order L6.
    Brighter values indicate better retention of earlier tasks.
    }
    \label{fig:oc_heat}
\end{figure}

\begin{figure}[!ht]
    \centering
    \small
    \includegraphics[width=0.45\textwidth, clip]{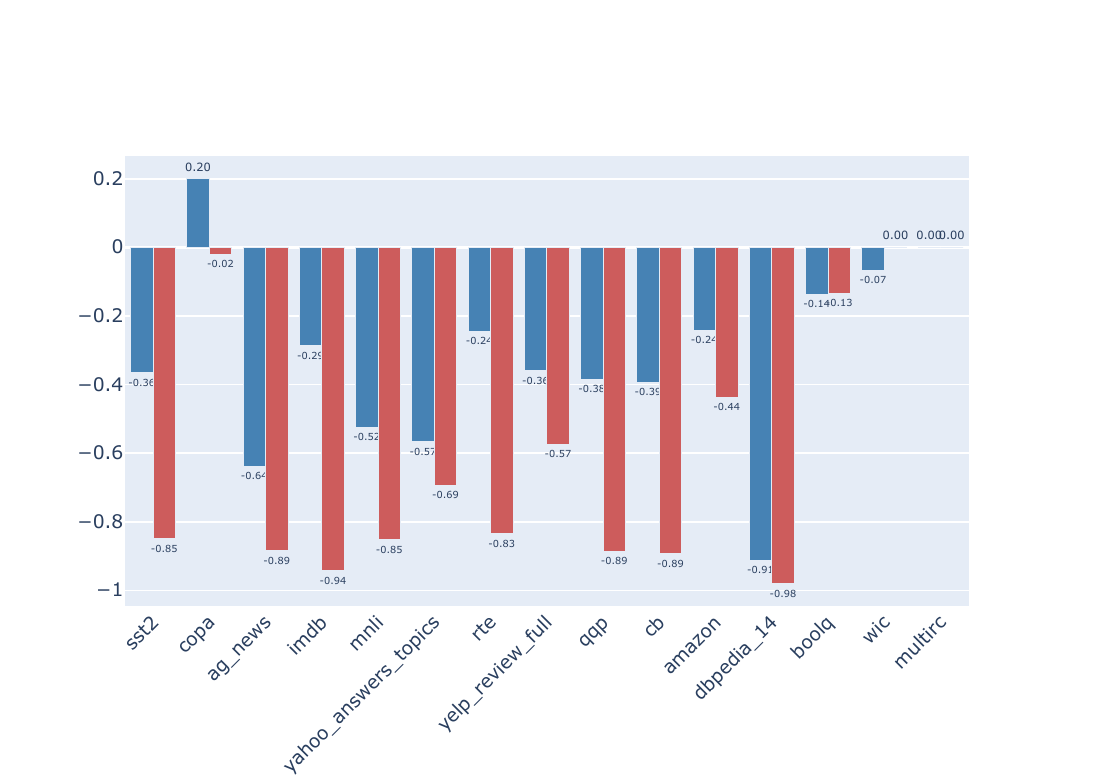}
    \caption{\small Per-task BWT comparison between our method (blue) and the ProgPrompt (red) for Order L6. Positive bars indicate improved retention of prior tasks.}
\label{fig:oc_bwt}
\end{figure}

We further observe that \textbf{Order L6 (mixed)} mostly outperforms L4 but remains slightly inferior to L5, 
suggesting that interleaving tasks of varying difficulty offers a compromise between stability and adaptability. 
This trend holds across model sizes, with larger models exhibiting higher absolute forgetting but benefiting more 
from favorable task curricula. Overall, these results confirm that task order plays a critical role in long-horizon 
task-agnostic continual learning, and that hard-to-easy curricula are particularly effective for mitigating forgetting 
under prompt-based adaptation. Figures~\ref{fig:bwt_l1_l2} -~\ref{fig:oc_bwt} provide complementary per-task BWT bar plots and heatmaps for Orders L2-L6.

\subsection{Forgotten Task Count Analysis}
\label{apdx:ftc}
We define the \emph{Forgotten Task Count (FTC)} as the number of tasks whose accuracy drops below a fixed fraction of their standalone performance. Formally, task $T_i$ is considered forgotten at step $t$ if $a_i^{(t)} < \tau \cdot \min_j a_j^{(0)}$, where $a_i^{(0)}$ denotes standalone accuracy and $\tau \in (0,1)$. Using zero accuracy as a cutoff is misleading, as performance may degrade to small but nonzero values; our criterion instead enforces a principled absolute threshold.\footnotemark  
GRID substantially reduces forgetting compared to ProgPrompt and SHLPT, forgetting on average only 13.8 tasks versus 78.7 and 64.0, respectively. These results demonstrate that GRID mitigates catastrophic forgetting in long-horizon continual learning while preserving task-relevant knowledge in a compact form. Results across model variants are reported in Table~\ref{tab:forgotten_tasks_comparison}.

\subsection{Forward transfer on long-sequence benchmarks.}
Table~\ref{tab:FWT} reports FWT on the long-sequence task orders L1--L6. Overall, GRID achieves better average FWT than Progressive Prompts across most model variants, improving the mean FWT from $-4.45$ to $-2.16$ across the six T5/Flan-T5 backbones. The gains are especially clear for Flan-T5-base, where FWT improves from $-4.92$ to $1.84$, and Flan-T5-large, where it improves from $-2.42$ to $-0.09$. These results indicate that GRID's reduction in backward forgetting does not come at the cost of forward transfer. Although T5-base shows lower FWT under GRID, the overall trend suggests that prompt-pool compression preserves useful transferable knowledge while reducing interference from redundant prompts.

\subsection{Analysis of Results on Negative Transfer Benchmarks.}
Tables~\ref{Table5}, \ref{Table6}, and \ref{tab:model_results_10} report BWT, accuracy, 
and FWT on the negative transfer sequences NT1--NT3, where tasks are deliberately 
dissimilar to test cross-task interference.

\textbf{BWT (Table~\ref{Table5}).}
GRID consistently and substantially reduces forgetting across all model scales. 
Improvements over PP range from ${\sim}51\%$ (T5-large) to ${\sim}72\%$ 
(Flan-T5-small), confirming that constrained decoding and gradient-guided compression 
together suppress interference even when tasks share little semantic overlap.

\textbf{Accuracy (Table~\ref{Table6}).}
Results are competitive but mixed. GRID outperforms PP on Flan-T5-small ($+1.6\%$), 
Flan-T5-base ($+1.2\%$), and matches closely on most others. PP holds a marginal edge 
on T5-base ($-1.1\%$) and T5-large ($-0.4\%$). The pattern suggests that GRID's 
accuracy advantage is less pronounced under high task dissimilarity compared to 
long-sequence settings, likely because constrained decoding provides smaller gains 
when task boundaries are already well-separated.

\textbf{FWT (Table~\ref{tab:model_results_10}).}
Forward transfer is similarly mixed. GRID improves FWT on Flan-T5-small 
($-5.34$ vs.\ $-7.27$) and Flan-T5-base ($-5.07$ vs.\ $-6.44$), but underperforms 
PP on T5-base and T5-large. This indicates that prompt pool compression occasionally 
discards prompts with transferable signals in low-similarity settings.

\subsection{Extension to Open-Ended Generation (SuperNI Benchmark).}

We evaluate GRID on the SuperNI benchmark described in Table~\ref{tab:superni_orders} and Table~\ref{tab:superni_details},
which contains 15 instruction-following tasks spanning dialogue generation,
information extraction, question answering, summarization, and sentiment analysis.
Unlike the fixed-label classification benchmarks used in the main experiments,
most SuperNI tasks require open-ended generation and are evaluated with Rouge-L,
while sentiment tasks are evaluated using accuracy.

As shown in Table~\ref{tab:superni_openended}, GRID improves both average accuracy and BWT over ProgPrompt and SHLPT across two task orders, suggesting that its gains are not solely due to constrained decoding but also arise from reducing cross-task prompt interference. These results suggest that GRID is not limited to fixed-label classification settings: even when constrained label decoding is not applicable, prompt-pool compression improves retention by reducing cross-task prompt interference.

\subsection{Results on MTCL-Bench}
Table~\ref{tab:mtcl_bench_results} shows that MTCL-Bench is a challenging continual learning benchmark across mathematical reasoning and tool-selection tasks. Although ProgPrompt achieves slightly higher average accuracy on Qwen, GRID consistently provides better task retention across both backbones, as shown by less negative BWT and less forgetting. On LLaMA, GRID improves average accuracy over ProgPrompt while reducing forgetting by a large margin, with BWT improving from $-35.09$ to $-25.34$. On Qwen, GRID also improves retention, reducing BWT from $-6.12$ to $-5.14$. These results suggest that GRID's selective prompt compression better preserves previously learned task knowledge than naive prompt accumulation, leading to stronger backward transfer and more stable long-horizon retention.

\begin{figure}[t]
    \centering
\includegraphics[width=0.45\textwidth]{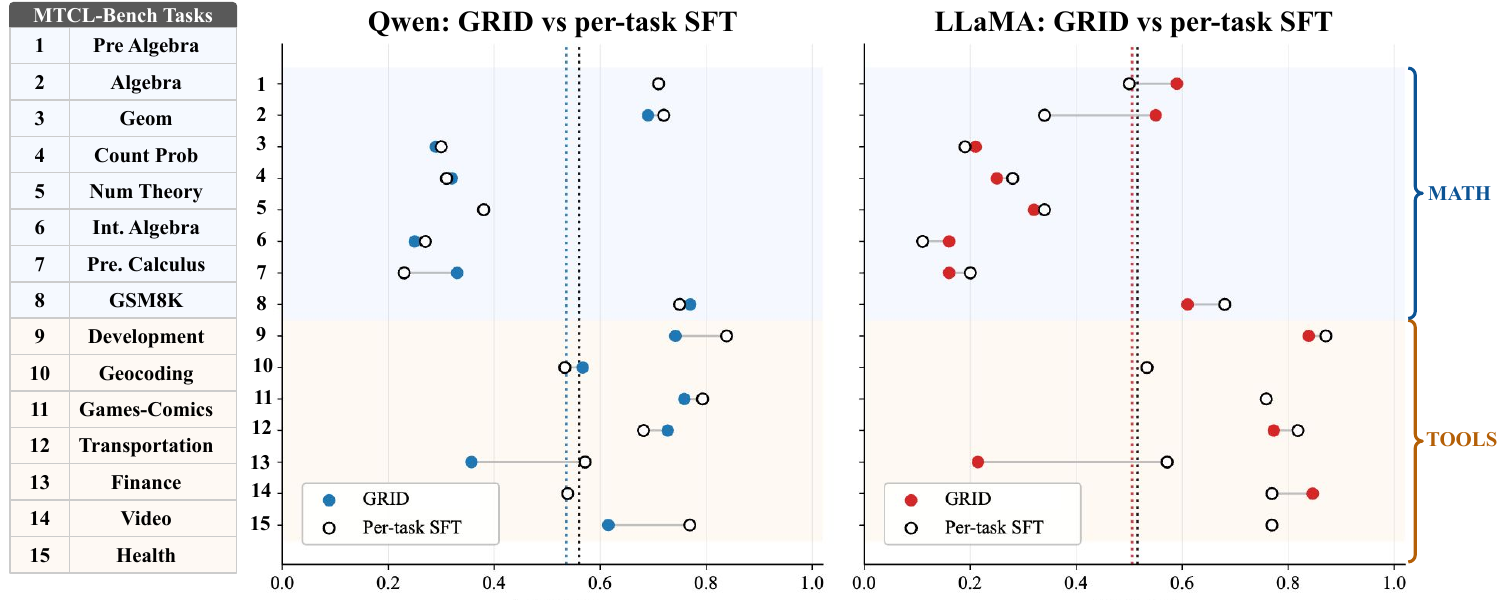}
    % \vspace{-0.4cm}

    \caption{\small Task-level comparison between GRID and per-task LoRA SFT on MTCL-Bench. Each point reports task-level accuracy for Qwen and LLaMA, with shaded regions separating math and tool-use tasks. Per-task LoRA SFT trains an independent LoRA adapter for each task and serves as a task-specific reference, while GRID uses bounded-memory continual soft-prompt learning.}
\label{fig:mtcl_taskwise}
\end{figure}

\paragraph{Comparison with per-task LoRA SFT.}
Figure~\ref{fig:mtcl_taskwise} compares GRID with a per-task LoRA SFT reference on MTCL-Bench. Unlike per-task LoRA SFT, which trains an independent LoRA adapter for each task, GRID learns continually using a bounded soft-prompt pool. GRID remains competitive on several tasks, while the remaining gaps highlight cases where continual prompt learning still suffers from task interference.

\subsection{Hyperparameter Sensitivity: Prompt Selection Threshold $\alpha$.}
\label{apdx:alpha}
The hyperparameter $\alpha$ controls the aggressiveness of gradient-based prompt selection, and therefore determines both the number of retained prompts and the prompt memory footprint. As shown in Table~\ref{tab:alpha_sensitivity}, GRID exhibits a stable memory--performance trade-off. Smaller values of $\alpha$ retain more prompts, leading to slightly better retention but higher memory usage, whereas larger values increase compression with gradual degradation. Importantly, a broad range of $\alpha \in [0.6, 0.8]$ consistently retains about five prompts ($\sim$200KB) and achieves near-optimal BWT, indicating that GRID is robust and does not require precise tuning. Notably, this retained capacity also matches the best queue size observed in our queue-size analysis, suggesting that gradient-based selection can automatically identify an effective memory budget.

\begin{table}[t]
\centering
\tiny
\setlength{\tabcolsep}{7pt}
\renewcommand{\arraystretch}{1.1}
\begin{tabular}{c|c|c|c}
\toprule
\rowcolor{gray!25}
$\alpha$ & Retained Prompts & BWT $\uparrow$ & Prompt Mem. (KB) $\downarrow$ \\
\midrule
0.0 & 7 & -0.40 & 280 \\
0.2 & 6 & -0.35 & 240 \\
0.4 & 6 & -0.35 & 240 \\
0.6 & 5 & \textbf{-0.34} & \textbf{200} \\
0.8 & 5 & \textbf{-0.34} & \textbf{200} \\
1.0 & 4 & -0.38 & 160 \\
1.2 & 4 & -0.38 & 160 \\
1.4 & 3 & -0.45 & 120 \\
1.6 & 3 & -0.45 & 120 \\
\bottomrule
\end{tabular}
\caption{Sensitivity to the prompt-selection threshold $\alpha$. GRID remains stable over a broad range of $\alpha$, with $\alpha=0.6$--$0.8$ providing the best memory--BWT trade-off.}
\label{tab:alpha_sensitivity}
\end{table}

\begin{table*}[t]
  \centering
  \tiny
  \setlength{\tabcolsep}{4pt}
  \renewcommand{\arraystretch}{1.1}

  \begin{tabular}{l l c c c c c c c c}
    \toprule
    \rowcolor{gray!25}
    \textbf{Model} & \textbf{Setting} &
    \textbf{L1} & \textbf{L2} & \textbf{L3} &
    \textbf{L4} & \textbf{L5} & \textbf{L6} &
    \textbf{Avg} & \textbf{Imp. (\%)} \\
    \midrule

    \multirow{2}{*}{T5-small}
      & PP & -0.6081 & -0.5414 & -0.5126 & -0.5243 & -0.5377 & -0.5006 & -0.5374 & \multirow{2}{*}{\textbf{54.4}$\uparrow$} \\
      & GRID       & -0.1402 & -0.1894 & -0.2504 & -0.3141 & -0.2734 & -0.3006 & \textbf{-0.2447} & \\

    \midrule
    \multirow{2}{*}{FT5-small}
      & PP & -0.5711 & -0.5574 & -0.4916 & -0.4914 & -0.5075 & -0.4987 & -0.5196 & \multirow{2}{*}{\textbf{68.4}$\uparrow$} \\
      & GRID       & -0.0518 & -0.1449 & -0.2106 & -0.2129 & -0.1517 & -0.2396 & \textbf{-0.1686} & \\

    \midrule
    \multirow{2}{*}{T5-base}
      & PP & -0.6918 & -0.6520 & -0.5948 & -0.5956 & -0.6471 & -0.5881 & -0.6282 & \multirow{2}{*}{\textbf{52.7}$\uparrow$} \\
      & GRID       & -0.3060 & -0.2404 & -0.3549 & -0.2893 & -0.3439 & -0.3259 & \textbf{-0.3101} & \\

    \midrule
    \multirow{2}{*}{FT5-base}
      & PP & -0.6092 & -0.7222 & -0.6488 & -0.6273 & -0.6631 & -0.6225 & -0.6489 & \multirow{2}{*}{\textbf{58.7}$\uparrow$} \\
      & GRID       & -0.2616 & -0.3243 & -0.3319 & -0.3245 & -0.3240 & -0.2743 & \textbf{-0.3067} & \\

    \midrule
    \multirow{2}{*}{T5-large}
      & PP & -0.7275 & -0.7625 & -0.6137 & -0.6257 & -0.6351 & -0.6416 & -0.6677 & \multirow{2}{*}{\textbf{51.7}$\uparrow$} \\
      & GRID       & -0.3243 & -0.3336 & -0.3979 & -0.3912 & -0.2956 & -0.3512 & \textbf{-0.3490} & \\

    \midrule
    \multirow{2}{*}{FT5-large}
      & PP & -0.7679 & -0.7444 & -0.6195 & -0.6616 & -0.7107 & -0.6631 & -0.6945 & \multirow{2}{*}{\textbf{50.4}$\uparrow$} \\
      & GRID       & -0.3115 & -0.3656 & -0.3544 & -0.3514 & -0.3282 & -0.3540 & \textbf{-0.3442} & \\

    \bottomrule
  \end{tabular}
    \caption{\small Backward Transfer (BWT) comparison between our method and ProgPrompt across multiple task orderings (Orders L1--L6) for T5 and Flan-T5 models. Less negative scores indicate reduced forgetting. Our method consistently improves BWT across all model variants.}
  \label{tab:bwt_comparison}
\end{table*}

\begin{table}[htbp]
\centering
\tiny
\setlength\tabcolsep{6pt}
\renewcommand{\arraystretch}{1.15}

\begin{tabular}{c c c c}
\toprule
\rowcolor{gray!25}
\tiny
\textbf{k} & \textbf{DBPedia} & \textbf{Amazon} & \textbf{AG News} \\ 
\midrule
20   & 0.5203 & 0.0000 & 0.0000 \\
200  & 0.9674 & 0.0000 & 0.8180 \\
1000 & 0.9880 & 0.5136 & 0.8900 \\
2000 & 0.9892 & 0.5534 & --     \\
\bottomrule
\end{tabular}
\caption{\small Accuracy for different datasets with varying sample sizes (k) highlights how model performance scales with more training samples.}
\label{samples}

\end{table}

\subsection{Detailed Ablation Study}

\label{sec:detailed_abalation}

To evaluate the contribution of each component in the GRID framework, we conduct an ablation study on task order L1, L2, and L3, measuring BWT under four variants: (0) the full \textbf{G.R.I.D.} model, (1) without gradient-based prompt selection (G), (2) without constrained decoding (D), (3) without representative input selection (R), and (4) without all components. As shown in Table~\ref{ablation}, the full model yields the highest BWT for both T5-large (\textbf{-0.3511}) and Qwen-3-4B (\textbf{-0.4418}). Excluding gradient-based selection has minimal effect on BWT, consistent with its role in reducing memory rather than directly improving retention: GRID requires only \textbf{200 KB} of storage versus \textbf{600 KB} for ProgPrompt, a \textbf{66.7\% reduction}. In contrast, removing constrained decoding causes a pronounced BWT drop, and eliminating both components leads to the worst performance (\textbf{-0.7012} for T5-large, \textbf{-0.8114} for Qwen-3-4B). Overall, these results underscore constrained decoding as the primary driver of backward transfer, with gradient-based selection offering complementary scalability.

\subsubsection{Impact of Prompt Selection Strategies (Gradient vs. FIFO vs. Random).} 
As shown in Table~\ref{compact_strategy_comparison}, our gradient-based method consistently yields the fewest forgotten tasks and achieves competitive or superior BWT across all settings. FIFO occasionally matches our BWT but retains more redundant prompts, resulting in higher forgetting. Random selection performs worst, with unstable BWT and the largest number of forgotten tasks.

\subsubsection{Effect of Sample Size and Clustering on Performance .}
We empirically find that increasing the number of representative samples beyond 1k per class 
leads to accuracy saturation with only negligible improvements. 
This indicates that larger sample sizes may not yield substantial performance gains 
(Table~\ref{samples}). 
By contrast, clustering provides both data efficiency and better generalization, 
highlighting the importance of representative input selection in GRID.

\begin{table}[t]
\centering
\tiny

\setlength{\tabcolsep}{4pt}
\renewcommand{\arraystretch}{1.1}

\begin{tabular}{c|ccc|ccc|ccc}
\toprule
\textbf{PL} &
\multicolumn{3}{c|}{\textbf{L1}} &
\multicolumn{3}{c|}{\textbf{L2}} &
\multicolumn{3}{c}{\textbf{L3}} \\
\cmidrule(lr){2-4} \cmidrule(lr){5-7} \cmidrule(lr){8-10}
& \textbf{Acc} & \textbf{BWT} & \textbf{FT}
& \textbf{Acc} & \textbf{BWT} & \textbf{FT}
& \textbf{Acc} & \textbf{BWT} & \textbf{FT} \\
\midrule
5  & 78.23 & -0.3167 & 10 & 81.73 & -0.3565 & 5 & 84.98 & -0.4286 & 15 \\
10 & 79.12 & -0.3243 & 11 & 80.69 & -0.3336 & 5 & 81.09 & -0.3979 & 18 \\
20 & 80.58 & -0.3396 & 12 & 79.37 & -0.3159 & 5 & 75.55 & -0.3206 & 24 \\
\bottomrule
\end{tabular}
\caption{\small Impact of prompt length (PL) on T5-large performance across Order L1--L3. We observe a trade-off: longer prompts improve BWT (less forgetting) but increase the number of forgotten tasks and can degrade accuracy. Optimal performance is generally observed at a moderate prompt length (e.g., 10).}
\label{tab:prompt_length}
\end{table}

\subsubsection{Effect of Prompt Length.}
We analyze the impact of prompt length (PL) on performance in Table~\ref{tab:prompt_length} using the T5-large backbone across task orders L1--L3. We observe a clear trade-off between accuracy, backward transfer (BWT), and forgetting. Short prompts (PL=5) yield competitive accuracy but suffer from increased forgetting, while longer prompts (PL=20) improve BWT (i.e., reduce forgetting) at the cost of degraded accuracy and a higher number of forgotten tasks, likely due to over-parameterization and prompt interference. A moderate prompt length (PL=10) consistently provides the best balance across metrics, achieving strong accuracy with stable BWT and controlled forgetting. Based on these observations, we fix the prompt length to 10 tokens in all experiments.

\subsubsection{Effect of Queue Size.}
\begin{figure}[ht]
    \centering
\includegraphics[width=0.45\textwidth]{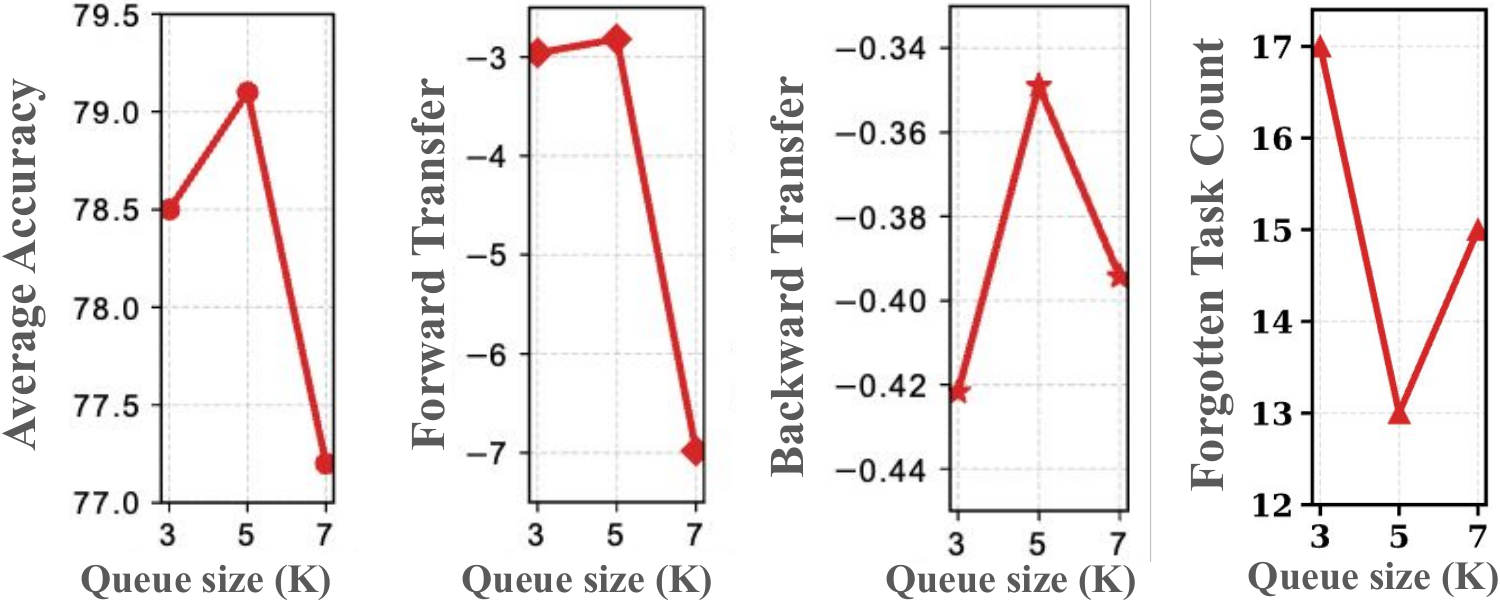}
% \vspace{-0.4cm}

    \caption{\small Effect of queue size
K. 
% \kj{make it bigger; it wastes the space a little bit}
}
    \label{fig:K}
\end{figure}

Figure~\ref{fig:K} shows that moderate queue sizes (e.g., $K{=}5$) provide the best trade-off between accuracy, forward transfer, and backward transfer, while larger queues degrade performance due to redundant and weakly task-relevant prompts. 

\section{ Detailed Training and Inference Overhead Analysis.}
\label{apdx:overhead}

\subsection{Complexity analysis }
GRID introduces an additional gradient-scoring step for prompt selection. Importantly, this scoring is performed only once per retained prompt when a new task arrives, rather than during the main prompt-training iterations. For a new task $T_t$, the scoring overhead is $O(|P_t|)$, where $|P_t|$ denotes the size of the current prompt pool. Across a sequence of $N$ tasks, the total scoring overhead is therefore $O(\sum_{t=1}^{N} |P_t|)$. In practice, GRID maintains a compact prompt pool of approximately five prompts, so the overhead scales nearly linearly with the number of tasks.

\subsection{Runtime and Hardware.}  
In addition to the A100 experiments, we evaluated GRID on H100 GPUs (80GB) with batch size $16$. On H100, long task sequences averaged $\sim 11$ hours per run with peak memory usage of $\sim 65$ GB, compared to $\sim 27$ hours and $\sim 21$ GB on A100 (40GB, batch size $8$). These results demonstrate that GRID scales efficiently across hardware generations.

\subsection{Training overhead analysis}
Table~\ref{tab:overhead_analysis} reports the empirical overhead on a 15-task sequence. The gradient-scoring step requires approximately 2.5--3.1 minutes per prompt. Since GRID retains about five prompts, the additional overhead remains around 12.9--15.6 minutes per task, totaling only 144.45 minutes across Tasks 6--15. Overall training time remains comparable to ProgPrompt: GRID requires about 27 hours compared with about 24 hours for ProgPrompt on an A100 with batch size 8. On an H100 with batch size 16, GRID requires about 12 hours compared with about 11 hours for ProgPrompt. These results show that prompt compression introduces only modest training overhead while substantially reducing prompt memory.

\begin{table}[t]
\centering
\tiny
\setlength{\tabcolsep}{5pt}
\renewcommand{\arraystretch}{1.08}
\begin{tabular}{c|l|c|c|c|c}
\toprule
\rowcolor{gray!25}
\textbf{Task} \# & \textbf{Task} & \textbf{$|P_t|$} & \textbf{min/prompt} & \textbf{Overhead (min)} & \textbf{Cum. (min}) \\
\midrule
6  & boolq   & 5 & 2.85 & 14.25 & 14.25 \\
7  & rte     & 5 & 2.92 & 14.60 & 28.85 \\
8  & imdb    & 5 & 2.58 & 12.90 & 41.75 \\
9  & yelp    & 5 & 2.95 & 14.75 & 56.50 \\
10 & amazon  & 5 & 3.11 & 15.55 & 72.05 \\
11 & sst2    & 5 & 2.88 & 14.40 & 86.45 \\
12 & dbpedia & 5 & 2.90 & 14.50 & 100.95 \\
13 & agnews  & 5 & 2.84 & 14.20 & 115.15 \\
14 & multirc & 5 & 2.97 & 14.85 & 130.00 \\
15 & yahoo   & 5 & 2.89 & 14.45 & 144.45 \\
\bottomrule
\end{tabular}
\caption{Empirical gradient-scoring overhead for GRID on a 15-task sequence. The scoring step is performed once per retained prompt when a new task arrives. Since GRID maintains a compact prompt pool of approximately five prompts, the additional overhead remains modest.}
\label{tab:overhead_analysis}
\end{table}

\subsection{Inference efficiency.}
At inference time, GRID uses a substantially smaller prompt pool than prompt-accumulation baselines. For example, after 15 tasks, ProgPrompt retains 15 task-specific prompts, whereas GRID retains approximately five prompts. This reduces the number of prompt tokens prepended to each input. For long inputs with sequence length around 2048, this yields a modest inference speedup of about 5--20\%. For shorter inputs with sequence length around 264, where prompts constitute a larger fraction of the total input length, inference becomes about 30--50\% faster. Thus, although GRID introduces a small one-time scoring overhead during training, it improves memory efficiency and reduces inference cost throughout deployment.

% \noindent\textbf{Remark (Gradient-score intuition).}
% For a prompt $\mathbf{p}_j$, let $L_t(\mathbf{p}_j)$ denote the current-task loss. 
% A first-order Taylor approximation gives
% \begin{equation}
% L_t(\mathbf{p}_j+\Delta \mathbf{p}_j)-L_t(\mathbf{p}_j)
% \approx
% \left\langle 
% \nabla_{\mathbf{p}_j}L_t(\mathbf{p}_j), 
% \Delta \mathbf{p}_j
% \right\rangle .
% \end{equation}
% By Cauchy--Schwarz,
% \begin{equation}
% \left|
% L_t(\mathbf{p}_j+\Delta \mathbf{p}_j)-L_t(\mathbf{p}_j)
% \right|
% \leq
% \|\nabla_{\mathbf{p}_j}L_t(\mathbf{p}_j)\|_2
% \|\Delta \mathbf{p}_j\|_2 .
% \end{equation}
% Thus, if $g_j=\|\nabla_{\mathbf{p}_j}L_t(\mathbf{p}_j)\|_2$ is small, changing or merging $\mathbf{p}_j$ has limited first-order effect on the current-task loss. This provides an intuitive justification for using gradient norm as a prompt relevance score.

% By Cauchy--Schwarz,
% \begin{equation}
% \begin{aligned}
% &\left|
% L_t(\mathbf{p}_j+\Delta \mathbf{p}_j)-L_t(\mathbf{p}_j)
% \right| \\
% &\qquad \leq
% \|\nabla_{\mathbf{p}_j}L_t(\mathbf{p}_j)\|_2
% \|\Delta \mathbf{p}_j\|_2 .
% \end{aligned}
% \end{equation}
% Thus, if $g_j=\|\nabla_{\mathbf{p}_j}L_t(\mathbf{p}_j)\|_2$ is small, changing or merging $\mathbf{p}_j$ has limited first-order effect on the current-task loss. This provides an intuitive justification for using gradient norm as a prompt relevance score.

\begin{table}[t]
\centering
\tiny
\setlength{\tabcolsep}{7pt}
\renewcommand{\arraystretch}{1.12}
\begin{tabular}{l|cc|cc}
\toprule
\multirow{2}{*}{\textbf{Method}} 
& \multicolumn{2}{c|}{\textbf{O1}} 
& \multicolumn{2}{c}{\textbf{O2}} \\
\cline{2-5}
& \textbf{Avg. Acc.} $\uparrow$ & \textbf{BWT} $\uparrow$ 
& \textbf{Avg. Acc.} $\uparrow$ & \textbf{BWT} $\uparrow$ \\
\midrule
ProgPrompt & 43.65 & -37.27 & 43.98 & -41.65 \\
SHLPT      & 44.97 & -32.20 & 46.97 & -33.97 \\
GRID       & \textbf{46.54} & \textbf{-31.15} & \textbf{48.12} & \textbf{-31.20} \\
\bottomrule
\end{tabular}
\caption{Results on open-ended instruction-based tasks from SuperNI across two task orders. Unlike fixed-label classification tasks, these tasks require open-ended generation. GRID improves both average accuracy and backward transfer (BWT), indicating that its gains are not limited to constrained label-space settings.}
\label{tab:superni_openended}
\end{table}

\begin{table}[t]
  \centering

  \tiny
  \setlength\tabcolsep{4pt}
  \begin{tabular}{@{}l l c c c c c c c@{}}
    \toprule
    \rowcolor{gray!25}
    \textbf{Model} & \textbf{Setting} & \textbf{L1} & \textbf{L2} & \textbf{L3} & \textbf{L4} & \textbf{L5} & \textbf{L6} & \textbf{Avg} \\
    \midrule
    \multirow{2}{*}{T5-small}  & PP & 63.18 & 57.18 & 61.12 & 62.73 & 59.11 & 60.33 & 60.61 \\
                               & GRID     & 59.75 & 64.97 & 59.90 & 64.64 & 64.51 & 62.39 & \textbf{62.69} \\
    \midrule
    \multirow{2}{*}{FT5-small} & PP & 60.32 & 59.25 & 59.43 & 59.18 & 59.35 & 59.96 & 59.58 \\
                               & GRID     & 58.35 & 57.37 & 64.73 & 61.74 & 58.30 & 57.07 & 59.59 \\
    \midrule
    \multirow{2}{*}{T5-base}   & PP & 71.34 & 67.60 & 69.57 & 69.71 & 72.43 & 69.93 & 70.10 \\
                               & GRID     & 72.74 & 71.24 & 72.09 & 67.16 & 75.72 & 65.23 & 70.70 \\
    \midrule
    \multirow{2}{*}{FT5-base}  & PP & 63.73 & 74.83 & 75.23 & 72.14 & 74.25 & 72.48 & 72.11 \\
                               & GRID     & 77.24 & 78.08 & 77.16 & 78.32 & 73.53 & 65.92 & \textbf{75.04} \\
    \midrule
    \multirow{2}{*}{T5-large}  & PP & 75.68 & 78.56 & 74.29 & 75.05 & 77.10 & 75.46 & 76.02 \\
                               & GRID     & 79.12 & 80.69 & 81.09 & 79.05 & 79.79 & 75.54 & \textbf{79.21} \\
    \midrule
    \multirow{2}{*}{FT5-large} & PP & 79.81 & 77.51 & 74.30 & 79.13 & 79.09 & 79.56 & \textbf{78.23} \\
                               & GRID     & 79.77 & 78.25 & 76.09 & 76.91 & 76.94 & 76.74 & 77.45 \\
    \bottomrule
  \end{tabular}
  \caption{\small Extensive comparison of \textbf{GRID} (ours) with Progressive Prompts on seq-to-seq backbones (T5 and Flan-T5) across long-sequence task orders (L1--L6), reporting average test accuracy.} 
  \label{tab:enc-dec}
\end{table}

\begin{table}[ht]
  \centering

  % \vspace{0.1cm}
  \tiny
  \setlength\tabcolsep{4pt}
  \begin{tabular}{@{}l l c c c c c c c c@{}}
    \toprule
    \rowcolor{gray!25}
    \textbf{Model} & \textbf{Setting} 
      & \textbf{L1} & \textbf{L2} & \textbf{L3} 
      & \textbf{L4} & \textbf{L5} & \textbf{L6} 
      & \textbf{Average} 
      & \textbf{Imp (\%)} \\
    \midrule
    \multirow{2}{*}{T5s}  
      & PP 
        & 85 & 85 & 88 & 90 & 84 & 93 
        & 87.5 
        & \multirow{2}{*}{\textbf{76.8\%}} \\
      & GRID       
        & \textbf{14} & \textbf{7}  & \textbf{28} & \textbf{36} & \textbf{16} & \textbf{21} 
        & \textbf{20.3} 
        &  \\ 
    \midrule
    \multirow{2}{*}{FT5s} 
      & PP 
        & 93 & 95 & 95 & 94 & 97 & 93 
        & 94.5 
        & \multirow{2}{*}{\textbf{76.7\%}} \\
      & GRID       
        & \textbf{17} & \textbf{14} & \textbf{28} & \textbf{26} & \textbf{20} & \textbf{27} 
        & \textbf{22.0} 
        &  \\ 
    \midrule
    \multirow{2}{*}{T5b}   
      & PP 
        & 88 & 82 & 76 & 90 & 83 & 88 
        & 84.5 
        & \multirow{2}{*}{\textbf{80.9\%}} \\
      & GRID       
        & \textbf{15} & \textbf{6}  & \textbf{19} & \textbf{29} & \textbf{13} & \textbf{15} 
        & \textbf{16.2} 
        &  \\ 
    \midrule
    \multirow{2}{*}{FT5b}  
      & PP 
        & 94 & 93 & 93 & 93 & 93 & 93 
        & 93.2 
        & \multirow{2}{*}{\textbf{83.0\%}} \\
      & GRID       
        & \textbf{10} & \textbf{10} & \textbf{15} & \textbf{25} & \textbf{14} & \textbf{21} 
        & \textbf{15.8} 
        &  \\ 
    \midrule
    \multirow{2}{*}{T5l}  
      & PP 
        & 77 & 72 & 80 & 87 & 71 & 85 
        & 78.7 
        & \multirow{2}{*}{\textbf{82.4\%}} \\
      & GRID       
        & \textbf{11} & \textbf{5}  & \textbf{18} & \textbf{26} & \textbf{10} & \textbf{13} 
        & \textbf{13.8} 
        &  \\ 
    \midrule
    \multirow{2}{*}{FT5l} 
      & PP 
        & 92 & 93 & 93 & 93 & 91 & 93 
        & 92.5 
        & \multirow{2}{*}{\textbf{79.8\%}} \\
      & GRID       
        & \textbf{15} & \textbf{10} & \textbf{24} & \textbf{24} & \textbf{14} & \textbf{25} 
        & \textbf{18.7} 
        &  \\ 
    \bottomrule
  \end{tabular}
    \caption{ \small Comparison of forgotten task counts between Progressive Prompts and our method for different models across various task orders (order L1–L6).}
  \label{tab:forgotten_tasks_comparison}
  % \vspace{-0.5cm}
\end{table}

\begin{table}[t]
\centering

\tiny
\setlength{\tabcolsep}{3pt}
\renewcommand{\arraystretch}{1.08}

\begin{tabular}{@{}l l r r r r@{}}
\toprule
\rowcolor{gray!25}
\textbf{Model} & \textbf{Method} & \textbf{NT1} & \textbf{NT2} & \textbf{NT3} & \textbf{Avg} \\
\midrule
T5-small  & PP & -0.5783 & -0.4882 & -0.5133 & -0.5266 \\
          & GRID       & -0.1957 & -0.1945 & -0.2043 & -0.1982 \\
\midrule
Flan-T5-small & PP & -0.4771 & -0.3924 & -0.4097 & -0.4264 \\
              & GRID       & -0.0807 & -0.1601 & -0.1171 & -0.1193 \\
\midrule
T5-base   & PP & -0.5707 & -0.5834 & -0.5890 & -0.5810 \\
          & GRID       & -0.1817 & -0.2103 & -0.2314 & -0.2078 \\
\midrule
Flan-T5-base  & PP & -0.6719 & -0.5335 & -0.6116 & -0.6057 \\
              & GRID       & -0.2911 & -0.2074 & -0.3162 & -0.2716 \\
\midrule
T5-large  & PP & -0.7088 & -0.6598 & -0.6536 & -0.6741 \\
          & Ours       & -0.3154 & -0.3658 & -0.3077 & -0.3296 \\
\midrule
Flan-T5-large & PP & -0.7156 & -0.6628 & -0.6886 & -0.6890 \\
              & GRID       & -0.2858 & -0.2888 & -0.3636 & -0.3127 \\
\bottomrule
\end{tabular}
\caption{\small Backward Transfer (BWT) on negative transfer benchmarks (NT1--NT3). Less negative is better.}
\label{Table5}
\end{table}

\begin{table}[t]
  \centering

  \tiny
  \setlength\tabcolsep{4pt}
  \begin{tabular}{@{}l l c c c c@{}}
    \toprule
    \rowcolor{gray!25}
    \textbf{Model} & \textbf{Setting} & \textbf{NT1} & \textbf{NT2} & \textbf{NT3} & \textbf{Avg} \\
    \midrule
    \multirow{2}{*}{T5-small}  & PP & 57.22 & 55.51 & 57.79 & 56.84 \\
                               & GRID     & 58.21 & 55.08 & 56.44 & 56.58 \\
    \midrule
    \multirow{2}{*}{FT5-small} & PP & 51.57 & 46.02 & 48.36 & 48.65 \\
                               & GRID     & 50.40 & 50.54 & 49.77 & 50.24 \\
    \midrule
    \multirow{2}{*}{T5-base}   & PP & 61.18 & 65.75 & 66.42 & 64.45 \\
                               & GRID     & 63.34 & 63.46 & 63.30 & 63.37 \\
    \midrule
    \multirow{2}{*}{FT5-base}  & PP & 70.81 & 60.83 & 67.25 & 66.30 \\
                               & GRID     & 70.76 & 61.54 & 70.05 & 67.45 \\
    \midrule
    \multirow{2}{*}{T5-large}  & PP & 73.57 & 73.63 & 74.26 & 73.82 \\
                               & GRID     & 72.29 & 74.84 & 73.07 & 73.40 \\
    \midrule
    \multirow{2}{*}{FT5-large} & PP & 75.98 & 73.98 & 74.84 & 74.93 \\
                               & GRID     & 76.28 & 71.56 & 74.72 & 74.19 \\
    \bottomrule
  \end{tabular}
\caption{\small Average test accuracy on negative transfer task sequences (NT1–NT3) comparing our method with Progressive Prompts. Our method consistently outperforms the baseline across all models, demonstrating better generalization under reduced task similarity.}
  \label{Table6}
\end{table}

\begin{table}[t]
  \centering
  \scriptsize
  \setlength\tabcolsep{4pt}
  \begin{tabular}{@{} l l r r r r r r r @{}}
    \toprule
    \rowcolor{gray!25}
    \textbf{Model}      & \textbf{Set.}
      & \textbf{L1} & \textbf{L2} & \textbf{L3}
      & \textbf{L4}   & \textbf{L5}   & \textbf{L6} & \textbf{Avg} \\
    \midrule
    \multirow{2}{*}{T5s}  & PP
      & -5.07  & -11.28 &  -1.48
      & -6.09  &  -9.58 &  -8.63 &  -7.02 \\
                               & GRID
      & -5.82  &  -0.11 &  -5.55
      & -0.96  &  -1.11 &  -3.33 &  -2.81 \\
    \midrule
    \multirow{2}{*}{FT5s} & PP
      & -3.57  &  -5.26 &  -4.98
      & -5.33  &  -4.92 &  -4.64 &  -4.78 \\
                               & GRID
      & -2.53  &  -4.09 &   3.33
      &  0.25  &  -3.02 &  -4.64 &  -1.78 \\
    \midrule
    \multirow{2}{*}{T5b}   & PP
      & -3.02  &  -7.02 &  -4.80
      & -4.55  &  -1.58 &  -4.23 &  -4.20 \\
                               & GRID
      & -4.93  &  -6.68 &  -5.78
      & -11.32 &  -2.21 & -13.28 &  -7.37 \\
    \midrule
    \multirow{2}{*}{FT5b}  & PP
      & -13.91 &  -2.08 &  -1.48
      & -4.90  &  -2.60 &  -4.53 &  -4.92 \\
                               & GRID
      &  4.27  &   5.10 &   4.04
      &  5.32  &   0.18 &  -7.90 &   1.84 \\
    \midrule
    \multirow{2}{*}{T5l}  & PP
      & -4.14  &  -1.19 &  -5.58
      & -2.66  &  -2.67 &  -3.81 &  -3.34 \\
                               & GRID
      & -2.82  &  -1.20 &  -0.90
      & -3.11  &  -2.20 &  -6.28 &  -2.75 \\
    \midrule
    \multirow{2}{*}{FT5l} & PP
      & -0.73  &  -3.17 &  -6.65
      & -1.43  &  -1.53 &  -0.99 &  -2.42 \\
                               & GRID
      &  2.35  &   0.75 &  -1.48
      & -0.64  &  -0.60 &  -0.90 &  -0.09 \\
    \bottomrule
  \end{tabular}
  \caption{Forward Transfer (FWT) scores across various task orderings (Order L1-L6) for T5 and Flan-T5 models. Higher scores indicate better transfer to future tasks. Our method achieves comparable or improved FWT in most cases, demonstrating that reducing forgetting does not come at the cost of forward transfer.}
  \label{tab:FWT}
\end{table}

\begin{table}[t]
  \centering

  \scriptsize
  \setlength\tabcolsep{4pt}
  \begin{tabular}{@{}l l c c c c@{}}
    \toprule
    \rowcolor{gray!25}
    \textbf{Model} & \textbf{Setting} & \textbf{NT1} & \textbf{NT2} & \textbf{NT3} & \textbf{Avg} \\
    \midrule
    \multirow{2}{*}{T5-small}  & PP & -5.86 & -7.62 & -5.62 & -6.37 \\
                               & GRID     & -4.67 & -7.65 & -6.87 & -6.40 \\
    \midrule
    \multirow{2}{*}{FT5-small} & PP & -3.93 & -9.99 & -7.88 & -7.27 \\
                               & GRID     & -4.96 & -4.58 & -6.47 & -5.34 \\
    \midrule
    \multirow{2}{*}{T5-base}   & PP & -7.44 & -1.35 & -1.44 & -3.41 \\
                               & GRID     & -6.52 & -5.56 & -6.64 & -6.24 \\
    \midrule
    \multirow{2}{*}{FT5-base}  & PP & -1.29 & -12.68 & -5.35 & -6.44 \\
                               & GRID     & -1.43 & -11.72 & -2.05 & -5.07 \\
    \midrule
    \multirow{2}{*}{T5-large}  & PP & -2.55 & -2.34 & -1.62 & -2.17 \\
                               & GRID     & -4.70 & -1.85 & -3.76 & -3.44 \\
    \midrule
    \multirow{2}{*}{FT5-large} & {PP} & 0.09  & -2.19 & -0.76 & -0.95 \\
                               & GRID     & 1.06  & -4.39 & -0.77 & -1.37 \\
    \bottomrule
  \end{tabular}
  \caption{Comparison of FWT scores between Progressive Prompts and Ours across different models. The table reports average performance on different tasks order , including Order NT1, NT2, NT3, and the overall average (Avg).}  
  \label{tab:model_results_10}
\end{table}

\begin{table*}[ht]
  \scriptsize
  \centering

\begin{tabular}{@{}p{2.5cm} p{2cm} p{2cm} >{\centering\arraybackslash}p{1.2cm} >{\centering\arraybackslash}p{0.8cm} >{\centering\arraybackslash}p{1.5cm} >{\centering\arraybackslash}p{0.8cm}@{}}
\toprule
\rowcolor{gray!25}
\textbf{Premise} & \textbf{Choice1} & \textbf{Choice2} & \textbf{Question} & \textbf{Label} & \textbf{ProgPrompt} & \textbf{GRID} \\
\midrule
The man lost the competition. & The competition was sabotaged. & He intimidated his competitors. & cause & 0 & false & choice1 \\
\midrule
I regained composure from my fit of anger. & My heart pounded. & I took deep breaths. & cause & 1 & true & choice2 \\
\midrule
The cook's eyes watered. & He ran out of onions. & He cut an onion. & cause & 1 & true & choice2 \\
\midrule
The tree branch landed in the river. & The branch moved downstream. & The river's current became stronger. & effect & 0 & false & choice1 \\
\midrule
The woman retired. & She received her pension. & She paid off her mortgage. & effect & 0 & false & choice1 \\
\bottomrule
\end{tabular}
\caption{
\small Examples highlighting \textbf{inconsistent label mappings} in COPA under task-agnostic inference. Prior works like Progressive Prompts use manual conversions (e.g., 0 $\rightarrow$ ``false''), often misaligning task semantics. GRID mitigates these inconsistencies by inferring appropriate labels automatically.
}
\label{appendix:incons_label}

\end{table*}

% \begin{figure}[htbp]
% \centering
% \includegraphics[width=0.4\textwidth]{images/Analysis/cluster2_v1.pdf}

%     \caption{\small Dataset embeddings using all-MiniLM-L6-v2.
% }
%     \label{fig:cluster2}
% \end{figure}

% \begin{figure*}[!ht]
%     \centering
%     \small
%     \includegraphics[width=\textwidth, clip]{images/T5-large/O9/o9_4.pdf}
%     \caption{\small Per-task BWT comparison between our method (blue) and the baseline (red) for Order L2. Positive bars indicate improved retention of prior tasks.}
% \label{fig:o5_bwt}
% \end{figure*}

% ----------------------------------------------------

% \begin{figure*}[t]
%     \centering

%     % ---------------- LEFT COLUMN ----------------
%     \begin{minipage}[t]{0.48\textwidth}
%         \centering
%         \includegraphics[width=0.9\columnwidth]{images/T5-large/easy/e4.pdf}
%         \captionof{figure}{
%         (a) Order L4.
%         }
%         \label{fig:oa_bwt}
%     \end{minipage}
%     \hfill
%     % ---------------- RIGHT COLUMN ----------------
%     \begin{minipage}[t]{0.48\textwidth}
%         \centering
%         \includegraphics[width=0.9\columnwidth]{images/T5-large/hard/h4.pdf}
%         \captionof{figure}{
%         (b) Order L5.
%         }
%         \label{fig:ob_bwt}
%     \end{minipage}

%     \caption{
%     Per-task BWT comparison between GRID (blue) and ProgPrompt (red)
%     under varying task difficulty.
%     Positive bars indicate improved retention of prior tasks,
%     demonstrating robustness across both easy and hard task orders.
%     }
%     \label{fig:bwt_easy}
% \end{figure*}

% \begin{figure*}[!ht]
%     \centering
%     \small
%     \includegraphics[width=\textwidth, clip]{images/T5-large/O10/o10_4.pdf}
%     \caption{\small Per-task BWT comparison between our method (blue) and the baseline (red) for Order L3. Positive bars indicate improved retention of prior tasks.}
% \label{fig:o6_bwt}
% \end{figure*}

% ----------------------------------------------------

% ----------------------------------------------------

% ----------------------------------------------------

\begin{figure}[htbp]
    \centering
    \includegraphics[width=\linewidth]{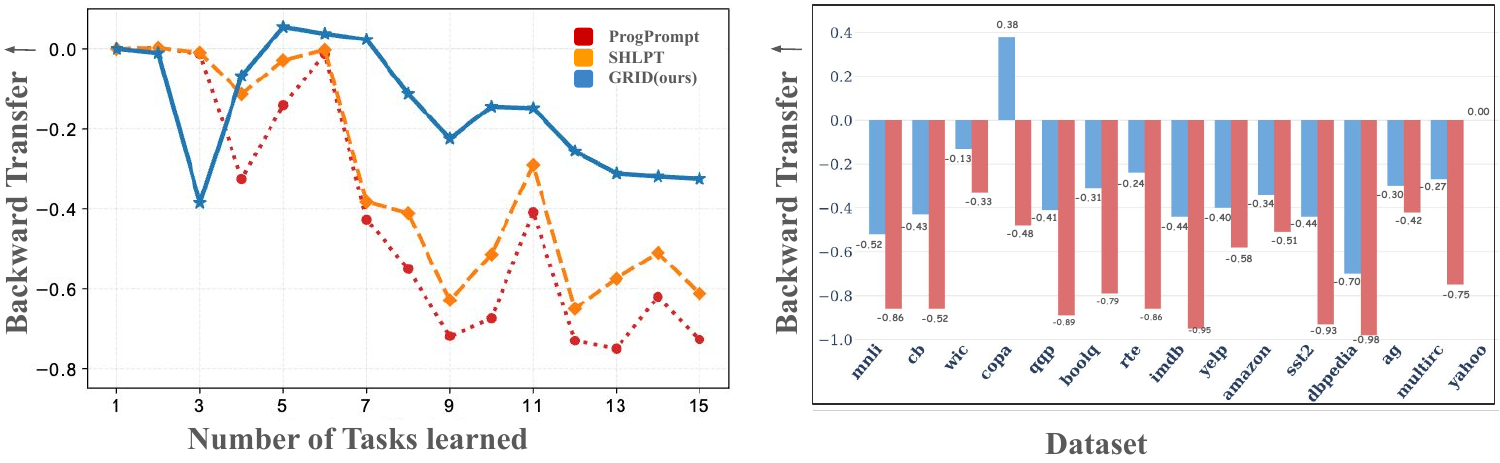}
    \caption{\small
    (a) Left - Backward transfer dynamics across sequence (L1).
    (b) Right - Per-task BWT comparison for Order L1, where GRID (in blue) consistently improves retention over  ProgPrompt (in red).
    }
    \label{fig:bwt_and_analysis}
\end{figure}

\begin{table*}[htbp]
% \vspace{-0.1cm}
\centering
\scriptsize
\setlength\tabcolsep{7pt}
\renewcommand{\arraystretch}{1.15}

\begin{tabular}{llccc ccc ccc}
\toprule
\multirow{2}{*}{\textbf{Model}} & \multirow{2}{*}{\textbf{Strategy}} & 
\multicolumn{3}{c}{\textbf{L1}} & \multicolumn{3}{c}{\textbf{L2}} & \multicolumn{3}{c}{\textbf{L3}} \\
\cmidrule(lr){3-5} \cmidrule(lr){6-8} \cmidrule(lr){9-11}
 & & \textbf{Acc} & \textbf{BWT} & \textbf{FT} & \textbf{Acc} & \textbf{BWT} & \textbf{FT} & \textbf{Acc} & \textbf{BWT} & \textbf{FT} \\
\midrule
\multirow{3}{*}{T5-large}
  & Random & 76.81 & -0.3482 & 15 & \textbf{82.07} & -0.3549 & 7  & \textbf{82.06} & -0.4065 & 21 \\
  & FIFO   & 76.08 & -0.3389 & 14 & 77.39 & -0.3780 & 11 & 81.34 & \textbf{-0.3962} & 20 \\
  & GRID   & \textbf{79.12} & \textbf{-0.3243} & \textbf{11} & 80.69 & \textbf{-0.3336} & \textbf{5} & 81.09 & -0.3979 & \textbf{18} \\
\midrule
\multirow{3}{*}{Qwen-3-4B}
  & Random & 81.75 & -0.4767 & 23 & 82.15 & -0.4835 & 17 & 82.28 & -0.4705 & 25 \\
  & FIFO   & 80.62 & -0.4781 & 23 & 77.05 & -0.4981 & 16 & \textbf{83.90} & -0.4633 & 24 \\
  & GRID   & \textbf{83.54} & \textbf{-0.4243 } & \textbf{18} & \textbf{82.32} & \textbf{-0.4198 } & \textbf{15} & 83.87 & \textbf{-0.4812} & \textbf{23} \\
\bottomrule
\end{tabular}
\caption{\small Comparison of prompt selection strategies (Random, FIFO, Gradient-based) for T5-large and Qwen-3-4B across order L1–L3. Metrics: average accuracy (Acc), BWT, and number of forgotten tasks (FT).}
\label{compact_strategy_comparison}
\end{table*}

\begin{table*}[htbp]
% \vspace{-0.2cm}
\centering
\scriptsize
% \footnotesize
\setlength{\tabcolsep}{7pt}
\renewcommand{\arraystretch}{1.15}

% S column formats: one sign + 1 digit + 4 decimals
\begin{tabular}{
l
l
S[table-format = -1.4]
S[table-format = -1.4]
S[table-format = -1.4]
S[table-format = -1.4]
S[table-format = 3.0]
l
}
\toprule
\rowcolor{gray!25}
\textbf{Model} & \textbf{Variant} & {\textbf{L1}} & {\textbf{L2}} & {\textbf{L3}} & {\textbf{Avg}} & {\textbf{Memory}} & {\textbf{GPU (h:m)}} \\
\midrule
\multirow{5}{*}{T5-large}
& \textbf{(0) G.R.I.D.}     & \bfseries -0.3243 & \bfseries -0.3310 & \bfseries -0.3979 & \bfseries -0.3511 & \bfseries 200 & \texttt{27:08} \\
& (1) w/o G                 & -0.3254 & -0.3321 & -0.3895 & -0.3490 & 600 & \texttt{25:35} \\
& (2) w/o G,D               & -0.7032 & -0.7589 & -0.5967 & -0.6863 & 600 & \texttt{26:42} \\
& (3) w/o G,D,R             & -0.7155 & -0.7612 & -0.5993 & -0.6954 & 600 & \texttt{24:58} \\
& (4) w/o all               & -0.7275 & -0.7625 & -0.6137 & -0.7012 & 600 & \texttt{23:58} \\
\midrule
\multirow{5}{*}{Qwen}
& \textbf{(0) G.R.I.D.}     & \bfseries -0.4243  & \bfseries -0.4198  & \bfseries -0.4812 & \bfseries -0.4418  & \bfseries 200 & \texttt{84:72} \\
& (1) w/o G                 & -0.4145 & -0.4210 & -0.4823 & -0.4393 & 600 & \texttt{81:38} \\
& (2) w/o G,D               & -0.8832  & -0.8289 & -0.6754 & -0.7958 & 600 & \texttt{81:35} \\
& (3) w/o G,D,R             & -0.8855 & -0.8297 & -0.6876 & -0.8009 & 600 & \texttt{80:46} \\
& (4) w/o all               & -0.8987 & -0.8421 & -0.6934 & -0.8114 & 600 & \texttt{78:47} \\
\bottomrule
\end{tabular}
\caption{\small Ablation on GRID over orders L1–L3. We report BWT (less negative is better), average across orders, prompt memory size in KB (slots), and GPU time per run on A100 (40GB) with batch size 8.}
\label{ablation}

\end{table*}

\end{document}